\documentclass [10pt]{article}
\usepackage[utf8]{inputenc}
\usepackage{amsmath}
\usepackage{amsfonts}
\usepackage{amssymb}
\usepackage{graphicx}
\usepackage{subcaption}
\usepackage{soul}
\usepackage{parskip}
\usepackage{natbib}
\usepackage{bbm}
\usepackage{algorithm}
\usepackage{bm}
\usepackage[noend]{algpseudocode}
\usepackage{booktabs}
\usepackage{tabularx}
\usepackage{xcolor}
\usepackage{url}
\usepackage{algorithmicx}
\usepackage{multirow}
\usepackage{lscape}
\usepackage{pifont}

\title{\textbf{A Multi-Task Evaluation of LLMs' Processing of Academic Text Input}}

\author{\textsc{Tianyi Li}\thanks{tianyi.li@cuhk.edu.hk. Co–first author.} \\[1ex] 
Department of Decisions, Operations and Technology, CUHK \\
\\
\textsc{Yu Qin}\thanks{yuqin@asu.edu. Co–first author.}, \textsc{Olivia R. Liu Sheng}\thanks{olivia.liu.sheng@asu.edu} \\[1ex] 
Department of Information Systems, Arizona State University}
\date{}

\begin{document}

\maketitle

\begin{abstract}

\noindent How much large language models (LLMs) can aid scientific discovery, notably in assisting academic peer review, is in heated debate. Between a literature digest and a human-comparable research assistant lies their practical application potential. We organize individual tasks that computer science studies employ in separate terms into a guided and robust workflow to evaluate LLMs’ processing of academic text input. We employ four tasks in the assessment: content reproduction/comparison/scoring/reflection, each demanding a specific role of the LLM (oracle/judgmental arbiter/knowledgeable arbiter/collaborator) in assisting scholarly works, and altogether testing LLMs with questions that increasingly require intellectual capabilities towards a solid understanding of scientific texts to yield desirable solutions. We exemplify a rigorous performance evaluation with detailed instructions on the prompts. Adopting first-rate Information Systems articles at three top journals as the input texts and an abundant set of text metrics, we record a compromised performance of the leading LLM - Google's Gemini: its summary and paraphrase of academic text is acceptably reliable; using it to rank texts through pairwise text comparison is faintly scalable; asking it to grade academic texts is prone to poor discrimination; its qualitative reflection on the text is self-consistent yet hardly insightful to inspire meaningful research. This evidence against an endorsement of LLMs' text-processing capabilities is consistent across metric-based internal (linguistic assessment), external (comparing to the ground truth), and human evaluation, and is robust to the variations of the prompt. Overall, we do not recommend an unchecked use of LLMs in constructing peer reviews. 
\end{abstract}


\section{Introduction}

Using artificial intelligence (AI) for scientific discovery is an ambitious vision for the future of science. The discussion has drawn much attention in past decades \citep[e.g.,][]{LBet1993,KWet2004,KRet2009,WB2009} and is gaining enormous momentum along the recent surge in AI \citep[e.g.,][]{Z2017,E2018,K2021,SE2023}, notably the AI for Science campaign \citep[e.g.,][]{WFet2023}. ``Self-driving labs'' \citep{C2020} transform science labs into automated factories of discovery \citep{ACAet2024}, aiming to help chemical syntheses \citep{DVet2024} especially drug discovery \citep{V2023}, material design \citep{JAet2023}, mathematical proofs \citep{FPet2024} and even the development of physical concepts \citep{IMet2020,He2024}, despite warnings such as the reproducibility crisis \citep{KN2023}, the danger of ``fast science'' \citep{F2020}, the illusions of scientific understanding \citep{MC2024}, and the over-reliance on AI \citep{NK2025}.

Generative AI, in particular large language models (LLMs), joins the landscape in a strong whirl. Advancing conventional computational tools which emphasize machines' capabilities in search, analysis, and optimization that help override ``human bottlenecks'' \citep{GGet2014}, LLMs proclaim more capabilities especially via advanced text-processing to help address intellectual tasks including creative works \citep{CC2024}. People envision employing LLMs to facilitate scholarly work during problem formulation, hypothesis generation, research design, data collection/analysis, interpretation/theorization, and composition/writing \citep{AMet2024,ACNet2024,N2024}, despite practical challenges \citep[e.g.,][]{B2023}, impediments from the LLM explosion \citep[e.g.,][]{APet2024}, biases in their output \citep[e.g.,][]{FCet2024}, and researchers' mixed feelings on this new technology \citep[e.g.,][]{P2024}. See Wiley's ExplanAItions study\footnote{https://www.wiley.com/en-cn/ai-study} on AI's role in research and insights into how researchers are or are not embracing AI. 

\subsection{The grand debate: How much can LLMs aid scientific discovery?}

On the ground level, LLMs' excellent capabilities in summarizing and paraphrasing promise to help researchers digest scientific literature, which is in tremendous growth over the years \citep{L2016,BHet2021}. We can use LLMs to filter content, condense texts, extract findings, and communicate the science \citep{ACet2024}. This application of LLMs is propelled by the increasing use of academic papers as the training data \citep{Gib2024} and gets mature along with the development of commercial products such as Elicit (https://elicit.com), Semantic Reader\footnote{https://www.semanticscholar.org/product/semantic-reader}, and Elsevier's Scopus AI. These tools nonetheless encounter barriers, such as the lack of accuracy and verifiability, restricted access to full texts, and non-machine-readable formats, bringing inevitable limitations \citep{C2023} and calling for tremendous system amelioration. As a result, although LLMs may help increase productivity and confidence in writing \citep{LLet2024a}, many people display negative perceptions of LLM-assisted writing \citep{LLet2024b}.

A prominent question is to what extent the other envisioned applications of LLMs aiding scientific discovery, which build on their basic function of literature digest, can be validated. 

(+) To one extreme, some argue that LLMs have the potential to evolve into a full-fledged \textit{research assistant} with solid scientific understanding; in this picture, with the continuous advancement of their technical core, LLMs will have a prodigious impact on occupations closely related to scientific discovery which are among those having the highest share of tasks exposed to LLMs \citep[][Figure 16 in Supplementary Materials]{EMet2024}. For example, the recent development of an ``AI-scientist for fully automated open-ended scientific discovery'' \citep{LLet2024} which can produce formal manuscripts for computer science (CS) conference submissions from a complete workflow of idea generation, code building, experiment execution, and writing and review, adds a sounding weight to this conjecture; the multimodal AI copilot PathChat \citep{LCet2024} can produce ``accurate and pathologist-preferable responses to diverse queries related to pathology'' and considerably help education, research, and clinical decision-making.

(-) Despite the promising landscape, many remain skeptical of this ambition and contend that scientific discovery is different from scientific understanding and the current AI-assisted approach is not and will not easily reach the level of understanding (e.g., see the discussions in \textit{Science} by Melanie Mitchell)\footnote{https://www.science.org/authored-by/Mitchell/Melanie} which should ``recognize qualitatively characteristic consequences of a theory without performing exact computations and use them in a new context'' or effectively ``teach human experts'' \citep{DD2005,GLet2024}. Therefore, instead of viewing AI tools as an ``agent of understanding'' \citep{KPet2022}, we should view them as an artificial agency that embodies sufficient complexity but without the capability of understanding. For example, with the AI scientist in \citet{LLet2024}, one would argue that its workflow of science production closely follows human instructions and the produced manuscript is strictly structured by human input; the ``automatic'' discovery relies on much intervention from the human supervisor.

The practical application potential of LLMs lies somewhere between that of a literature digest and a human-comparable research assistant—the two ends that span the spectrum of LLMs' intellectual capabilities. Studies evaluating LLMs should look into this spectrum and investigate LLMs' capabilities from the ground up to the full-fledged scenario.

\subsection{The focal debate: How feasible is LLM-assisted peer review?}

Specifically, of AI's many potential applications for scientific discovery, AI-assisted peer review \citep{CBet2021,VTet2021,WLet2022} stands out as having promising conditions to materialize in the near term. Peer review lies at the core of science production \citep{S2002,R2016,SWet2023,B2024} yet increasingly suffers from speed and objectivity issues, among others \citep[e.g.,][]{S2022}; academia has been testing innovative ways \citep[e.g.,][]{TDet2017,KPet2020} to accelerate this heavy-laden research evaluation process \citep[e.g.,][]{KPet2016,A2020} and mitigate the bias within it \citep[e.g.,][]{LSet2013,LWet2018}.

Progress in natural language processing and machine learning in general allows people to imagine employing computational tools to automate the peer review of academic texts \citep[e.g.,][]{KAet2024} presumably with an enhanced speed and objectivity, and potentially with better structuring \citep{M2024}. Predictive AI tools such as machine classifiers have already been implemented in the peer review workflow, e.g., for plagiarism/misinformation/error detection and statistics/structure/method checks during the initial quality control of submitted manuscripts \citep[e.g.,][]{KT2024}. Generative AI and in particular LLMs holds great promise in this application, and researchers are extensively investigating the feasibility of LLM-assisted peer review \citep[e.g.,][]{LS2023,LZet2024,R2023,DWet2024} and are developing trial tools \citep[e.g.,][]{DHet2024,JZet2024,SCet2024}. 

The results from initial investigations reside in two camps. 

(+) Some studies support the utility of LLMs. For example, in a large-scale empirical study, \citet{LZet2024} reported that a dataset of GPT-4-generated reviews aligned reasonably with human reviews and a large portion of the surveyed user base found LLM-generated reviews useful; however, GPT-4 focused on certain aspects of the feedback and struggled to provide in-depth critiques.

(-) Other studies question the augmented use of the tool. For example, \citet{LS2023} reported that GPT-4, outperforming several other LLMs, did well in error identification and submission checklist verification but not so in choosing the better abstract from two candidates; the tested LLM was thus not yet suitable for a complete evaluation of papers or proposals despite a promising use in specific review tasks. 

The potential and possible pitfalls of employing LLMs in aiding academic peer review need careful examination to release the power of the tool and avoid bad practice \citep{SMet2023,LIet2024,LZ2024,Z2024,BBet2025,N2025}. While the literature on leveraging LLMs in text assessment such as in dialogue quality measurement and evaluation \citep[e.g.,][]{HJet2023,JKet2024} expands rapidly, studies focusing on academic text input are at an infant stage. Existing works also often consider particular tasks around peer review while lacking an overarching investigation. 

\subsection{Our guided evaluation}

In a seminal discussion, \citet{MC2024} proposed a conceptual framework to help address the ongoing debates. They argue that AI can serve four roles -- \textit{oracle}, \textit{surrogate}, \textit{quant}, and \textit{arbiter} -- in scientific assistantship, where the digital tools are primarily used for literature digest, data collection, data analysis, and content evaluation, respectively. The idea of AI serving as a human-like collaborator with solid scientific understanding is rejected, and they imply that the increasing epistemic trust people grant to AI that builds on its presumed objectivity, proclaimed analytical depth, and pre-acknowledged efficiency incites an over-optimism on employing AI in assisting scientific research.

Among the four roles of AI in scientific assistantship, the oracle role and the arbiter role, which ask the AI to directly process academic text input, apply readily to LLMs. In this study, we develop a guided evaluation echoing \citet{MC2024}'s conceptual framework to assess LLMs' processing of academic text input and investigate its practical potential in scientific discovery particularly in assisting academic peer review. We focus on LLMs' processing of academic text \textit{input} instead of their processing of academic texts \textit{in general}, e.g., searching and summarizing the existing literature.

We employ four tasks assembled into a guided workflow to conduct the assessment (see details in Section 2): \textit{content reproduction} (Task 1), \textit{content comparison} (Task 2), \textit{content scoring} (Task 3), and \textit{content reflection} (Task 4), while existing studies only employ a subset of them to evaluate LLMs (see Appendix A for these studies). We propose that each task demands a specific role of LLMs in assisting scientific discovery and tests different capabilities of the LLM; from Tasks 1 to 4, the difficulty for LLMs to satisfactorily complete the task increases, as a good solution there calls more and more for the proximity to a solid scientific understanding. 

The four tasks adhere to the components of a typical academic peer review:

\begin{itemize}

    \item \textit{Content reproduction} can occur when editors screen articles to determine whether to send manuscripts out for review or when they create content outlines (e.g., when reviewers produce summaries at the beginning of review reports).

    \item \textit{Content comparison} can assist editors, who are subject to limited reviewer pools and small publication selection quotas, in handling a large number of manuscripts, which is typical for conference submissions.

    \item \textit{Content scoring} can help evaluate the differential publishability of manuscripts where reviewers give quantitative scores to the reviewed manuscripts for editors' consideration and decision.

    \item \textit{Content reflection} can be called for when reviewers provide authors with qualitative and instructive feedback on their manuscripts, which constitutes the essential content in review reports.

\end{itemize}

These tasks thus help assess LLMs' application potential at different stages of an academic peer review: authors to editors, editors to reviewers, back to editors, and finally back to authors \citep{SMet2023}.

\subsection{Do NOT take LLMs' capabilities for granted}

Our experiments across the four tasks examine a spectrum of LLMs' capabilities to investigate their application potential, noting that LLMs' different capabilities should not be taken for granted.

In \citet{EMet2024}'s high-impact work on GPTs' influence on labor markets, LLMs' capabilities at summarizing medium-length documents (corresponding to \textit{content reproduction}) and providing feedback on documents (corresponding to \textit{content reflection}) are juxtaposed as their basic capabilities. Similarly, in \citet{YBet2025}, multiple LLMs are assembled in a workflow to optimize the final output through the backpropagation of feedback, with each LLM playing a different function in the system: summarize, rate, or critique; the three functionalities are investigated in our evaluation (Tasks 1, 3, and 4). These two are typical examples in social science \citep{EMet2024} or engineering fields \citep{YBet2025} where one overlooks the difference in tasks for LLMs and stays over-confident in their capabilities. We warn that this blind assumption that LLMs could perform well at different functionalities is unwarranted.

Moreover, evidence suggests that although people found LLM-generated reviews helpful to a certain extent, e.g., at providing timely feedback, for early-stage manuscripts, or when the review has been well structured \citep{LZet2024,DT2024,G2024}, limitations of the current leading LLMs with regard to the features (in italic font) evaluated at our four tasks are apparent: 

\begin{itemize}

    \item At \textit{content reproduction}, \citet{TOet2024} reported that GPT-4 responded with 86.9$\%$ accuracy in reproducing content upon queries (acceptable \textit{reliability}).
    
    \item At \textit{content comparison}, \citet{LS2023} reported that the LLM committed errors six out of ten times when comparing two abstracts and selecting the outperforming one (unwarranted \textit{scalability}).

    \item At \textit{content scoring}, \citet{DT2024} reported that LLM scores were higher than human scores by a substantial positive bias of around 23$\%$ (skewed \textit{discrimination}).

    \item At \textit{content reflection}, \citet{LZet2024} reported that despite the ability to generate non-generic feedback, LLM commented on research implications 7.27 times more frequently, yet on research novelty 10.69 times less frequently, than humans did (bounded \textit{insightfulness}). 
    
\end{itemize}

Noting the limitations, we organize these tasks into an integrated workflow to measure these four important features of the LLM output. Our standpoint thus contrasts with those studies \citep[e.g.,][]{WPet2023} that advocate for LLMs’ intellectual capabilities in particular with respect to human intelligence. Overall, our results provide evidence of unwarranted over confidence in LLMs' capabilities across the four tasks. 

\subsection{Establishing an exemplary analysis}

LLMs' text-processing capabilities vary by the LLM and the input text. For the LLM, open- (to certain degrees \citep{Gia2024}) or closed-source products \citep{OLet2025} based on small- \citep[e.g.,][]{ZHet2024} to large-scale architectures are developed for special \citep[e.g.,][]{STet2024} or general applications, and are getting constantly upgraded under market competition. For the input academic text, articles from diverse scientific disciplines can differ considerably in content ingredients and semantic styles. To allow a comprehensive evaluation of LLMs' processing capabilities, it is imperative to conduct extensive studies with various LLM products and text samples.

A feasible idea for recruiting miscellaneous academic content is to use the articles from interdisciplinary journals or essentially, to focus on an interdisciplinary field. The discipline of Information Systems (IS) provides a suitable option, which is a field known for exhibiting interdisciplinary research across social, physical, and managerial sciences and supporting diverse scientific approaches in studying information technologies \citep{BW1996,R1996}. As such, academic texts in IS cover an extended range of topics, methodologies, and materials and demonstrate various semantic styles, compared to texts from a more concentrated discipline; this variety brings the necessary diversity of the input for testing the LLM. 

In this study, we present an exemplary analysis for our evaluation framework, focusing on demonstrating the effectiveness and robustness of its tasks and the computational workflow in processing text input from a single discipline by an LLM of choice. For the academic texts, we collect 246 articles recently published in three top journals (Information Systems Research, Journal of Management Information Systems, and IS Department of Management Sciences), all publishing diverse first-rate articles devoted to the IS field. For the LLM, we use the popular Google's Gemini Pro\footnote{https://ai.google.dev/gemini-api/docs/models/gemini}. The focal assessment is on whether it is desirable to recruit the tested LLM in aiding the review of IS articles, which has not been directly investigated by previous studies. Based on the focal evaluation, we show the potential of the guided framework in providing insights that benefit future evaluations applied to academic content in broad disciplines or other LLMs.

\begin{figure}[h]
  \centering
  \includegraphics[width=6.5in]{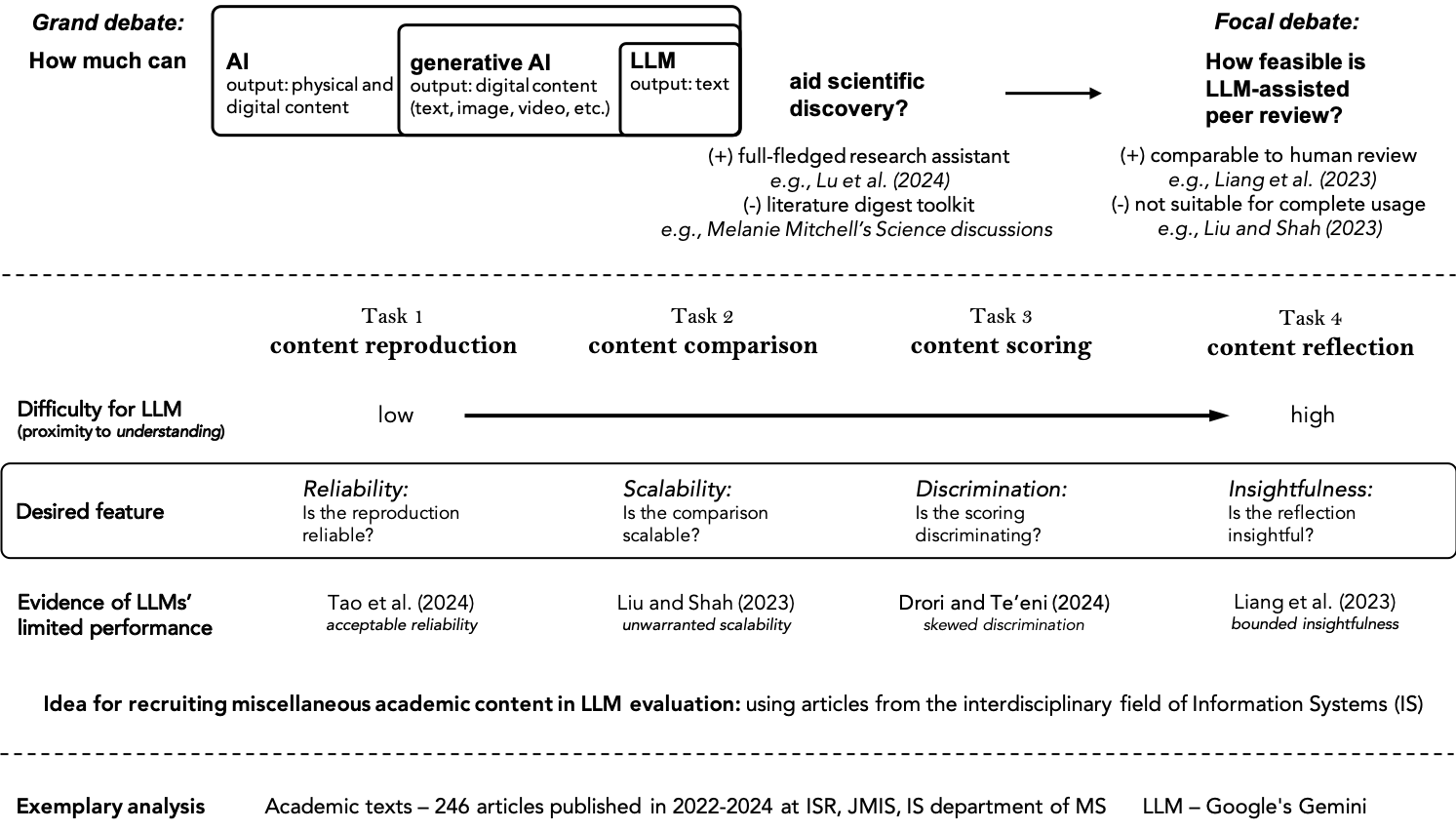}
  \caption{Research motivations and outline.}
  \label{fig1}
\end{figure}

Overall, the motivations and outline of this research are summarized in Figure \ref{fig1}. 

\subsection{Research contributions}

This study investigates a pivotal topic that draws increasing attention from different research disciplines. With global investment in AI—particularly in large language models (LLMs)—estimated to reach trillions over the past and coming decades \citep{Le2024}, research on LLMs is expected to remain active. Information Systems (IS) scholars bear the responsibility of evaluating their role, performance, and deployment.

Our evaluation workflow follows the design science principles in IS studies \citep{HMet2004,GnH2013,RBet2017,APet2024} and the steps of classical artifacts \citep{PCet2015} of computational frameworks \citep[e.g.,][]{CW1994,AC2008,SGet2024}. Supplementing technical viability to the conceptual discussion on AI and the future of research, we contribute a context to help connect IS to broader audiences \citep{PFet2022} in surrounding fields with qualitative \citep[e.g.,][]{MC2024} or quantitative (e.g., computer sciences) methodologies.

Studying the behavior of LLMs in aiding future research during our human-machine interaction \citep{JPet2021,CCet2022} is essential for developing the abilities to control the actions of these artificial agents, reap their benefits, and minimize their harms \citep{RCet2019}. This is important for understanding the ``Janus effects'' of AI tools and promoting their responsible use \citep{SGet2023}. As such, we answer the call in \citet{YHet2024} for future IS studies to have ``balanced attention to both the positive and negative externalities of digital innovations'' and in particular ``being aware of and actively engaging with the potential negative consequences of digital innovation'', while ``[believing] that nuanced, human-centered, responsibly designed technological innovations can play essential roles in addressing challenges.'' 

In a fresh debate, \citet{BAet2025} collects scientists' perspectives on LLMs' effect on the future practice of science: some see working with LLMs similar to working with humans, some deny the possibilities of AI scientists replacing humans and caution for misuse, while some envision AI and humans working together with us remaining responsible for determining the research roadmap. Following \citet{BBet2025}, we contend that the integration of generative AI in scholarly work is a ``means to augment, not replace, human contributions.'' In a future landscape where LLMs keep advancing, our work contributes a viable tool for their assessment; nevertheless, as issues such as the shortage of training data and the fall of the ``scaling law'' \citep[e.g.,][]{VHet2024} start to question the optimism around LLMs' continued upgrades, we emphasize on LLMs' limitations and the imperfect LLM output. Overall, securing the right path for future science asks for extensive interdisciplinary efforts; we try to contribute a responsible discussion from the IS field, where scholars are ``positioned to offer guidelines that ensure the ethical and effective use of generative AI across diverse business communities, facilitating interdisciplinary collaboration.'' \citep{BBet2025}


LLMs' imperfect processing of academic texts is nonetheless useful. An LLM acts effectively as an ``intelligent and well-rounded layman'': this artificial audience possesses the knowledge base of diverse scientific fields. This allows LLMs to produce a \textit{fair} evaluation of academic texts across disciplines: texts teemed with nomenclature (e.g., in biology), equations (e.g., in maths), or algorithms (e.g., in CS), can be assessed with an indifferent eye. A perfect literature digest may not help distinguish the texts as it outputs excellent summaries; an intermediate-level ``knowing-all-by-a-little'', which suitably describes early-stage LLMs, is effective for discriminating texts of diverse genres and qualities with its flawed output. Our evaluation thus does not chase state-of-the-art LLMs but embraces the imperfect LLM output, dismissing the worry about the obsolescence or sub-optimality of LLMs, which currently undergo rapid market transitions. 

Promisingly, LLMs' processing can be utilized to indicate text quality. This helps design measures for articles/journals/scholars, complementing current diffusion-based metrics (e.g., impact factor, H-Index \citep{W2016,AC2017}) that are rather discipline-specific \citep{SPet2021}, contributing to the amelioration of science evaluation \citep{A2025,S2025} (observing the Leiden Manifesto \citep{HWet2015}).


\newpage

\section{The Multi-task Evaluation}

\begin{figure}[h]
  \centering
  \includegraphics[width=6.5in]{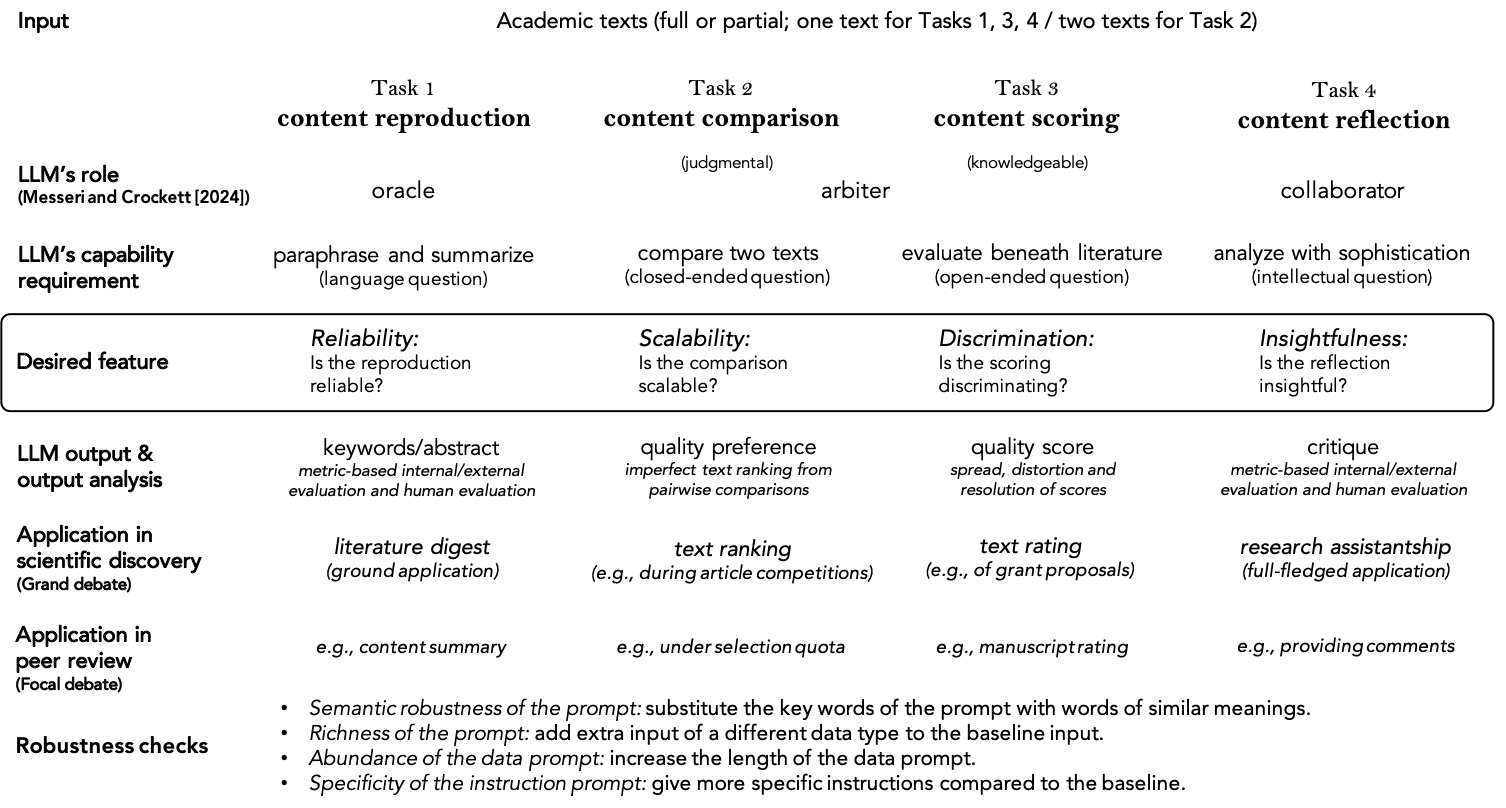}
  \caption{The multi-task evaluation of LLMs' processing of academic text input.}
  \label{fig2}
\end{figure}

We employ four tasks to study LLMs' processing of academic text input (Figure \ref{fig2}): \textit{content reproduction} (Task 1), \textit{content comparison} (Task 2), \textit{content scoring} (Task 3), and \textit{content reflection} (Task 4). 

At \textit{content reproduction}, we ask the LLM to reproduce short texts (keywords and abstract) based on the long input text. The LLM plays the role of \textit{oracle}. The task is intrinsically a language generation question, the least difficult for LLMs, and from the results, we evaluate LLMs' basic capability of paraphrasing and summarizing. We desire \textit{reliability} from the LLM output: the reproduced texts are reliable to the original text. This task assesses LLMs' basic application potential as literature digests.

At \textit{content comparison}, we ask the LLM to compare two input texts and output a quality preference. The LLM plays the role of arbiter; we further note that here it is a \textit{judgmental arbiter} that works based on the input material. The task is a closed-ended question, calling for LLMs' capability of making comparisons. We desire \textit{scalability} from the LLM output: the pairwise preferences can scale up. Scalable pairwise comparisons can underlie the application potential of LLMs at text ranking (e.g., during manuscript competitions).

At \textit{content scoring}, we ask the LLM to evaluate a single input text and output a quality score. The LLM plays the role of arbiter and in particular a \textit{knowledgeable arbiter}, as the text evaluation should rely on the knowledge of relevant scientific literature. The task is then an open-ended question, testing LLMs' capability of making quantitative evaluations. We desire \textit{discrimination} from the LLM output: the quality scores can effectively discriminate texts. A good content scoring can support LLMs' application in text rating (e.g., of grant proposals).

At \textit{content reflection}, we ask the LLM to examine the input text and output a list of critiques. In this most difficult task, the LLM eventually plays the role of \textit{collaborator} during science production. The task poses an intellectual question to the LLM, asking it to analyze the text with sophistication that adheres to scientific understanding. We desire \textit{insightfulness} from the LLM output: the qualitative reflections on the text bring insights. LLMs' potential as a full-fledged research assistant is not impossible if they succeed in this task.

Across the four tasks, the LLM is posed with questions that increasingly require intellectual capabilities towards a solid scientific understanding to yield desirable solutions. 
In each task, we assess whether a desired feature (reliability, scalability, discrimination, insightfulness) can be established from the LLM output, using different analytical tools to study the LLM output (Section 3). Notably, Tasks 1 and 4 evaluate LLMs' capabilities in outputting texts, while Tasks 2 and 3 evaluate LLMs' capabilities as judges \citep{ZCet2023}. Desirable performances in these tasks endorse LLMs' application potential at literature digest (upon reliable content reproduction), text ranking (upon scalable content comparison), text rating (upon discriminating content scoring), and finally full research assistantship (upon insightful content reflection), respectively. The explicit delineation of the roles of the LLM, the means of the evaluation, and the objectives of the assessment conforms closely to IS design science principles \citep{HMet2004,PTet2007}.

A major issue in LLM evaluation is the lack of robustness due to prompt diversity and probabilistic nature of the LLM output. The engineering of LLM prompts, which consists of the data part (input data for LLMs to process) and the instruction part (output instructions for LLMs to follow), is a central topic in LLM research that denies simple discussions \citep[e.g.,][]{L2024, SIet2024}. In our experiments, we vary the baseline prompts and conduct robustness checks, considering the following four aspects of the LLM prompts' robustness:

\begin{itemize}
    \item \textit{Semantic robustness of the prompt (R1).} We substitute the keywords of the prompt with words of similar meanings; we test if the prompt is robust to semantic variations.

    \item \textit{Richness of the prompt (R2).} We add extra input of a different data type to the baseline input; we test if enhancing the richness of the prompt affects the quality of the LLM output.

    \item \textit{Abundance of the data prompt (R3).} We increase the length of the data prompt; we test if enhancing the abundance of the data prompt affects the quality of the LLM output. 

    \item \textit{Specificity of the instruction prompt (R4).} We give more specific instructions compared to the baseline; we test if enhancing the specificity of the instruction prompt affects the quality of the LLM output. 
    
\end{itemize}

Below, we explain the baseline prompts and their variants (following the rationales of R1-R4). To account for LLMs' probabilistic output, we conduct five runs under each prompt and analyze the result ensemble. Our experiments are summarized in Table \ref{tab2}. \\

\textbf{\normalsize Task 1: Content Reproduction} 

Experiment 1-0: Input the introduction and conclusion sections of a scientific article. Ask the LLM to generate a list of keywords for this text.
 
\textit{Baseline Prompt.} ``For a scientific article (in the field of Information Systems) with the following `Introduction' and `Conclusion' sections, generate a list of appropriate keywords (please only output necessary keywords and separate the keywords by commas). `Introduction:' $\{$introduction$\}$. `Conclusion:' $\{$conclusion$\}$.''

\textit{Robustness checks.} Building on Experiment 1-0, investigate the following variants of the baseline prompt.

\begin{itemize}
    \item Experiment 1-1, 1-2: Input only the introduction or the conclusion section. (R3)

    \item Experiment 1-3: Include the article title in the prompt. (R2)

    \item Experiment 1-4: Specify the number of keywords to be generated. (R4)

    \item Experiment 1-5: Demand that the output keywords be ordered by importance. (R4)
\end{itemize}

Experiment 2-0: Input the introduction and conclusion sections of a scientific article. Ask the LLM to generate an abstract for this text.

\textit{Baseline Prompt.} ``For a scientific article (in the field of Information Systems) with the following `Introduction' and `Conclusion' sections, generate a one-paragraph abstract that contains no more than 300 words (please only output the main abstract). `Introduction:' $\{$introduction$\}$. `Conclusion:' $\{$conclusion$\}$.''

\textit{Robustness checks.} Building on Experiment 2-0, investigate the following variants of the baseline prompt.

\begin{itemize}
    \item Experiment 2-1, 2-2: Input only the introduction or the conclusion section. (R3)

    \item Experiment 2-3: Include the ground-truth keywords in the prompt. (R2)
    
    \item Experiment 2-4: Include the LLM-generated keywords in the prompt. (R2)
    
    \item Experiment 2-5: Specify the length of the abstract to be generated. (R4)\\
\end{itemize}

\textbf{\normalsize Task 2: Content Comparison} 

Experiment 3-0: Input the introduction and conclusion sections of two scientific articles. Ask the LLM to compare the two texts and output a quality preference.

\textit{Baseline Prompt.} ``For two scientific articles (in the field of Information Systems) with the following `Introduction' and `Conclusion' sections, compare their quality and output which one is better. Article A: `Introduction:' $\{$introduction$\}$. `Conclusion:' $\{$conclusion$\}$. Article B: `Introduction:' $\{$introduction$\}$. `Conclusion:' $\{$conclusion$\}$.''

\textit{Robustness checks.} Building on Experiment 3-0, investigate the following variants of the baseline prompt.

\begin{itemize}
    \item Experiment 3-1/2/3: Change the word ``quality'' in the prompt to ``information density''/``scientific value''/``comprehension difficulty''. (R1)\\

\end{itemize}

\textbf{\normalsize Task 3: Content Scoring} 

Experiment 4-0: Input the introduction and conclusion sections of a scientific article. Ask the LLM to evaluate this text and output a quality score in [1, 10].

\textit{Baseline Prompt.} ``For a scientific article (in the field of Information Systems) with the following `Introduction' and `Conclusion' sections, evaluate the quality of this text on a scale of 1 (worst) to 10 (best) and output the score. `Introduction:' $\{$introduction$\}$. `Conclusion:' $\{$conclusion$\}$.''

\textit{Robustness checks.} Building on Experiment 4-0, investigate the following variants of the baseline prompt.

\begin{itemize}
    \item Experiment 4-1/2/3: Change the word ``quality'' in the prompt to ``information density''/``scientific value''/``comprehension difficulty''. (R1)
    \item Experiment 4-4: Include the ground-truth abstract in the prompt. (R2)
    \item Experiment 4-5: Specify the score interval (set as 0.1). (R4)\\
\end{itemize}

\textbf{\normalsize Task 4: Content Reflection} 

Experiment 5-0: Input the introduction and conclusion sections of a scientific article. Ask the LLM to reflect on this text and generate a list of critiques. 

\textit{Baseline Prompt.} ``For a scientific article (in the field of Information Systems) with the following `Introduction' and `Conclusion' sections, output a list of critiques (in at most 3 bullet points) on its content. `Introduction:' $\{$introduction$\}$. `Conclusion:' $\{$conclusion$\}$.''\\ 

\textbf{\large Output analysis} 

We collect the LLM output to conduct analyses. (A) \textit{Objective analyses.} For Tasks 1 and 4, we conduct metric-based internal (linguistic assessment) and external (comparing to the ground truth (Task 1) or the text input (Task 4)) evaluation of the LLM-generated keywords, abstracts, and critiques. For Task 2, we construct a ranking over the text ensemble based on LLM's pairwise preferences. For Task 3, we investigate the effectiveness of discrimination of the LLM-output text scores. (B) \textit{Subjective analyses.} We recruit human arbiters to evaluate LLM-generated abstracts (Task 1) and critiques (Task 4). See details in Section 3.

\newpage

\begin{landscape}
\begin{table}[h!]
\centering
\resizebox{9in}{!}{%
\begin{tabular}{c|c|c|c|c|c}
\hline
 &  &  &  & \multicolumn{2}{c}{} \\
 & \textbf{Input} & \textbf{Output} & \textbf{Prompt Variants} (for robustness checks) & \multicolumn{2}{c}{\textbf{Output Analysis} (objective $|$ subjective)} \\
 &  &  &  & \multicolumn{2}{c}{} \\
\hline \hline
\multirow{4}{*}{\textbf{Task 1 --}} & & & (E1-1/2) Input only Introduction/Conclusion. & &  \\
& & Keywords & (E1-3) Include the article title. & Internal/external evaluation of & \\
& & (E1-0) & (E1-4) Specify the number of keywords. & LLM-generated keywords & \\
& Introduction, Conclusion & & (E1-5) Order keywords by importance. & &  \\ 
\cline{3-6}
\textbf{Content} & of one scientific article & & (E2-1/2) Input only Introduction/Conclusion. & & \\
\textbf{Reproduction} & & Abstract & (E2-3) Include ground-truth keywords. & Internal/external evaluation of & Human evaluation of \\
& & (E2-0) & (E2-4) Include LLM-generated keywords. & LLM-generated abstracts & LLM-generated abstracts\\
& & & (E2-5) Specify the abstract length. & & \\ 

\hline

& & & & & \\

\textbf{Task 2 --} & Introduction, Conclusion & Quality & (E3-1/2/3) Compare the texts' ``information density/scientific & Evaluation of the constructed &   \\
\textbf{Content Comparison} & of two scientific articles & preference & value/comprehension difficulty''instead of ``quality''. & text ranking & \\ 
& & (E3-0) & & & \\

\hline

& & & & & \\

& & & (E4-1/2/3) Evaluate the text's ``information density/scientific & & \\
\textbf{Task 3 --} & Introduction, Conclusion & Quality & value/comprehension difficulty''instead of ``quality''. & Evaluation of the output & \\
\textbf{Content Scoring} & of one scientific article & score & (E4-4) Include the ground-truth abstract. & text quality scores & \\
& &(E4-0) & (E4-5) Specify the score interval. & & \\
& & & & & \\

\hline

& & & & & \\
\textbf{Task 4 --} & Introduction, Conclusion & Critiques & 
& Internal/external evaluation of & Human evaluation of \\
\textbf{Content Reflection} & of one scientific article & (E5-0) & 
& LLM-generated critiques & LLM-generated critiques \\ 
& & & & & \\

\hline

\end{tabular}
}
\caption{Experimental setup. e.g., E1-0: Experiment 1-0. Across the five experiments, there are a panel of robustness checks concerning four aspects of the variation of the prompt for this one-round human-LLM interaction: (1) semantic robustness of the prompt (E3-1/2/3, 4-1/2/3), (2) richness of the prompt (E1-3, 2-3, 2-4, 4-4), (3) abundance of the data prompt (E1-1, 1-2, 2-1, 2-2), and (4) specificity of the instruction prompt (E1-4, 1-5, 2-5, 4-5).}
\label{tab2}
\end{table}
\end{landscape}

\newpage

\section{Analytical Tools}

\subsection{Metrics for text evaluation}

For content reproduction (Task 1) and content reflection (Task 4), we employ complementary metrics to evaluate the quality of LLM-generated texts from two angles: internal and external. For internal evaluation, we assess the linguistic quality of the text \citep{K2019}, employing four metrics: information density (via the Halliday lexical density \citep{H1985}), richness (via the Shannon information entropy \citep{SL2022} or the type-token ratio (TTR) \citep{R1987}), and readability (via the Flesch-Kincaid (FK) score \citep{KFet1975}). For external evaluation, we compare the LLM-generated keywords/abstracts/critiques to the ground truth (Task 1) or the text input (Task 4), employing four metrics to indicate text similarity: Jaccard index, cosine similarity (upon the Term Frequency-Inverse Document Frequency (TF-IDF) \citep{LPet2016}), bilingual evaluation understudy (BLEU) \citep{PRet2002}, and recall-oriented understudy for gisting evaluation (ROUGE) \citep{L2004}. The metrics for text evaluation are summarized in Table \ref{tab5}.

\begin{table}[h!]
\centering
\resizebox{\textwidth}{!}
{
\begin{tabular}{cc}

\multicolumn{2}{l}{\textbf{Internal Evaluation (IE) - linguistic quality of the text}} \\ \hline 
\hline
Information density & \multirow{2}{*}{$IE_{H-density} = \frac{number\ of\ lexical\ items}{number\ of\ clauses}$} \\
(via the Halliday lexical density) & \\ \hline
Richness & \multirow{2}{*}{$IE_{entropy} = - \sum^n_{i=1} p_i\log_2{p_i}$} \\
(via the Shannon information entropy) & \\ \hline
Richness & \multirow{2}{*}{$IE_{TTR} = \frac{number\ of\ unique\ words\ (types)}{total\ number\ of\ words\ (tokens)}$} \\
(via the type-token ratio (TTR)) & \\ \hline
Readability & \multirow{2}{*}{$IE_{FK} = 206.835 - 1.015*(\frac{number\ of\ words}{number\ of\ sentences}) - 84.6*(\frac{number\ of\ syllables}{number\ of\ words})$} \\
(via the Flesch-Kincaid (FK) score) & \\ \hline
\hline

& \\

\multicolumn{2}{l}{\textbf{External Evaluation (EE) - similarity to the ground truth}} \\ \hline 
\hline
\multirow{2}{*}{Jaccard index} & \multirow{2}{*}{$EE_{Jac} = \frac{|n_{original} \cap n_{generated}|}{|n_{original} \cup n_{generated}|} $} \\
& \\ \hline
\multirow{2}{*}{Cosine similarity} & \multirow{2}{*}{$EE_{cos} = \frac{\sum^n_{i=1}TFIDF_{original}*TFIDF_{generated}}{\sqrt{\sum^n_{i=1}TFIDF_{original}^2}\sqrt{\sum^n_{i=1}TFIDF_{generated}^2}}$} \\
& \\ \hline
\multirow{2}{*}{BLEU} & \multirow{2}{*}{$EE_{BLEU} = BP * exp(\sum^N_{n=1}w_n\log{p_n})$} \\
& \\ \hline
\multirow{2}{*}{ROUGE} & \multirow{2}{*}{$EE_{ROUGE} = \frac{(1+\beta^2)EE_{ROUGE_{L}-recall}EE_{ROUGE_{L}-precision}}{EE_{ROUGE_{L}-recall} + \beta^2 EE_{ROUGE_{L}-precision}}$} \\
& \\ \hline
\hline
\end{tabular}
}
\caption{Metrics for internal/external text evaluation.}
\label{tab5}
\end{table}

\subsubsection{Internal Evaluation}

\textbf{Information density (via the Halliday lexical density).} The information density of the text is measured using the Halliday lexical density, defined as the ratio of the number of lexical items to the number of clauses in the text. Following \citet{TFet2013}, we determine the number of lexical items/clauses based on the presence of finite verbs, i.e., the ratio of all words to verbs. The metric $IE_{H-density}$ is given by
\begin{equation}
IE_{H-density} = \frac{number\ of\ lexical\ items}{number\ of\ clauses}.
\end{equation}

\textbf{Richness (via the Shannon information entropy).} To evaluate the richness of the text, we compute the Shannon entropy based on the frequency of words in the text. High entropy indicates a rich text employing a large vocabulary. The metric $IE_{entropy}$ is given by
\begin{equation} 
IE_{entropy} = - \sum^n_{i=1} p_i\log_2{p_i},
\end{equation}
where $n$ is the number of unique words and $p_i$ is the probability of the occurrence of word $i$.

\textbf{Richness (via the type-token ratio (TTR)).} TTR measures the richness of the text by the ratio of unique words (``types'') to the total number of words (``tokens'') in the text. A high TTR indicates a rich vocabulary. The metric $IE_{TTR}$ is given by

\vspace{-0.2cm}

\begin{equation}
IE_{TTR} = \frac{number\ of\ unique\ words\ (types)}{total\ number\ of\ words\ (tokens)}.
\end{equation}

\vspace{-0.3cm}

\textbf{Readability (via the Flesch-Kincaid (FK) score).} The FK score indicates the level of difficulty in text understanding, widely used in educational settings, etc. The score is based on the number of syllables, words, and sentences in the text, with a high score denoting great readability. The metric $IE_{FK}$ is given by
\begin{equation}
IE_{FK} = 206.835 - 1.015*(\frac{number\ of\ words}{number\ of\ sentences}) - 84.6*(\frac{number\ of\ syllables}{number\ of\ words}).
\end{equation}
We use the toolkit \textit{syllapy} (https://github.com/mholtzscher/syllapy) to count text syllables.

\vspace{-0.2cm}

\subsubsection{External Evaluation}

\textbf{Jaccard index.} Jaccard index measures the similarity of two sets of elements. Here the elements are unique words in the text. The metric $EE_{Jac}$ is given by ($n$ is the number of unique words)
\begin{equation}
EE_{Jac} = \frac{|n_{original} \cap n_{generated}|}{|n_{original} \cup n_{generated}|},
\end{equation}

\textbf{Cosine similarity.} Cosine similarity measures the angle between two vectors in a multi-dimensional space. Here text vectors are represented with the TF-IDF, which is proportional to word frequency in the text while offset by word frequency in a background corpus (English dictionary), adjusting for common words. The metric $EE_{cos}$ is given by ($n$ is the number of unique words)
\begin{equation}
EE_{cos} = \frac{\sum^n_{i=1}TFIDF_{original}*TFIDF_{generated}}{\sqrt{\sum^n_{i=1}TFIDF_{original}^2}\sqrt{\sum^n_{i=1}TFIDF_{generated}^2}},
\end{equation}

\textbf{Bilingual Evaluation Understudy (BLEU).} BLEU evaluates machine-translated texts against reference translations. Here it calculates the precision of thewords in the LLM-generated text that appear in the ground-truth text, applying a penalty to discourage short texts. The metric $EE_{BLEU}$ is given by

\begin{equation}
EE_{BLEU} = BP * exp(\sum^N_{n=1}w_n\log{p_n}),
\end{equation}

where $p_n$ and $w_n$ are the precision and weight of each word; $BP = e^{-\max{(0, r/c-1)}}$ is penalizing short texts, with $c$/$r$ as the respective lengths of the LLM-generated/ground-truth text.

\textbf{Recall-Oriented Understudy for Gisting Evaluation (ROUGE).} ROUGE evaluates machine translations against references by measuring the match of words, word pairs, and word sequences between the two texts. Here we use ROUGE-L, which calculates the F-measure of the longest common subsequences (LCSs) between the two texts. We construct the metric $EE_{ROUGE}$, given by 
\begin{equation}
\begin{aligned}
EE_{ROUGE_{L}-recall} &= \frac{LCS(reference\ text,\ generated\ text)}{\sum_{words \in {reference\ text}}Count(words)},\\
EE_{ROUGE_{L}-precision} &= \frac{LCS (reference\ text,\ generated\ text)}{\sum_{words \in {generated\ text}}Count(words)},\\
EE_{ROUGE} &= \frac{(1+\beta^2)EE_{ROUGE_{L}-recall}EE_{ROUGE_{L}-precision}}{EE_{ROUGE_{L}-recall} + \beta^2 EE_{ROUGE_{L}-precision}},
\end{aligned}
\end{equation}

where $\beta$ is the weight for recall relative to precision (set to 1 in our calculation to give equal importance to precision and recall).

\subsection{Text ranking from quality preferences}

For content comparison (Task 2), we collect the LLM's pairwise preferences to construct a ranking over the text ensemble. Ranking objects based on pairwise comparisons is an important problem \citep{WJet2013} with various algorithmic solutions \citep[e.g.,][]{JN2011,NOet2012}. We use the classic Copeland counting \citep{C1951,SW2018} to construct the ranking. For two objects $i$ and $j$, $y_{ij} \in [-1,0,1]$ is the ground-truth outcome of their comparison: $j$ beating $i$ ($y_{ij} = -1$), dual ($y_{ij} = 0$), or $i$ beating $j$ ($y_{ij} = 1$); the sum of an object's scores from $N-1$ comparisons is its Copeland score. For error-free comparisons, the resulting Copeland scores for the compared items should form the perfect sequence $N-1$, $N-3$, $N-5$, ..., $-(N-3)$, $-(N-1)$.

Notably, due to the stochasticity of the LLM, it is possible that the outcome of comparing $i$ to $j$, $\hat{y_{ij}}$, differs from the (opposite) outcome of comparing $j$ to $i$, $-\hat{y_{ji}}$. We average the two outcomes, $\hat{z_{ij}} = 9\hat{y_{ij}}+(-\hat{y_{ji}}))/2$, and obtain the matrix $\mathbf{Z} = \{\hat{z_{ij}}\}$, which is asymmetric and deviates from the symmetric ground-truth outcome matrix $\mathbf{Y} = \{y_{ij}\}$. The distance between the measured Copeland scores (computed on $\mathbf{Z}$) and the perfect sequence (computed on $\mathbf{Y}$) can help estimate the error rate of the LLM output (Section 5.2).

\subsection{Assessing text quality scores}

For content scoring (Task 3), we assess the discrimination of the LLM-output text scores $\hat{w}_i$. We indicate the spread of $\hat{w}_i$ over the $N$ texts using mean/median/mode, std, and percentiles. We demonstrate the sharpness \citep{GBet2007} of the score distribution using skewness and kurtosis: $skewness = m_3/m_2^{3/2}$, $kurtosis = m_4/m_2^2 - 3$, where $m_p$ is the $p$-th moment of the score distribution:
\vspace{-0.2cm}
\begin{equation}
   m_p = \frac{1}{N} \sum_{i=1}^{N} (\hat{w}_i - \hat{w}_{ave})^p.
\end{equation}

We consider shrinkage bias in the LLM-output scores which narrows the default score range [1, 10] to an internal range [$H_l$, $H_h$]. Under this bias, we discuss the condition for LLM-output scores to distinguish ground-truth score differences (Section 5.3).
\vspace{-0.1cm}
\subsection{Human evaluation of LLM-generated texts}

We recruited 87 graduate students (including 1 undergraduate and 1 unreported) from business schools and engineering schools as paid human arbiters to evaluate LLM-generated texts; recruited arbiters' English reading skills are endorsed by sufficient language test scores (Appendix B). At content reproduction (Task 1), arbiters compare the LLM-generated abstract to the ground-truth abstract and submit an integer score $HE^{abs}$ from 1 (least) to 5 (most) to indicate the LLM output's reliability. At content reflection (Task 4), arbiters read the input article sections and the LLM-generated critiques and submit an integer score $HE^{cri}$ from 1 (least) to 5 (most) to indicate the LLM output's insightfulness. 

We follow the guidelines of prior studies on asking human arbiters to evaluate AI-generated content (texts or images) \citep[e.g.,][]{NF2022,HUet2023,MSet2023}. The Institutional Review Board of the University approved the study protocols. The evaluations were performed in accordance with all relevant guidelines and regulations; informed consent was obtained from all participants. The collected data was stored securely and did not include any direct information that could reveal participants' identities or any sensitive information that could link to participants' personal records.

We developed an online interface for this crowdsourced evaluation (Appendix C). Arbiters were encouraged to complete the assignment remotely through the web page. After reading a short instruction, each arbiter was randomly assigned 12 abstract pairs and 8 article-critique pairs, displayed in a sequence; we thus gathered six samples for each $HE^{abs}$ and four samples for each $HE^{cri}$. The time limit was 3 minutes for abstract comparison and 8 minutes for critique assessment. We recorded arbiters' judgment time and discarded the samples from very fast judgment, i.e., $<$15 seconds at abstract comparison or $<$60 seconds at critique assessment. There were multiple breaks between evaluation sessions; each arbiter devoted a maximum of 110 minutes (including a maximum of 100 minutes for the evaluation) to the assignment. 

\newpage

\section{Materials}

\subsection{Academic Texts}

We collect articles published in \textit{Information Systems Research} (ISR) from Dec. 2022 to Mar. 2024, articles published in \textit{Journal of Management Information Systems} (JMIS) from Apr. 2022 to Jun. 2024, and articles published in the IS department of \textit{Management Sciences} (MS) in 2022 and 2023. The dataset consists of 246 articles, including 84 regular articles (RA), 19 research notes (RN), and 13 special issue articles (SI) from ISR, 81 articles (JM) from JMIS, and 49 articles (MS) from MS (Table \ref{tab3}). We compile the keywords, abstract, main text, and metadata (for ISR/MS, online publication date, received/accepted date, and download count\footnote{https://pubsonline.informs.org/}; for JMIS, online publication date, view count, and CrossRef citation count\footnote{https://www.tandfonline.com/journals/mmis20}) of each article. An article's acceptance time is from the received date to the accepted date; the download, view, and citation counts are normalized by the elapsed time (from the online publication date to the current date).

\begin{table}[h!]
\centering
\resizebox{0.55\textwidth}{!}
{
\begin{tabular}{cccccc}
 
\hline
\textbf{Article Type} & \textbf{RA} & \textbf{RN} & \textbf{SI} & \textbf{JM} & \textbf{MS} \\ \hline \hline
{No. Articles} & 84 & 19 & 13 & 81 & 49\\ \hline
{Ave. No. Keywords} & 5.9 & 6.2 & 5.4 & 7.3 & 5.0 \\
{Ave. Len. Abstract (words)} & 268 & 274 & 247 & 221 & 247 \\
{Ave. Len. Main Text (words)} & 12486 & 9885 & 11911 & 11776 & 11284 \\
{Ave. Acceptance Time (days)} & 835.5 & 824.7 & 628.6 & - & 754 \\
{Ave. Norm. Download (times)} & 3.4 & 3.0 & 1.1 & - & 1.38 \\
{Ave. Norm. View (times)} & - & - & - & 1351.7 & - \\
{Ave. Norm. Citation (times)} & - & - & - & 0.9 & - \\
\hline
\end{tabular}
}
\caption{Information on article data. No.: number of. Ave.: average. Len.: length of. Norm.: normalized.}
\label{tab3}
\end{table}

\subsection{Large Language Model}

LLMs have seen rapid development in recent years; leading tech corporations compete in launching and upgrading their LLMs that adopt an open- or closed-source design (see \citet{L2024} for a comparison of several major open-/closed-source LLMs). Users evaluate LLMs' performance on the chatbot arena leaderboard (https://chat.lmsys.org/leaderboard) supported by leading research institutions. On this crowdsourced leaderboard, participants are presented with text responses from two anonymous LLMs to each query they input; the dialogue proceeds until one discerns that an LLM yields the superior response, at which point the participant casts a vote. As of Aug 2025, over 3,820,000 effective votes have been collected. 

In this study, we adopt Google's Gemini, which ranks high in the LLM leaderboard and is reported to perform well in academic writing tasks \citep{G2024}. Google applies strict training restrictions to ensure that no LLM input is used for model training or fine-tuning; the sole purpose of logging prompt data is to detect potential misuse or policy violations\footnote{https://cloud.google.com/vertex-ai/generative-ai/docs/data-governance}. The academic content that we use in this study is thus secure and does not spill beyond the scope of our evaluation, precluding copyright concerns. We use the stable model versions Gemini Pro 1.0 and 1.5 (1.0 discontinued on Apr. 2025) which support text input/output and are optimized for natural language tasks. The application parameters of Gemini models include \textit{input/output token limit} (maximum number of tokens\footnote{One token is approximately four English characters; 100 tokens correspond to 60-80 English words.} allowed in the prompt/response), \textit{temperature} (degree of randomness in token selection; low temperatures corresponding to more deterministic responses), \textit{topK} (the next token is to be selected from how many candidates), and \textit{topP} (selecting tokens from the most to the least probable until the probability sum equals this value). We use the default configuration in our experiments; temperature is set at 0.9 to encourage stochastic output (Table \ref{tab4}). Under each prompt, we conduct five runs and build results over the output ensemble. We employ zero-shot prompts throughout our interaction with the LLM.
 
\begin{table}[H]
\centering
\resizebox{0.85\textwidth}{!}
{
\begin{tabular}{c|ccccc}
 
\hline
& \textbf{input token limit} & \textbf{output token limit} & \textbf{temperature} & \textbf{topK} & \textbf{topP} \\ \hline 
\textbf{Google's Gemini (Pro 1.0/1.5)} & 30720 & 2048 & 0.9 & None & 1.0\\
\hline
\end{tabular}
}
\caption{Parameters of the LLM (Google's Gemini) used in this study.}
\label{tab4}
\end{table}

\newpage

\section{Results}


\subsection{\textit{Reliability} of content reproduction}

We report the statistics of word use of ground-truth/LLM-generated keywords and abstracts at Task 1 (Table \ref{tab6}). Different forms of the same word are grouped. We compare the top-30 frequent words.

\begin{table}[h!]
\centering
\resizebox{0.8\textwidth}{!}
{
\begin{tabular}{ccccc}
\hline
\multirow{2}{*}{} & \multicolumn{2}{c}{\textbf{Keyword Reproduction}} & \multicolumn{2}{c}{\textbf{Abstract Reproduction}} \\ \cline{2-5}
 & \textbf{Ground-truth} & \textbf{LLM-gen.} & \textbf{Ground-truth} & \textbf{LLM-gen.} \\ \hline \hline
\textbf{No. of unique words} & 3053 & 6308 & 35443 & 25676 \\ \hline
\textbf{Gini index} & 0.54 & 0.70 & 0.68 & 0.76 \\ \hline
\textbf{\textit{top-30 frequent words}} & \multicolumn{4}{c}{} \\ \hline
1 & online$^*$  & information$^*$  & platform$^*$  & study \\ \hline
2 & platform$^*$  & platform$^*$  & effect$^*$  & platform$^*$  \\ \hline
3 & information$^*$ & system$^*$ & user  & effect$^*$  \\ \hline
4 & social$^*$  & social$^*$  & study & impact \\ \hline
5 & learning & market$^*$ & information$^*$  & finding \\ \hline
6 & theory  & effect$^*$ & data$^*$ & user  \\ \hline
7 & market$^*$  & online$^*$  & firm  & information$^*$ \\ \hline
8 & design  & theory & social$^*$  & data$^*$  \\ \hline
9 & data$^*$ & learning & model$^*$ & market$^*$  \\ \hline
10 & system$^*$  & data$^*$  & consumer  & model$^*$  \\ \hline
11 & network & network & online$^*$ & social$^*$ \\ \hline
12 & effect$^*$  & technology  & result & research  \\ \hline
13 & experiment & model$^*$ & market$^*$  & online$^*$ \\ \hline
14 & digital  & behavior  & research  & firm \\ \hline
15 & analysis  & design & find  & using  \\ \hline
16 & medium  & digital & based & consumer  \\ \hline
17 & model$^*$ & analysis  & system$^*$ & system$^*$  \\ \hline
18 & technology & health & show & service$^*$ \\ \hline
19 & health & cost & impact  & performance  \\ \hline
20 & field  & user  & finding  & examine \\ \hline
21 & analytics  & medium  & service$^*$  & based \\ \hline
22 & behavior  & performance & use  & cost \\ \hline
23 & difference & consumer  & technology & influence \\ \hline
24 & mobile  & customer  & value & investigate \\ \hline
25 & deep  & decision & mechanism & strategy  \\ \hline
26 & intelligence  & privacy & increase  & behavior  \\ \hline
27 & security & service$^*$  & performance & AI  \\ \hline
28 & healthcare  & AI & design  & highlight \\ \hline
29 & service$^*$  & healthcare & customer  & decision  \\ \hline
30 & value  & value  & cost & analysis \\ \hline
\textbf{Shared elements} & \multicolumn{4}{c}{10/30}  \\ \hline
\textbf{Unique elements} & 9/30 & 1/30 & 7/30 & 6/30 \\ \hline
\end{tabular}
}
\caption{Content reproduction (Task 1) -- Statistics of word use in ground-truth/LLM-generated (LLM-gen.) keywords and abstracts. No.: Number. $^*$: words shared in the four lists.}
\label{tab6}
\end{table}

At keywords reproduction, the LLM output employs a clearly larger vocabulary than the ground truth (6308 vs. 3053), yet much of the vocabulary is less frequently used, manifested in the higher Gini index (0.70 vs. 0.54) and fewer unique elements (1 vs. 9) in the top-30 word list. This suggests a more concentrated word use of LLM-generated texts. The word use of ground-truth/LLM-generated abstracts is similar.

\begin{figure}[h!]
    \centering
    \includegraphics[width=0.8\linewidth]{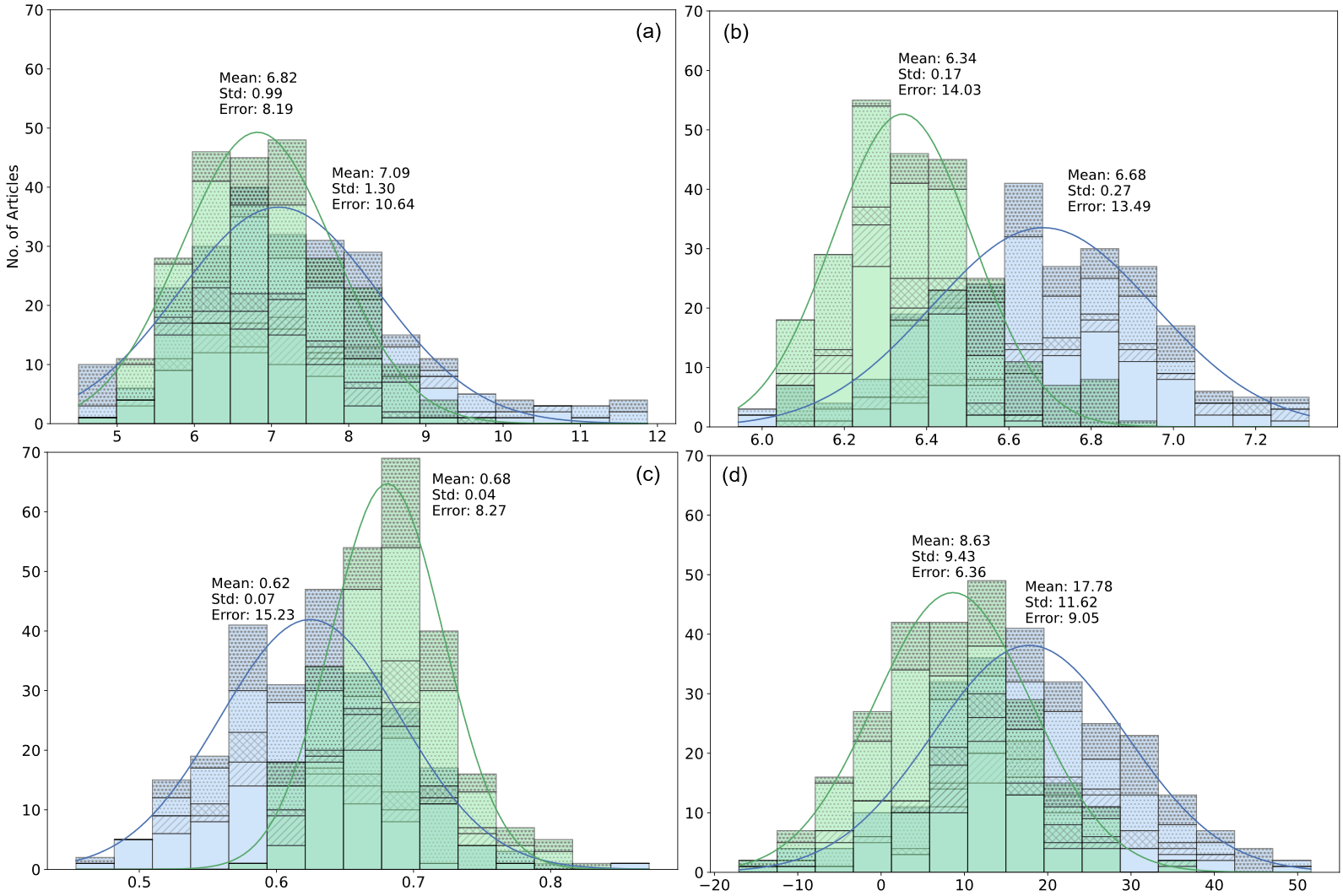}
    \caption{Content reproduction (Task 1) -- Internal evaluation of ground-truth (blue)/LLM-generated (green) abstracts for different article types (RA [empty], RN [slash], SI [cross], JM [dot], MS [star]) under metrics $IE_{H-density/entropy/TTR/FK}$ (a/b/c/d; x-axis: scores). Solid lines: best-fit Gaussian distributions.}
    \label{fig3}
\end{figure}

\textbf{Internal evaluation of abstracts.} We show internal evaluation results of ground-truth/LLM-generated abstracts for the five article types (RA, RN, SI, JM, MS). For each article, we evaluated the linguistic quality of the two abstracts under the four metrics $IE_{H-density/entropy/TTR/FK}$ (Figure \ref{fig3} and Table \ref{tab7}). 

On the frequency plots (Figure \ref{fig3}) of $IE_{\bullet}$, both ground-truth and LLM-generated abstracts exhibit approximate Gaussian distributions yet with different means and variances. LLM-generated abstracts have a lower mean and smaller variance under $IE_{H-density}$, $IE_{entropy}$, and $IE_{FK}$, suggesting more significant text homogeneity. LLM-generated abstracts obtain a higher mean score than ground-truth abstracts at $IE_{TTR}$, revealing a larger vocabulary use in the LLM output. We conduct the paired t-test on the abstract pairs (Table 6a) and find these differences in $IE_{\bullet}$ significant across the article set and also within the five subsets, except for a few cases. We conduct a one-way ANOVA on the five article types; in many cases (Table 6c), results suggest a significant difference across article types for both ground-truth and LLM-generated texts.

\begin{figure}[h!]
\centering
\includegraphics[width=0.8\linewidth]{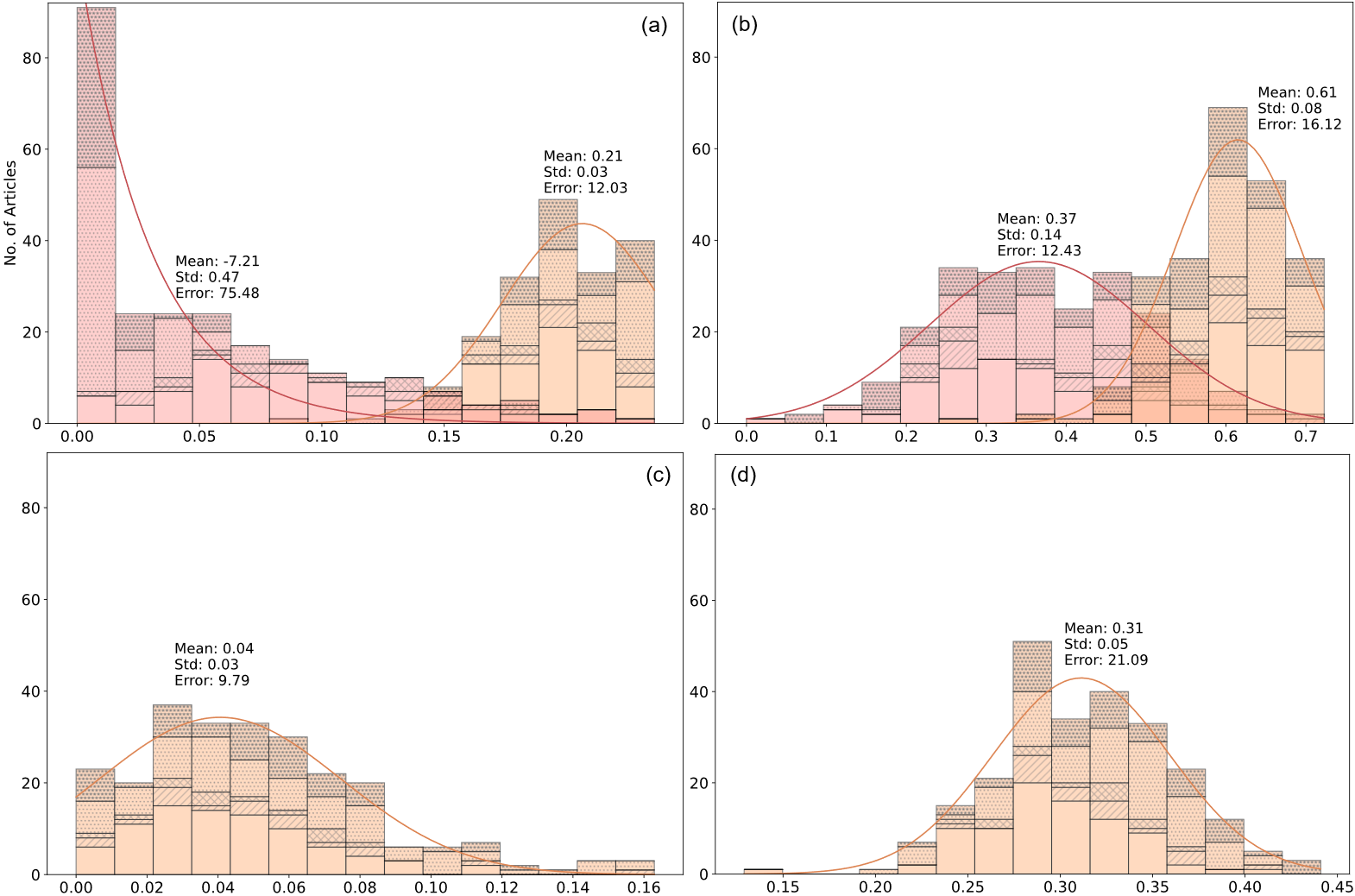}
\caption{Content reproduction (Task 1) -- External evaluation results comparing the ground-truth/LLM-generated keywords (pink) and abstracts (orange) for different article types (RA [empty], RN [slash], SI [cross], JM [dot], MS [star]) under metrics $EE_{Jac/cos}$ (keywords) (a/b) and $EE_{Jac/cos/BLEU/ROUGE}$ (abstracts) (a/b/c/d; x-axis: scores). Solid lines: best-fit Gaussian distributions.}
\label{fig4}
\end{figure}

\textbf{External evaluation of keywords/abstracts.} We show external evaluation results comparing ground-truth/LLM-generated keywords and abstracts. For each article, the similarity of the two keyword lists/abstracts is evaluated under $EE_{Jac/cos/BLEU/ROUGE}$ (Figure \ref{fig4}). $EE_{BLEU/ROUGE}$ does not apply to the comparison of the keyword lists due to the absence of sentences.

Similar to $IE_{\bullet}$, the measured $EE_{\bullet}$ displays approximate Gaussian distributions but now with greater deviations. Except for at $EE_{Jac}$ comparing ground-truth/LLM-generated keywords, where the distribution is highly right-skewed and fitting a symmetric normal distribution thus results in a negative fitted mean. For the LLM, the keyword output obtains smaller similarity scores ($EE_{Jac/cos}$) than the abstract output (Table 6b), as it is more challenging to reproduce the exact ground-truth keywords than to reproduce a similar abstract. Across five article types, one-way ANOVA results (Table 6c) reveal a significant difference in the Jaccard similarity at both keywords and the abstract, and in the ROUGE score at the abstract.

\textbf{Human evaluation of abstracts.} We show human evaluation results of LLM-generated abstracts. For each article, the reliability of the LLM-generated abstract with respect to the ground truth is indicated by the average score $HE^{abs}\in [1,5]$ from human arbiters. 

\begin{figure}[h!]
    \centering
    \includegraphics[width=\linewidth]{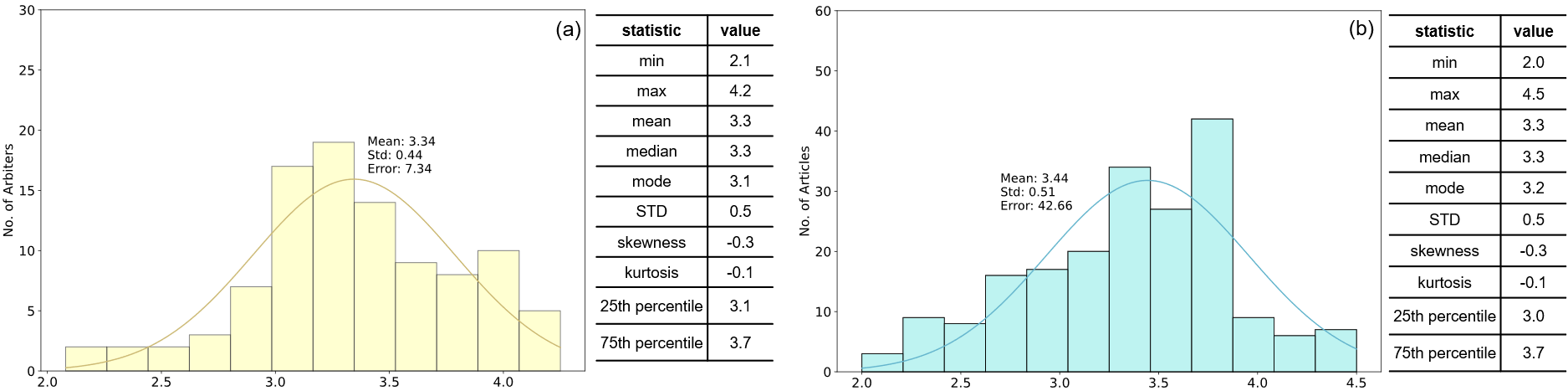}
    \caption{Content reproduction (Task 1) -- Human evaluation results of professional arbiters grading LLM-generated abstracts. (a) Distribution and statistics of arbiters' mean scores given to the assigned abstracts. (b) Distribution and statistics of average scores $HE^{abs}$ for LLM-generated abstracts. Solid lines: best-fit Gaussian distributions.}
    \label{fig5}
\end{figure}

The arbiters' mean score given to the 12 assigned abstracts skewed upwards (Figure \ref{fig5}a), the lowest mean score is 2.1 and the average is 3.3. Collecting the six (five in a few cases) evaluations from arbiters, abstracts' mean score $HE^{abs}$ is distributed between 2.0 and 4.5, averaging at 3.3 and mostly concentrated within [3, 4] (Figure \ref{fig5}b). These scores suggest an acceptable acknowledgment of LLM-generated abstracts by professional human arbiters. $HE^{abs}$ does not show a significant difference across article types (Table 6c).

\begin{table}[H]
\centering

\begin{subtable}[t]{0.9\textwidth}
    \centering
    \caption{Paired t-test between the ground-truth and LLM-generated abstracts (internal evaluation)}
    \resizebox{0.85\textwidth}{!}
{   \begin{tabular}{lccc|cc|cc|ccc}
    \hline
     & \multicolumn{1}{c}{Count} & \multicolumn{4}{c}{\textbf{$IE_{H-density}$}} & \multicolumn{4}{c}{\textbf{$IE_{entropy}$}} \\
    Type & & \multicolumn{2}{c}{ground-truth} & \multicolumn{2}{c}{LLM-generated} & \multicolumn{2}{c}{ground-truth} & \multicolumn{2}{c}{LLM-generated} \\
    \hline
     & & Mean & Std & Mean & Std & Mean & Std & Mean & Std \\
    \hline
    RA & 84 & 7.374 & 1.368 & \textbf{6.616}*** & 0.898 & 6.716 & 0.253 & \textbf{6.272}*** & 0.422 \\
    RN & 19 & 7.592 & 1.376 & 7.134~~~ & 0.753 & 6.743 & 0.342 & \textbf{6.295}*** & 0.142 \\
    SI & 13 & 6.858 & 0.905 & 7.068~~~ & 1.128 & 6.626 & 0.333 & \textbf{6.359}**~ & 0.109 \\
    JM & 81 & 7.368 & 1.415 & \textbf{6.783}*** & 0.898 & 6.599 & 0.223 & \textbf{6.310}*** & 0.134 \\
    MS & 49 & 7.239 & 1.685 & 7.452 & 0.927 & 6.669 & 0.319 & 6.609 & 0.153 \\
    All & 246 & 7.334 & 1.430 & \textbf{6.902}*** & 0.953 & 6.665 & 0.273 & \textbf{6.358}*** & 0.299 \\
    \hline
    \hline
     & \multicolumn{1}{c}{Count} & \multicolumn{4}{c}{\textbf{$IE_{TTR}$}} & \multicolumn{4}{c}{\textbf{$IE_{FK}$}} \\
    Type & & \multicolumn{2}{c}{ground-truth} & \multicolumn{2}{c}{LLM-generated} & \multicolumn{2}{c}{ground-truth} & \multicolumn{2}{c}{LLM-generated} \\
    \hline
     & & Mean & Std & Mean & Std & Mean & Std & Mean & Std \\
    \hline
    RA & 84 & 0.608 & 0.066 & \textbf{0.673}*** & 0.036 & 16.709 & 11.524 & \textbf{9.980}*** & 9.541 \\
    RN & 19 & 0.604 & 0.054 & \textbf{0.663}*** & 0.045 & 17.445 & 10.091 & \textbf{10.288}*** & 9.968 \\
    SI & 13 & 0.617 & 0.053 & \textbf{0.681}*** & 0.036 & 21.412 & 9.274 & \textbf{15.353}** & 5.711 \\
    JM & 81 & 0.646 & 0.059 & \textbf{0.689}*** & 0.044 & 14.733 & 12.306 & \textbf{5.567}*** & 9.541 \\
    MS & 49 & 0.629 & 0.064 & \textbf{0.699}*** & 0.048 & 24.855 & 11.769 & \textbf{8.035}*** & 10.459 \\
    All & 246 & 0.625 & 0.064 & \textbf{0.684}*** & 0.045 & 17.987 & 12.141 & \textbf{8.447}*** & 10.512 \\
    \hline
    \end{tabular}
    }
\end{subtable}

\begin{subtable}[t]{0.9\textwidth}
    \centering
    \caption{Comparing ground-truth/LLM-generated keywords and abstracts (external evaluation)}
    \resizebox{0.85\textwidth}{!}
{   \begin{tabular}{lccc|cc|cc|ccc}
    \hline
     & \multicolumn{1}{c}{Count} & \multicolumn{4}{c}{\textbf{$EE_{Jac}$}} & \multicolumn{4}{c}{\textbf{$EE_{cos}$}} \\
    Type & & \multicolumn{2}{c}{keywords} & \multicolumn{2}{c}{abstract} & \multicolumn{2}{c}{keywords} & \multicolumn{2}{c}{abstract} \\
    \hline
     & & Mean & Std & Mean & Std & Mean & Std & Mean & Std \\
    \hline
    RA & 84 & 0.089 & 0.054 & 0.197 & 0.030 & 0.357 & 0.134 & 0.605 & 0.082 \\
    RN & 19 & 0.099 & 0.053 & 0.216 & 0.033 & 0.400 & 0.134 & 0.634 & 0.060 \\
    SI & 13 & 0.094 & 0.051 & 0.205 & 0.026 & 0.364 & 0.125 & 0.590 & 0.061 \\
    JM & 81 & 0.019 & 0.029 & 0.220 & 0.036 & 0.381 & 0.127 & 0.600 & 0.076 \\
    MS & 49 & 0.011 & 0.021 & 0.221 & 0.038 & 0.332 & 0.121 & 0.595 & 0.071 \\
    All & 246 & 0.052 & 0.056 & 0.211 & 0.035 & 0.372 & 0.127 & 0.603 & 0.075 \\
    \hline
    \hline
     & \multicolumn{1}{c}{Count} & \multicolumn{4}{c}{\textbf{$EE_{BLEU}$}} & \multicolumn{4}{c}{\textbf{$EE_{ROUGE}$}} \\
    Type & & \multicolumn{2}{c}{keywords} & \multicolumn{2}{c}{abstract} & \multicolumn{2}{c}{keywords} & \multicolumn{2}{c}{abstract} \\
    \hline
     & & Mean & Std & Mean & Std & Mean & Std & Mean & Std \\
    \hline
    RA & 84 & - & - & 0.043 & 0.025 & - & - & 0.296 & 0.042 \\
    RN & 19 & - & - & 0.054 & 0.039 & - & - & 0.318 & 0.040 \\
    SI & 13 & - & - & 0.046 & 0.023 & - & - & 0.309 & 0.039 \\
    JM & 81 & - & - & 0.060 & 0.040 & - & - & 0.324 & 0.048 \\
    MS & 49 & - & - & 0.051 & 0.030 & - & - & 0.329 & 0.049 \\
    All & 246 & - & - & 0.051 & 0.033 & - & - & 0.314 & 0.047 \\
    \hline
    \end{tabular}
    }
\end{subtable}


\begin{subtable}[t]{0.9\textwidth}
    \centering
    \caption{One-Way ANOVA results on different article types (internal/external/human evaluation)}
    \resizebox{0.85\textwidth}{!}
{   \begin{tabular}{lccc|ccc}
    \hline
    Measure ($IE_{\bullet}$) & \multicolumn{3}{c|}{ground-truth abstract} & \multicolumn{3}{c}{LLM-generated abstract} \\
     & sum\_sq & F & p-value & sum\_sq & F & p-value \\
    \hline
    \textbf{$IE_{H-density}$} & 4.879 & 0.593 & 0.668 & 24.227 & 7.368 & \textbf{\textless0.001}*** \\
    \hline
    \textbf{$IE_{entropy}$} & 0.708 & 2.426 & \textbf{0.048}* & 3.968 & 13.347 & \textbf{\textless0.001}*** \\
    \hline
    \textbf{$IE_{TTR}$} & 0.069 & 4.450 & \textbf{0.001}*** & 0.026 & 3.285 & \textbf{0.012}* \\
    \hline
    \textbf{$IE_{FK}$} & 3464.762 & 6.393 & \textbf{\textless0.001}*** & 1561.978 & 3.689 & \textbf{0.006}** \\
    \hline
    \hline
    Measure ($EE_{\bullet}$) & \multicolumn{3}{c|}{keywords} & \multicolumn{3}{c}{abstract} \\
     & sum\_sq & F & p-value & sum\_sq & F & p-value \\
    \hline
    \textbf{$EE_{Jac}$} & 0.352 & 52.01 & \textbf{\textless0.001}*** & 0.030 & 6.469 & \textbf{\textless0.001}***  \\
    \hline
    \textbf{$EE_{cos}$} & 0.206 & 3.291 & \textbf{0.012}* & 0.024 & 1.069 & 0.372 \\
    \hline
    \textbf{$EE_{BLEU}$} & - & - & - & 0.012 & 2.902 & \textbf{0.023}* \\
    \hline
    \textbf{$EE_{ROUGE}$} & - & - & - & 0.047 & 5.776 & \textbf{\textless0.001}*** \\
    \hline
    \hline
     Measure ($HE^{abs}$) & \multicolumn{3}{c|}{abstract} &  \\
     & sum\_sq & F & p-value &  \\
    \hline
    \textbf{$HE^{abs}$} & 1.920 & 2.580 & 0.054 & & &  \\
    \hline
    \end{tabular}
    }
\end{subtable}
\caption{Content reproduction (Task 1) – Statistics and inference analyses. $^*/^{**}/^{***}$: $p<0.05/0.01/0.001$.}
\label{tab7}
\end{table}

\textbf{Summary on \textit{reliability}.} Overall, our results from internal, external, and human evaluations of LLM-generated keywords and abstracts suggest that LLMs' ability to summarize and paraphrase academic texts is acceptably reliable, although in some cases discouraging. The limitation may arise from its hallucinations \citep[e.g.,][]{JLet2023}. That said, the LLM can reasonably play the role of \textit{oracle}, and its application potential is admissible at the level of literature digest.

\newpage

\subsection{\textit{Scalability} of content comparison}

\begin{figure}[h!]
    \centering
    \includegraphics[width=0.7\linewidth]{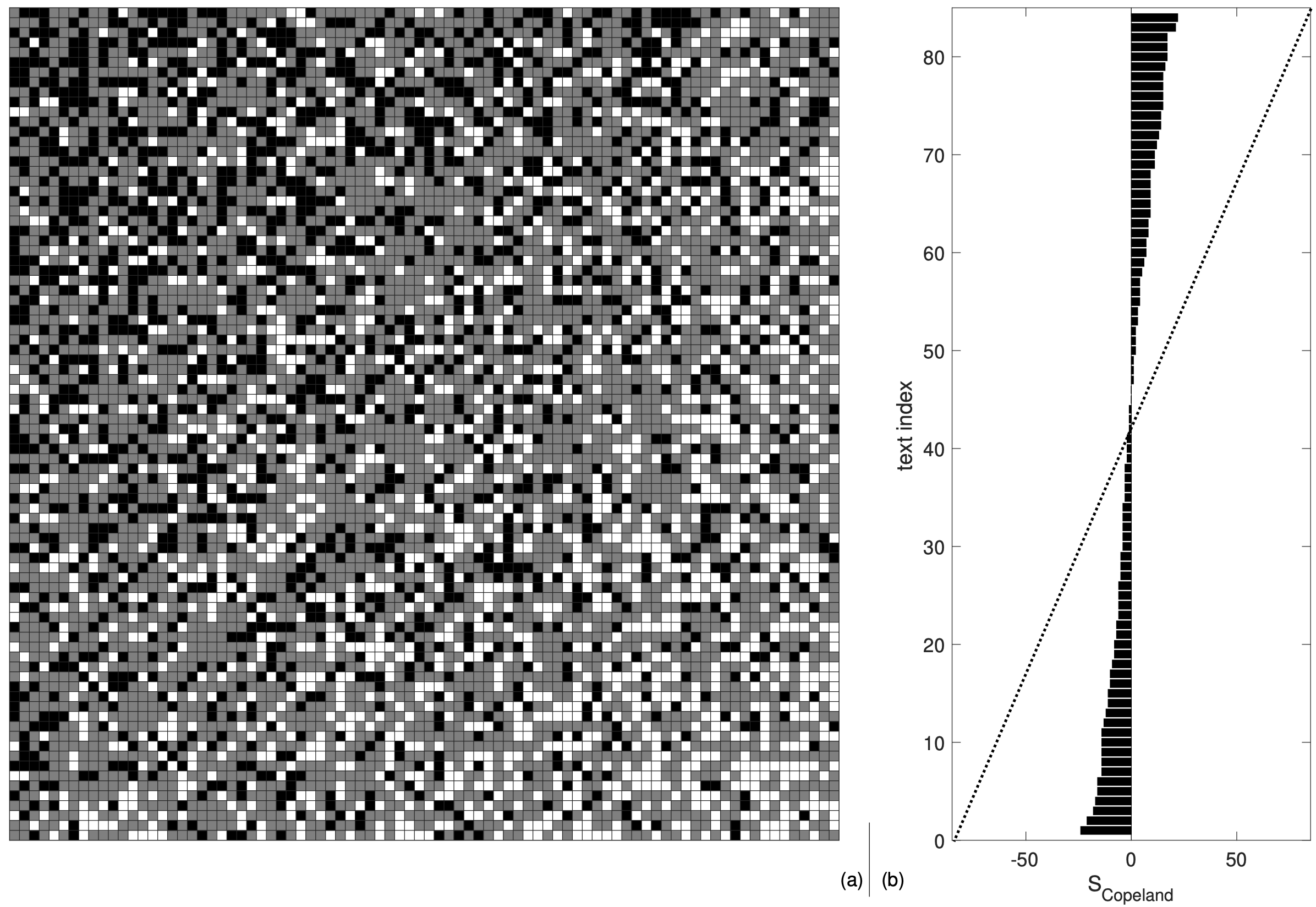}
    \caption{Content comparison (Task 2) - Demonstrating the RA case. (a) Preference matrix $\mathbf{Z}$ averaging two pairwise comparisons with either text in the front of the prompt. Entry $-1$ (black) or $1$ (white) indicates the row text beating or beaten by the column text twice in the comparison; $0$ (gray) indicates a tie. (b) Copeland scores (summing the row entries of $\mathbf{Z}$). Dotted line: perfect scores. Row indices in (a) correspond to the text indices in (b).}
    \label{fig6}
\end{figure}

We report the results of pairwise content comparison at Task 2. We compare the texts within each type (RA, RN, SI, JM, MS) for computational feasibility (due to the quadratic pairs); tests suggest that text comparisons across types bring similar results. A key feature of the results is that the LLM-output preference when asked to compare article $i$ to $j$, $\hat{y}_{ij}$, does not always agree with when asked to compare $j$ to $i$, $\hat{y}_{ji}$. Thus, for each article pair, we conduct two comparisons with each article in the front of the prompt and average the two outcomes, obtaining $\hat{z}_{ij} = (\hat{y}_{ij}+(-\hat{y}_{ji}))/2$. On the preference matrix $\mathbf{Z}=\{\hat{z}_{ij}\}$ (Figure \ref{fig6}a), a value of $-1$ (black) or $1$ (white) then indicates that the row article beats or is beaten by the column article twice in the comparison, and a value of $0$ (gray) indicates that the row and the column article each wins one comparison. $\mathbf{Z}$ is not diagonally symmetric but has the -1 elements mirroring the 1 elements. 

\begin{figure}[H]
    \centering
    \includegraphics[width=0.85\linewidth]{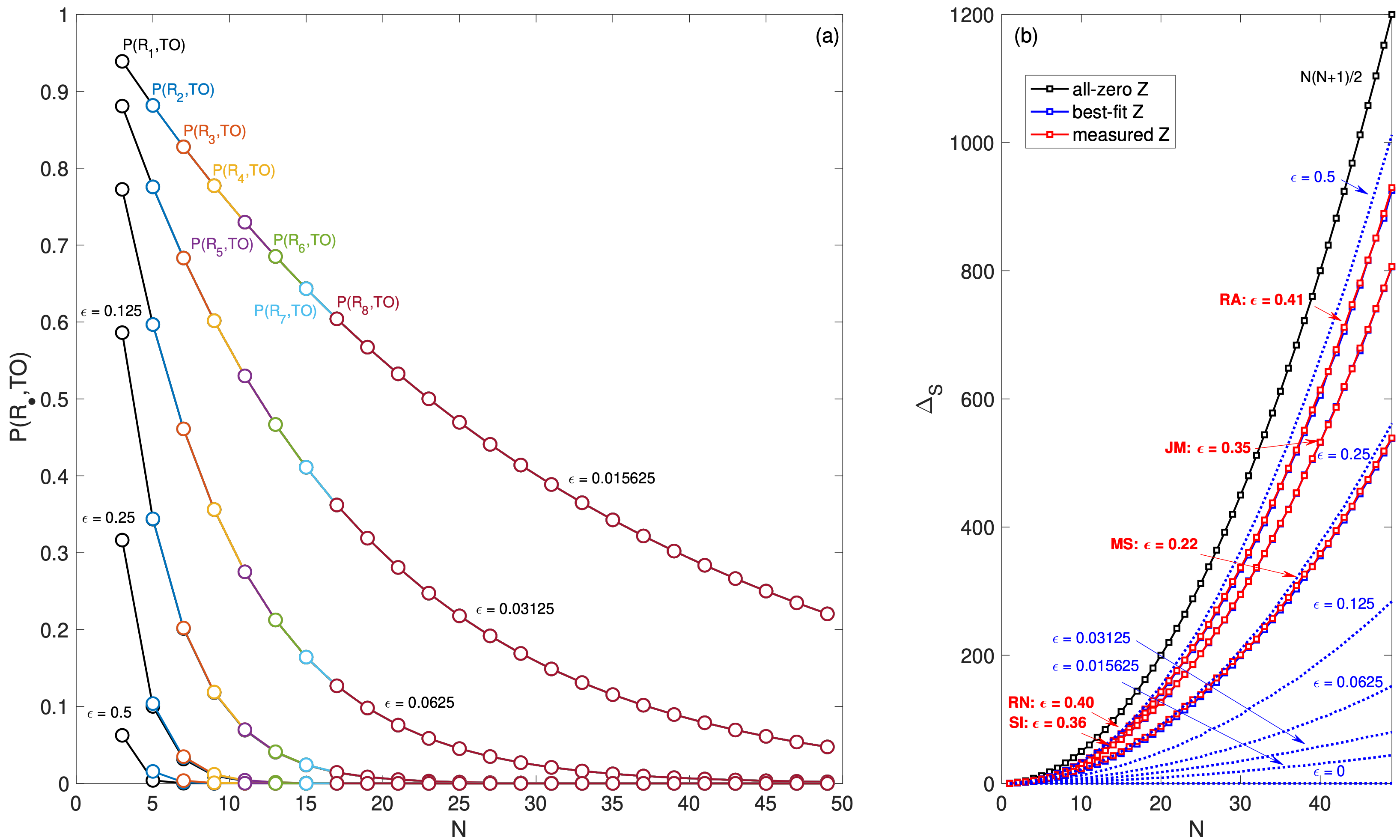}
    \caption{Content comparison (Task 2). (a) Probabilities of obtaining true $S_{Copeland}$ from content comparisons (six $\epsilon$ levels). As the number of texts $N$ increases, such probabilities uniformly decrease for all $\epsilon$ and the text of any ground-truth ranking (see Appendix D for $P(R_{\bullet},\text{TO})$). (b) Quantifying the loss of scalability with the deviation of $S_{Copeland}$ (denoted as $\Delta_S$) from the perfect sequence. Output error $\epsilon$ is estimated.}
    \label{fig7}
\end{figure}

We sum the row entries of $\mathbf{Z}$ to obtain the Copeland score for the row article; the articles' ranking is shown in Figure \ref{fig6}b. The obtained Copeland scores deviate from the (unknown) ground-truth scores: if there is no error in comparison outcomes, articles' Copeland scores should form the perfect sequence $N-1$, $N-3$, $N-5$, ..., $-(N-3)$, $-(N-1)$. The imperfect Copeland scores result from errors in the LLM-output preferences. We analyze this error as follows. Suppose the ground-truth outcome of comparing text $i$ and $j$ is $y_{ij}$. The LLM-output preference $\hat{y}_{ij}$ contains error with probability $\epsilon$:
\begin{equation}
\hat{y}_{ij} = \left\{
\begin{aligned}
&y_{ij}, \ \ P = 1-\epsilon, \\
-&y_{ij}, \ \ P = \epsilon. \\
\end{aligned}
\right.
\end{equation}
$\hat{z}_{ij}$ averages the outcomes $\hat{y}_{ij}$ and $\hat{y}_{ji}$ from the two instances of comparison between $i$ and $j$. Suppose the ground truth is $y_{ij}=-1$. There are four cases for the outcome at $\hat{z}_{ij}$: (i) $\hat{y}_{ij}$ is false and $\hat{y}_{ji}$ is true, $\hat{z}_{ij} = 0$; (ii) $\hat{y}_{ij}$ and $\hat{y}_{ji}$ are both true, $\hat{z}_{ij} = -1$; (iii) $\hat{y}_{ij}$ and $\hat{y}_{ji}$ are both false, $\hat{z}_{ij} = 1$; (iv) $\hat{y}_{ij}$ is true and $\hat{y}_{ji}$ is false, $\hat{z}_{ij} = 0$. Cases (i) and (iv) yield the same outcome; the probabilities for different values of $\hat{z}_{ij}$ are
\begin{equation}
\left\{
\begin{aligned}
&P(\hat{z}_{ij} = 0) &&= 2\epsilon(1-\epsilon), \\
&P(\hat{z}_{ij} = y_{ij}\ \text{[True outcome (TO)]}) &&= (1-\epsilon)^2, \\
&P(\hat{z}_{ij} = -y_{ij}\ \text{[Inverse outcome (IO)]}) &&= \epsilon^2. \\
\end{aligned}
\right.
\end{equation}
The Copeland ranking is based on the scores that sum on the matrix $\mathbf{Z}$: $S_{Copeland}(i) = \sum_j^{N-1} \hat{z}_{ij}$. We analyze the probability of having the true score $S_{Copeland}(i)$. In Appendix D, we prove that for a text of any true ranking in the list, the probability of obtaining its true score $S_{Copeland}$ decreases uniformly as $N$ increases (see Figure \ref{fig7}a for the visualization of this result). This directly means that the content comparison is \textit{poorly scalable}, in the sense of using it to construct the ranking over a large set of texts. 

We use the subset of $\mathbf{Z}$ (randomly selecting $n$ texts; averaging over 500 random selections) to compute $S_{Copeland}$ and then its deviation from the perfect sequence: the sum of the absolute distances between elements in the two ordered sequences, denoted as $\Delta_S$. We use $\Delta_S$ to quantify the scalability loss (Figure \ref{fig7}b). Results confirm that $\Delta_S$ grows as $n$ increases, at a speed inferior to the case of an all-zero $\mathbf{Z}$ matrix, where there is the closed form of $\Delta_S = N(N+1)/2$. 

Notably, with this model, the $\Delta_S$ plot can help yield an estimate for the error probability $\epsilon$ in the LLM output. For a specific value of $\epsilon$, equation (11) realizes a set of probabilities for elements in $\mathbf{Z}$ to take values of -1, 0, 1, respectively; the resulting $\mathbf{Z}$ matrix is used to compute $\Delta_S$ at this $\epsilon$ level. The error-free case ($\epsilon=0$) reproduces the perfect sequence ($\Delta_S = 0$); as $\epsilon$ increases, the elements in $\mathbf{Z}$ become more random, and the deviation $\Delta_S$ increases. Fitting the $\Delta_S$ plot from the measured $\mathbf{Z}$ to different $\Delta_S$ plots from $\epsilon$-specific $\mathbf{Z}$s then helps reveal the $\epsilon$ level in the LLM output. From our experiments, results show that the best-fit $\epsilon$ is 0.41/0.40/0.35/0.36/0.22 for RA/RN/SI/JM/MS articles (Figure \ref{fig7}b), i.e., the LLM outputs content preferences with an error rate of 22$\%$-41$\%$, lower than the 60$\%$ reported in \citet{LS2023}. 

\textbf{Summary on \textit{scalability}.} Our results suggest that using LLMs to rank texts through pairwise text comparison is feasible yet faintly scalable (i.e., the performance gets worse when more texts are compared) due to the lack of robustness regarding the order of comparison. The limitation may arise from its stochasticity \citep[e.g.,][]{BGet2021}. As a result, the LLM does not serve well as a \textit{judgmental arbiter}, and its application potential in text ranking is much limited.

\newpage

\subsection{\textit{Discrimination} of content scoring}

\begin{figure}[H]
    \centering
    \includegraphics[width=\linewidth]{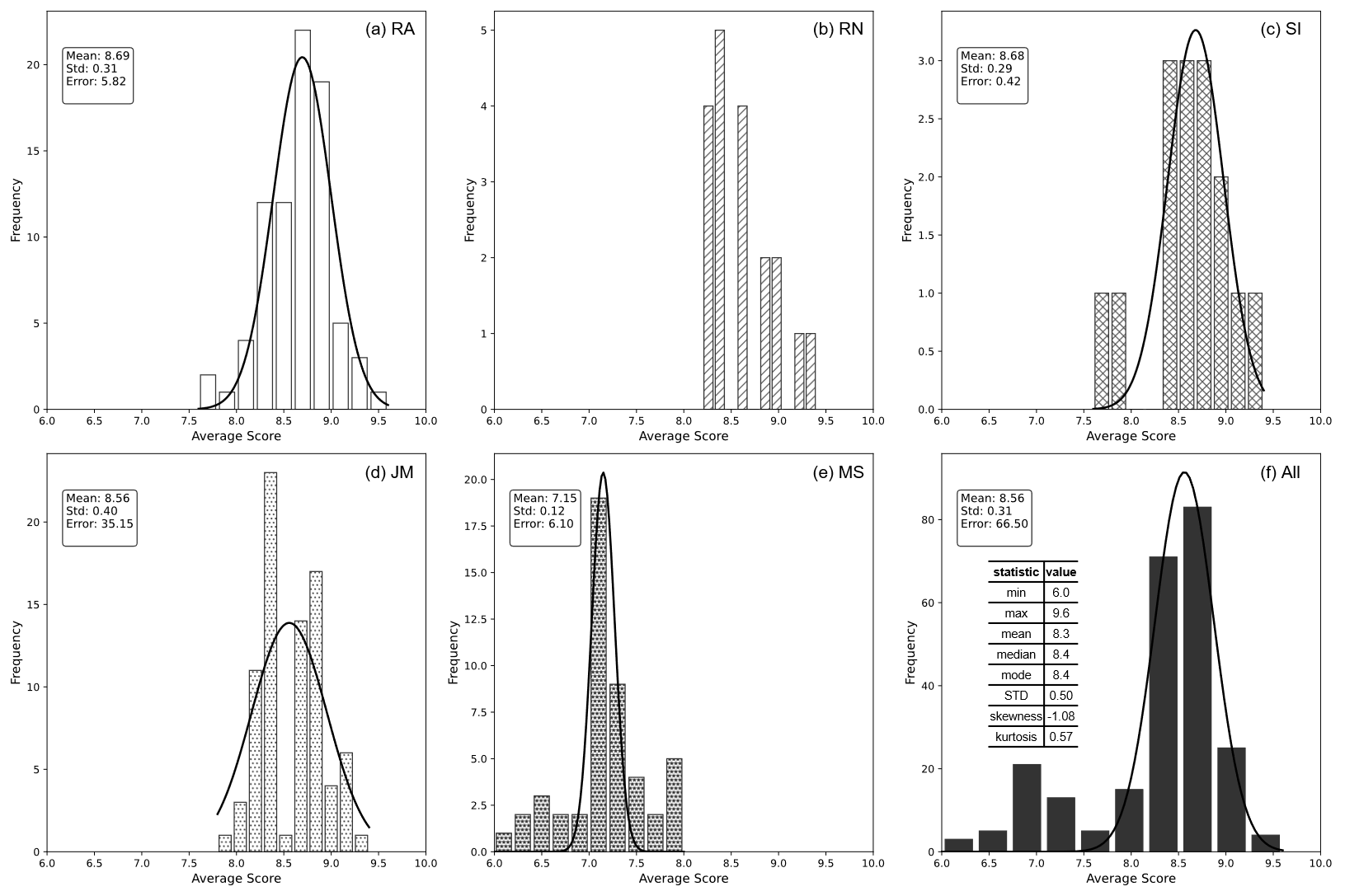}
    \caption{Content Scoring (Task 3) -- Distribution of the LLM-output scores $\hat{w}_i$ at different article types and for the whole set of articles (inset: statistics). Solid lines: best-fit Gaussian distributions.}
    \label{fig8}
\end{figure}

To report the results of content scoring at Task 3, we show the distribution of the LLM-output scores $\hat{w}_i$ at different article types (Figure \ref{fig8}). For the whole set of articles, the mean score is 8.3; median, mode, STD, skewness, and kurtosis are 8.4, 8.4, 0.50, -1.08, and 0.57, respectively; minimum/maximum score is 6.0/9.6.

We investigate a shrinkage bias in the LLM-output scores, considering that the LLM scores are skewed (presumably upwards, similar to human arbiters' scores (Figures 4 and 12); see evidence in e.g., \citet{DT2024}). Suppose that the (unknown) ground-truth score $w$ is drawn from the scale [1,10], while the LLM-output score $\hat{w}$ is drawn from a narrower scale $[1\leq H_l, H_h\leq10]$. The score distortion then follows:
\begin{equation}
\begin{aligned}
& (10-w):(w-1) = (H_h-\hat{w}):(\hat{w}-H_l) \\
\Rightarrow & \hat{w} = \frac{H_h-H_l}{9}w + \frac{10}{9}H_l - \frac{1}{9}H_h; w = \frac{9}{H_h-H_l}\hat{w} + \frac{H_h-10H_l}{H_h-H_l}.
\end{aligned}
\end{equation}
Suppose the score interval for $\hat{w}$ is 0.1; the condition for two ground-truth scores to be distinguishable is:
\begin{equation}
\begin{aligned}
\Delta \hat{w} \geq 0.1 \Rightarrow \frac{H_h-H_l}{9} \Delta w \geq 0.1 \Rightarrow \Delta w \geq \frac{0.9}{H_h-H_l}.
\end{aligned}
\end{equation}
The LLM-output scores can thus only discriminate a sufficiently large ground-truth score difference $\Delta w$. When the biased score range $(H_h-H_l)$ is narrow, the condition for score discrimination is more stringent: two ground-truth scores must have a very large difference so that their discrimination can manifest in the LLM-output scores. Note that $H_l$ and $H_h$ are not equal to the min and max of the measured scores $\hat{w}$; the $\hat{w}$ falls within $[H_l, H_h]$ but its distances from the two boundaries can be substantial. 

To further analyze the condition for score discrimination, we may impose constraints on the distribution of ground-truth scores $w$. For example, we consider that these articles' quality averages at $w_{ave}$ and that the minimum is $w_{min}$. Imposing these two metrics is valid as the ``ground-truth'' scores are eventually subjective and we can constrain their distribution with reasonable empirical considerations. This gives rise to
\begin{equation}
\begin{aligned}
\frac{9}{H_h-H_l}\hat{w}_{ave} + \frac{H_h-10H_l}{H_h-H_l} = w_{ave} \Rightarrow H_h-H_l = \frac{9(\hat{w}_{ave} - H_l)}{w_{ave}-1},\\
\frac{9}{H_h-H_l}\hat{w}_{min} + \frac{H_h-10H_l}{H_h-H_l} = w_{min} \Rightarrow H_h-H_l = \frac{9(\hat{w}_{min} - H_l)}{w_{min}-1}. 
\end{aligned}
\end{equation}
The above equation set solves
\begin{equation}
\begin{aligned}
&H_h = \frac{(10-w_{min})\hat{w}_{ave}-(10-w_{ave})\hat{w}_{min}}{w_{ave} - w_{min}}; H_l = \frac{(w_{ave}-1)\hat{w}_{min}-(w_{min}-1)\hat{w}_{ave}}{w_{ave} - w_{min}}; \\
&\Rightarrow \Delta w \geq 0.1\frac{w_{ave} - w_{min}}{\hat{w}_{ave} - \hat{w}_{min}},
\end{aligned}
\end{equation}
proportional to the LLM score interval 0.1. When the LLM score range shrinks, $w_{ave} - w_{min}$ is always no less than $\hat{w}_{ave} - \hat{w}_{min}$, and the condition for discriminating ground-truth scores is stricter.

\begin{figure}[h!]
    \centering
    \includegraphics[width=0.85\linewidth]{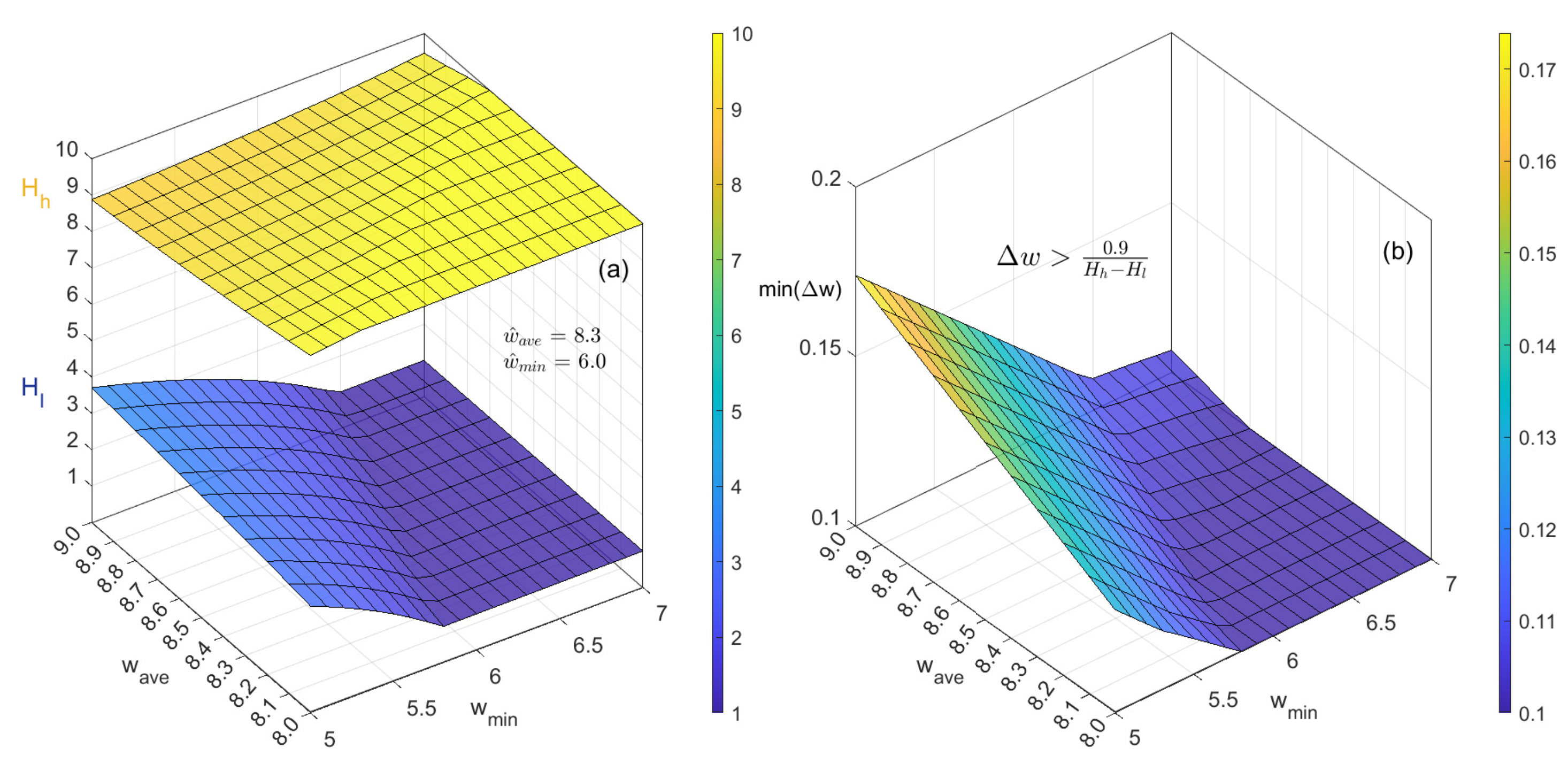}
    \caption{Content Scoring (Task 3) -- Theoretical results. (a) LLM-output score range $[H_l,H_h]$ as a function of the ground-truth score minimum $w_{min}$ and average $w_{ave}$. (b) Condition for discriminating ground-truth scores, $min(\Delta w)$. Measured $\hat{w}_{ave} = 8.3$, $\hat{w}_{min} = 6.0$.}
    \label{fig9}
\end{figure}

Metrics $\hat{w}_{ave}$, $\hat{w}_{min}$ can be measured from the LLM-output scores; for our results (Figure \ref{fig8}), $\hat{w}_{ave} = 8.3$, $\hat{w}_{min} = 6.0$. Metrics $w_{ave}$, $w_{min}$ can be determined with empirical considerations: in the current case, as articles in the dataset are peer-reviewed and published in top journals, their quality is endorsed and thus relatively large values for $w_{ave}$ and $w_{min}$ are appropriate. We consider $w_{ave}\in [8.0, 9.0]$, $w_{min}\in [5.0, 7.0]$ and show the resulting $H_h$, $H_l$, and $min(\Delta w)$ (Figure \ref{fig9}). 

When the desired score spread is large, i.e., the ground-truth $w_{min}$ is small and $w_{ave}$ is large, the shrinkage of the LLM score range is severe: low $H_h$, high $H_l$, narrow $H_h-H_l$, and large $min(\Delta w)$, i.e., the condition for score discrimination is strict. By contrast, when the ground-truth $w_{min}$ approaches $w_{ave}$, which means that ground-truth scores hardly spread, or when they spread similarly to the LLM-output scores ($w_{ave} - w_{min} \sim \hat{w}_{ave} - \hat{w}_{min}$), the corresponding $[H_l,H_h]$ then approaches the ground-truth scale [1, 10], and $min(\Delta w)\sim 0.1$.

\textbf{Summary on \textit{discrimination}.} Our results suggest that adopting LLMs to grade academic texts is feasible but prone to poor discrimination due to the possibility of skewed scores and score range shrinkage. The limitation may arise from its tempered objectivity \citep[e.g.,][]{DMet2024}. The LLM does not play the role of \textit{knowledgeable arbiter} well, and its application potential in text rating is questionable.

\newpage

\subsection{\textit{Insightfulness} of content reflection}

We report the results of content reflection at Task 4. The LLM-generated article critiques are itemized. We obtain three pieces of critiques in one run and in total 15 pieces of critiques in the five runs. We split each critique into the subject and the detail: for example, one critique reads ``Lack of Theoretical Grounding: While the introduction mentions the relevance of process virtualization theory (PVT), it fails to provide a detailed theoretical grounding for the study. A more robust theoretical framework would strengthen the paper's structure and clarify the research objectives.'' Here, the subject is ``Lack of Theoretical Grounding'' and the rest is the detail.

Across the 3690 pieces of critique in total, there are 2376 unique subjects, and 287 of them appear more than once; the most frequent three are ``lack of empirical evidence,'' ``limited generalizability,'' and ``limited scope of analysis.'' We group similar terms (e.g., ``lack of empirical evidence,'' ``empirical validation,'' ``empirical data''). The grouped top-25 critique subjects are shown in Figure \ref{fig10} (see Appendix E for the full list).

\begin{figure}[h!]
    \centering
    \includegraphics[width=0.72\linewidth]{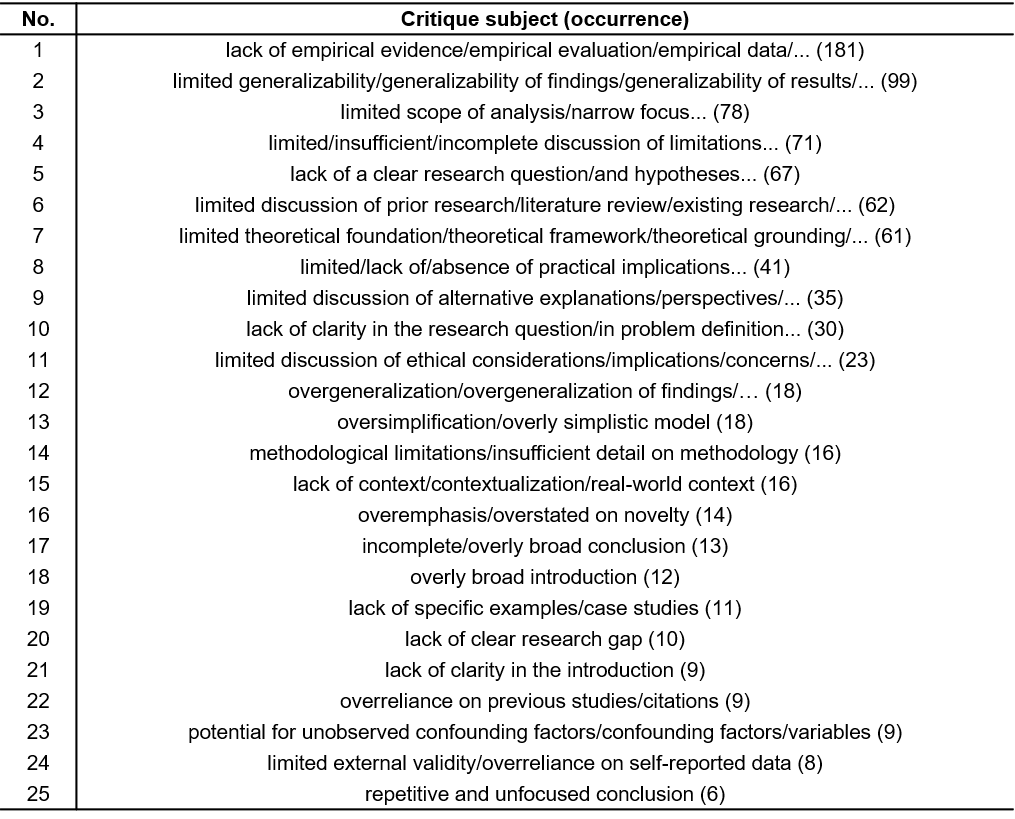}
    \caption{Content reflection (Task 4) -- Statistics of critique subjects. Grouped top-25 critique subjects.}
    \label{fig10}
\end{figure}

\textbf{Internal/external evaluation of article critiques.} We evaluate the entire critique (subject and detail) on its linguistic quality (internal evaluation) using the $IE_{\bullet}$ scores and its similarity to the input text (external evaluation) using the $EE_{\bullet}$ scores. For internal evaluation, we calculate the linguistic quality of the input text as the background. For external evaluation, we consider (i) comparing the three critiques at one run to the input text and then taking the average $EE_{\bullet}$ of the five runs, and also (ii) merging the 15 critiques in a single text and then comparing it to the input text. 

\begin{figure}[htbp]
    \centering
    \includegraphics[width=0.85\linewidth]{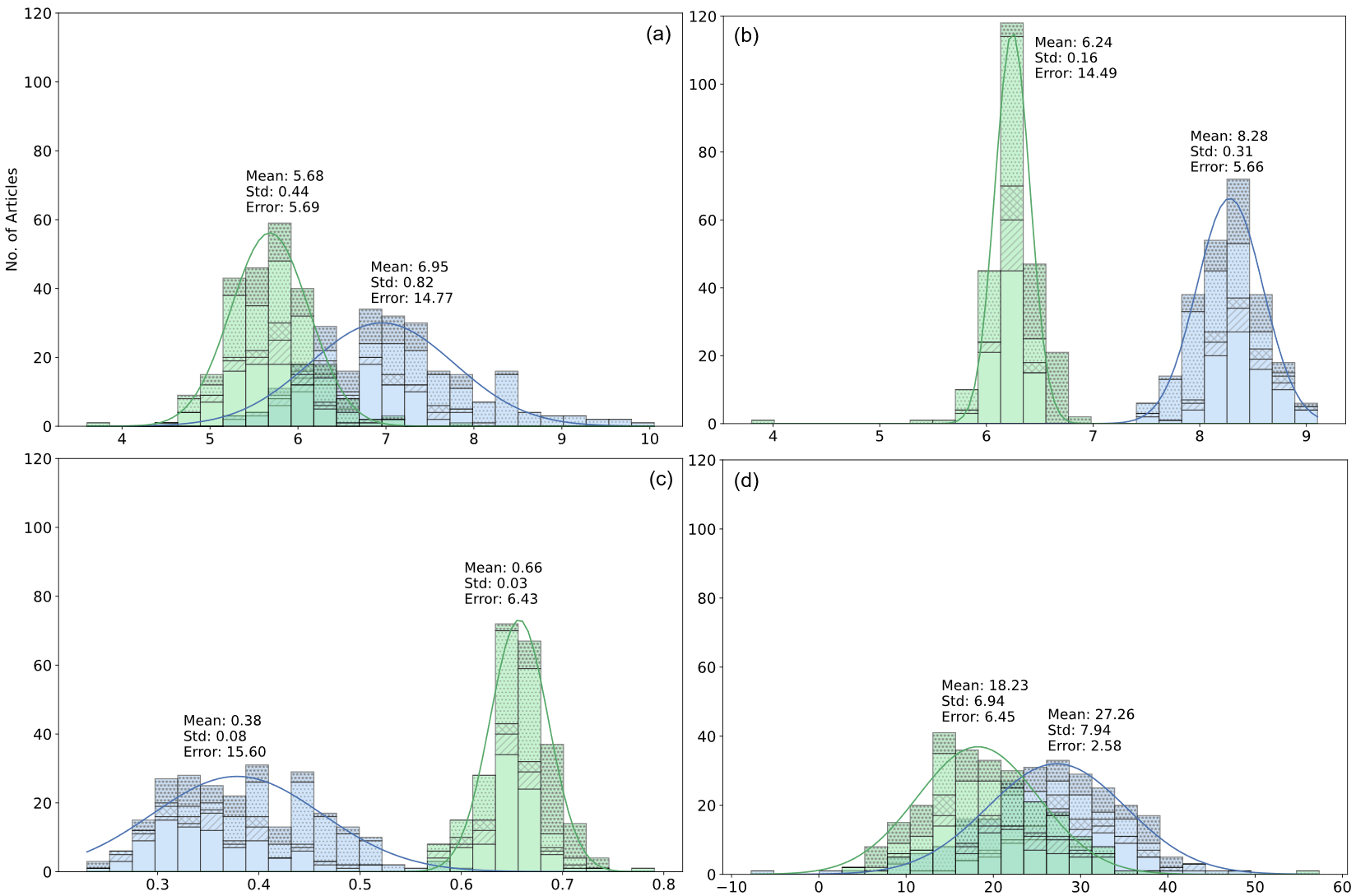}
    \caption{Content reflection (Task 4) -- Internal evaluation of the input text (blue)/LLM-generated critiques (green) for different article types (RA [empty], RN [slash], SI [cross], JM [dot], MS [star]) under metrics $IE_{H-density/entropy/TTR/FK}$ (a/b/c/d; x-axis: scores). Solid lines: best-fit Gaussian distributions.}
    \label{fig11}
\end{figure}

Results suggest that (Figures 10, 11), for internal evaluation, the linguistic quality of the input text (blue) is significantly higher than that of the LLM-generated critiques (green), when measured by $IE_{H-density/entropy/FK}$. This quality discrepancy is larger than that of comparing ground-truth/LLM-generated abstracts (Figure \ref{fig3}), as revealed by the paired t-test (Table \ref{tab8}). This suggests that LLM-generated critiques have a smaller linguistic complexity than LLM-generated abstracts. When measured by $IE_{TTR}$ (left bottom of Figure \ref{fig11}), the LLM-generated text obtains a higher score than the input article text, consistent with Figure \ref{fig3} that suggests a larger vocabulary use in the LLM output.

\begin{figure}[htbp]
    \centering
    \includegraphics[width=0.85\linewidth]{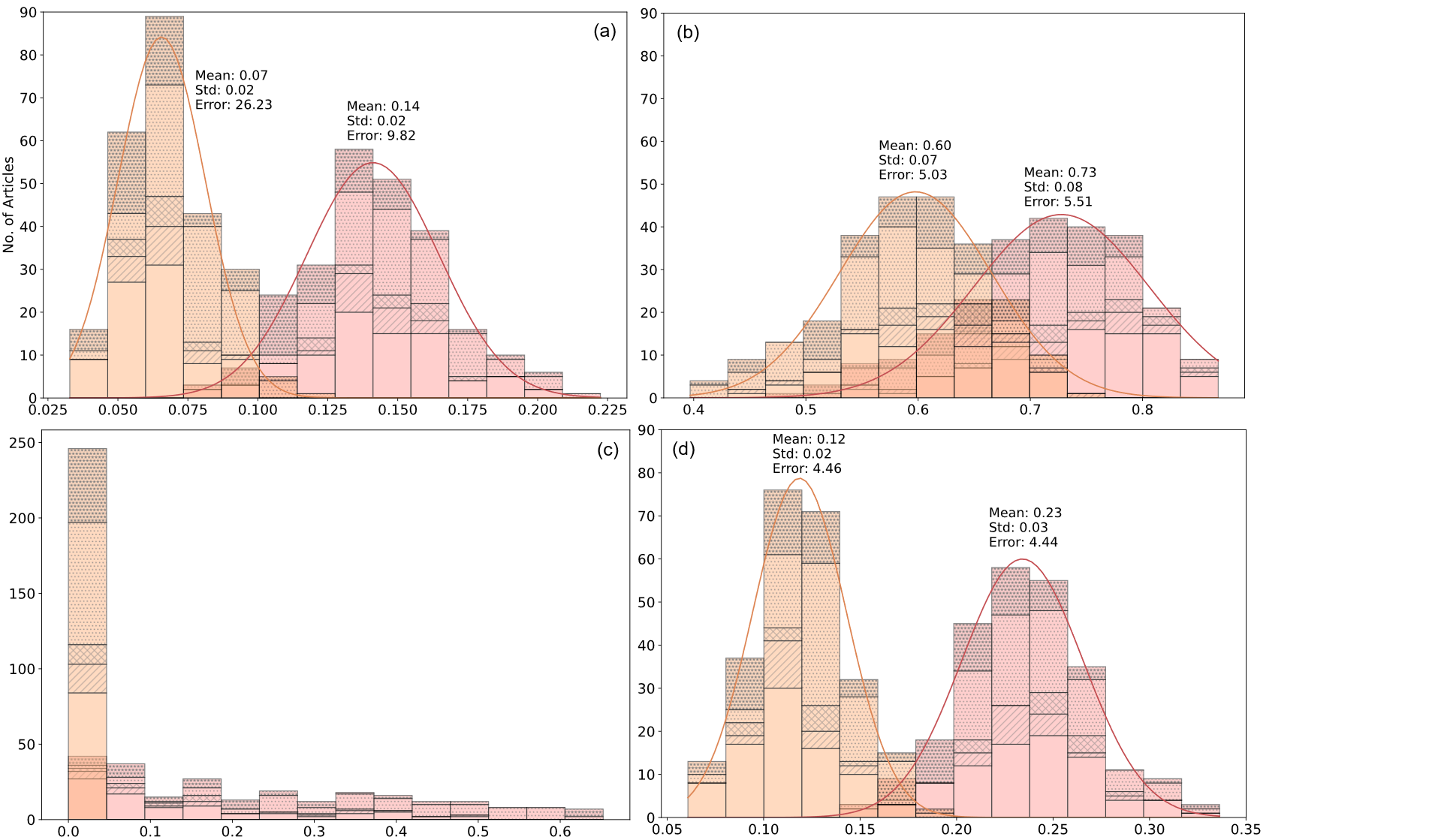}
    \caption{Content reflection (Task 4) -- External evaluation comparing the
merged critiques (red) or the three-piece critiques (orange) to the input text for different article types (RA [empty], RN [slash], SI [cross], JM [dot], MS [star]) under metrics $EE_{Jac/cos/BLEU/ROUGE}$ (a/b/c/d; x-axis: scores). Solid lines: best-fit Gaussian distributions.}
    \label{fig12}
\end{figure}

\newpage

\begin{table}[H]
\centering
\begin{subtable}[t]{\textwidth}
    \centering
    \caption{Paired t-test for the $IE_{\bullet}$ difference between the input text and the LLM-generated critiques (Figure \ref{fig11}) and the $EE_{\bullet}$ difference between the merged critique case and the three-piece critique case (Figure \ref{fig12})}
    \resizebox{0.85\textwidth}{!}
{   \begin{tabular}{lccc|cc|cc|ccc}
    \hline
     & \multicolumn{1}{c}{Count} & \multicolumn{4}{c}{\textbf{$IE_{H-density}$}} & \multicolumn{4}{c}{\textbf{$IE_{entropy}$}} \\
    Type & & \multicolumn{2}{c}{input text} & \multicolumn{2}{c}{critique} & \multicolumn{2}{c}{input text} & \multicolumn{2}{c}{critique} \\
    \hline
     & & Mean & Std & Mean & Std & Mean & Std & Mean & Std \\
    \hline
    RA & 84 & 6.663 & 0.619 & \textbf{5.621}*** & 0.475 & 8.390 & 0.302 & \textbf{6.213}*** & 0.149 \\
    RN & 19 & 6.562 & 0.0.742 & \textbf{5.701}*** & 0.386 & 8.321 & 0.310 & \textbf{6.203}*** & 0.138 \\
    SI & 13 & 6.831 & 0.676 & \textbf{5.861}** & 0.370 & 8.450 & 0.298 & \textbf{6.294}*** & 0.110 \\
    JM & 81 & 7.833 & 0.899 & \textbf{5.648}*** & 0.475 & 8.064 & 0.260 & \textbf{6.125}*** & 0.318 \\
    MS & 49 & 6.919 & 0.566 & \textbf{5.831}*** & 0.552 & 8.342 & 0.249 & \textbf{6.546}*** &  0.134 \\
    All & 246 &  7.100 & 0.892 & \textbf{5.691}*** & 0.482 & 8.271 & 0.314 & \textbf{6.254}*** & 0.263 \\
    \hline
     & \multicolumn{1}{c}{Count} & \multicolumn{4}{c}{\textbf{$IE_{TTR}$}} & \multicolumn{4}{c}{\textbf{$IE_{FK}$}} \\
    Type & & \multicolumn{2}{c}{input text} & \multicolumn{2}{c}{critique} & \multicolumn{2}{c}{input text} & \multicolumn{2}{c}{critique} \\
    \hline
     & & Mean & Std & Mean & Std & Mean & Std & Mean & Std \\
    \hline
    RA & 84 & 0.354 & 0.058 & \textbf{0.647}*** & 0.027 & 28.332 & 6.528 & \textbf{19.681}*** & 7.209 \\
    RN & 19 & 0.361 & 0.063 & \textbf{0.640}*** & 0.029 & 29.909 & 6.374 & \textbf{22.209}*** & 6.351 \\
    SI & 13 & 0.348 & 0.050 & \textbf{0.645}*** & 0.035 & 30.825 & 5.321 & \textbf{18.470}*** & 5.193 \\
    JM & 81 & 0.447 & 0.060 & \textbf{0.649}*** & 0.030 & 21.575 & 8.743 & 20.149 & 7.093 \\
    MS & 49 & 0.363 & 0.063 & \textbf{0.692}*** & 0.021 & 29.813 & 7.278 & \textbf{14.802}*** & 5.558 \\
    All & 246 & 0.387 & 0.073 & \textbf{0.656}*** & 0.033 & 26.656 & 8.209 & \textbf{19.009}*** & 7.038 \\
    \hline
    \hline
    
     & \multicolumn{1}{c}{Count} & \multicolumn{4}{c}{\textbf{$EE_{Jac}$}} & \multicolumn{4}{c}{\textbf{$EE_{cos}$}} \\
    Type & & \multicolumn{2}{c}{merged critique} & \multicolumn{2}{c}{three-piece critique} & \multicolumn{2}{c}{merged critique} & \multicolumn{2}{c}{three-piece critique} \\
    \hline
     & & Mean & Std & Mean & Std & Mean & Std & Mean & Std \\
    \hline
    RA & 84 & 0.142 & 0.026 & \textbf{0.063}*** & 0.015 & 0.740 & 0.072 & \textbf{0.608}*** & 0.066 \\
    RN & 19 & 0.144 & 0.012 & \textbf{0.066}*** & 0.010 & 0.736 & 0.078 & \textbf{0.610}*** & 0.067 \\
    SI & 13 & 0.146 & 0.019 & \textbf{0.065}*** & 0.010 & 0.757 & 0.054 & \textbf{0.625}*** & 0.044 \\
    JM & 81 & 0.152 & 0.023 & \textbf{0.077}*** & 0.015 & 0.696 & 0.072 & \textbf{0.569}*** & 0.069 \\
    MS & 49 & 0.127 & 0.024 & \textbf{0.063}*** & 0.015 & 0.680 & 0.074 & \textbf{0.583}*** & 0.066 \\
    All & 246 & 0.143 & 0.025 & \textbf{0.068}*** & 0.016 & 0.714 & 0.076 & \textbf{0.591}*** & 0.068 \\
    \hline
     & \multicolumn{1}{c}{Count} & \multicolumn{4}{c}{\textbf{$EE_{BLEU}$}} & \multicolumn{4}{c}{\textbf{$EE_{ROUGE}$}} \\
    Type & & \multicolumn{2}{c}{merged critique} & \multicolumn{2}{c}{three-piece critique} & \multicolumn{2}{c}{merged critique} & \multicolumn{2}{c}{three-piece critique} \\
    \hline
     & & Mean & Std & Mean & Std & Mean & Std & Mean & Std \\
    \hline
    RA & 84 & 0.143 & 0.137 & \textbf{$<$0.001}*** & 0.001 & 0.236 & 0.038 & \textbf{0.113}*** & 0.024 \\
    RN & 19 & 0.170 & 0.149 & \textbf{$<$0.001}*** & 0.001 & 0.237 & 0.018 & \textbf{0.117}*** & 0.016 \\
    SI & 13 & 0.130 & 0.099 & \textbf{$<$0.001}*** & 0.000 & 0.246 & 0.027 & \textbf{0.118}*** & 0.017 \\
    JM & 81 & 0.368 & 0.157 & \textbf{0.003}*** & 0.005 & 0.243 & 0.031 & \textbf{0.132}*** & 0.022 \\
    MS & 49 & 0.237 & 0.184 & \textbf{0.002}*** & 0.004 & 0.217 & 0.033 & \textbf{0.115}*** & 0.025 \\
    All & 246 & 0.237 & 0.181  & \textbf{0.001}*** & 0.004 & 0.235 & 0.034 & \textbf{0.120}*** & 0.024 \\
    \hline
    \end{tabular}
    }
\end{subtable}


\begin{subtable}[t]{\textwidth}
    \centering
    \caption{One-Way ANOVA results on different article types (internal/external/human valuation)}
    \resizebox{0.85\textwidth}{!}
{   \begin{tabular}{lccc|ccc}
    \hline
    Measure ($IE_{\bullet}$) & \multicolumn{3}{c|}{input text} & \multicolumn{3}{c}{critique} \\
     & sum\_sq & F & p-value & sum\_sq & F & p-value \\
    \hline
    \textbf{$IE_{H-density}$} & 67.680 & 32.065 & \textbf{\textless0.001}*** & 1.897 & 2.05 & 0.087 \\
    \hline
    \textbf{$IE_{entropy}$} & 5.351 & 17.187 & \textbf{\textless0.001}*** & 5.739 & 30.755 & \textbf{\textless0.001}*** \\
    \hline
    \textbf{$IE_{TTR}$} & 0.448 & 31.242 & \textbf{\textless0.001}*** & 0.083 & 27.141 & \textbf{\textless0.001}*** \\
    \hline
    \textbf{$IE_{FK}$} & 3242.834 & 14.728 & \textbf{\textless0.001}*** & 1213.795 & 6.695 & \textbf{\textless0.001}*** \\
    \hline
    \hline
    Measure ($EE_{\bullet}$) & \multicolumn{3}{c|}{merged critique} & \multicolumn{3}{c}{three-piece critique} \\
     & sum\_sq & F & p-value & sum\_sq & F & p-value \\
    \hline
    \textbf{$EE_{Jac}$} & 0.019 & 8.463 & \textbf{\textless0.001}*** & 0.010 & 11.788 & \textbf{\textless0.001}***  \\
    \hline
    \textbf{$EE_{cos}$} & 0.172 & 8.195 & \textbf{\textless0.001}*** & 0.087 & 4.999 & \textbf{\textless0.001}*** \\
    \hline
    \textbf{$EE_{BLEU}$} & 2.375 & 25.147 & \textbf{\textless0.001}*** & 0.000 & 7.507 & \textbf{\textless0.001}*** \\
    \hline
    \textbf{$EE_{ROUGE}$} & 0.023 & 5.133 & \textbf{0.001}*** & 0.016 & 7.771 & \textbf{\textless0.001}*** \\
    \hline
    \hline
     Measure ($HE^{cri}$) & \multicolumn{3}{c|}{critique} &  \\
     & sum\_sq & F & p-value &  \\
    \hline
    \textbf{$HE^{cri}$} & 0.716 & 0.685 & 0.562 & & &  \\
    \hline
    \end{tabular}
    }
\end{subtable}

\caption{Content reflection (Task 4) – Inference analyses. $^*/^{**}/^{***}$: $p<0.05/0.01/0.001$.}
\label{tab8}
\end{table}

\newpage

For external evaluation, merging the 15 critiques in a single text (option (ii), red) leads to a higher similarity to the input text than the three-critique case (option (i), orange) at $EE_{Jac/cos/ROUGE}$, as revealed by the paired t-test (Table \ref{tab8}). This suggests that the content of critiques differs in different runs. Compared to LLM-generated abstracts (Table 6), LLM-generated critiques (Table 7) have a comparable similarity to the article text. This implies that LLM-generated critiques refer to the input text to a considerable extent, approaching content reproduction. For $EE_{BLEU}$ (left bottom of Figure \ref{fig12}), as the critique is substantially shorter than the article text, the score has very small values (equation (7)) and does not apply to this situation well, especially at the three-piece critique. Similar to Task 1, results suggest a significant difference across article types at both internal and external evaluation of LLM-generated critiques (Table 7b).

\textbf{Human evaluation of article critiques.} We then ask human arbiters to evaluate the critiques. For each article, the arbiter submits a score $HE^{cri}\in [1,5]$ to indicate the insightfulness of the critiques after reading them altogether with the input text. 

\begin{figure}[h!]
    \centering
    \includegraphics[width=\linewidth]{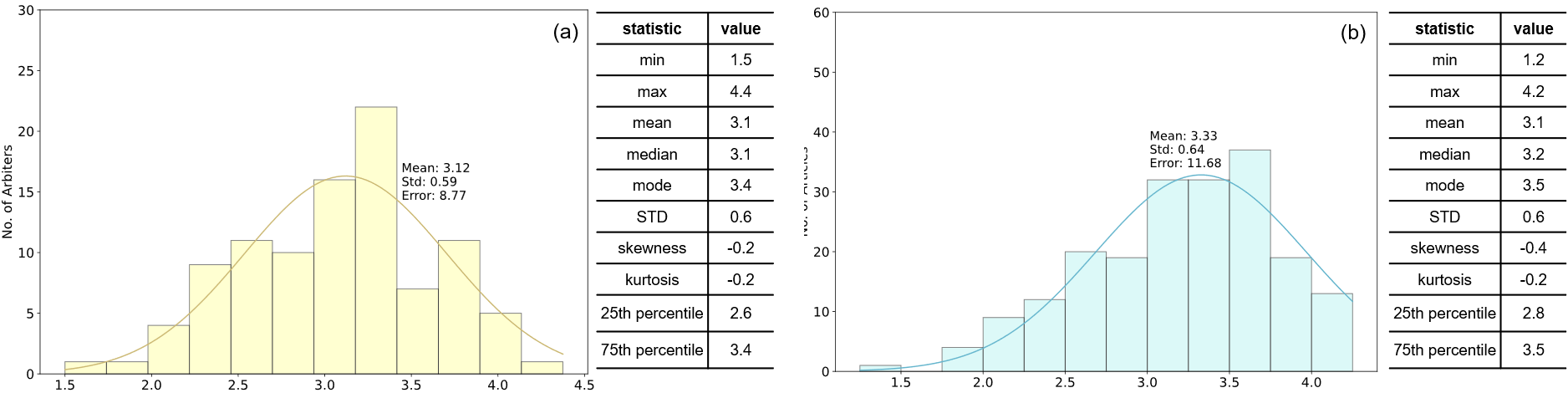}
    \caption{Content reflection (Task 4) -- Human evaluation results. (a) Distribution and statistics of average scores from arbiters. (b) Distribution and statistics of average scores $HE^{cri}$ for LLM-generated critiques. Solid lines: best-fit Gaussian distributions.}
    \label{fig13}
\end{figure}

Arbiters' mean scores given to the eight assigned critiques fell within [1.5, 4.4] and averaged at 3.1 (Figure \ref{fig13}a). Collecting the four (three in a few cases) evaluations from arbiters, the mean score for the set of critiques at each article, $HE^{cri}$, is distributed between 1.2 and 4.2, averaging at 3.1 and mostly concentrated within [2.5, 4.0] (Figure \ref{fig13}b). $HE^{cri}$ does not show a significant difference across article types (Table \ref{tab8}b).

We compare the distributions of $HE^{abs}$ and $HE^{cri}$ at the mean/median/mode, Cohen's d (effect size), variance (Levene's test), and shape (KS test). Results (Table \ref{tab9}) suggest significantly different score distributions and in particular a lower score ($p < 0.01$) for LLM-generated critiques than for LLM-generated abstracts, reflecting professional human arbiters' decreased confidence in LLM-generated critiques (see more details on human evaluation results in Appendix F).

\begin{table}[htbp]
\centering
\small
\begin{tabular}{lccc}
\hline
\textbf{Statistic} & $HE^{abs}$ & $HE^{cri}$ & \textbf{Difference} \\
\hline
Mean & 3.34 & 3.08 & \textbf{0.26}** \\
Median & 3.33 & 3.25 & 0.08 \\
Mode & 3.33 & 3.50 & -0.17 \\
Cohen's d (Effect Size) & -- & -- & 0.47 \\
Variance (Levene's Test) & -- & -- & ** \\
Distribution Shape (KS Test) & -- & -- & ** \\
\hline
\end{tabular}
\caption{Comparison of human evaluation results at Tasks 1 and 4. $^{**}$: $p < 0.01$. }
\label{tab9}
\end{table}

\textbf{Summary on \textit{insightfulness}.} Overall, our results from internal, external, and human evaluation of LLM-generated critiques suggest that the LLM's qualitative reflection on the text is self-consistent but hardly insightful to inspire meaningful research. Moreover, human users give less acknowledgment to the LLM's insightfulness than they do to its output from text summary and paraphrasing. The limitation may arise from the LLM's non-understanding \citep[e.g.,][]{M2023}. The LLM does not sufficiently qualify as a \textit{collaborator} in scientific research, and its application potential is far from endorsing a full research assistantship.

\newpage

\section{Robustness Checks}

We report the results of Experiments 1-4 under variants of the baseline prompts. These robustness checks (Table \ref{tab2}) concern four aspects of the input prompt: \textit{semantic robustness of the prompt} (E3-1/2/3, E4-1/2/3), \textit{richness of the prompt} (E1-3, E2-3, E2-4, E4-4), \textit{abundance of the data prompt} (E1-1, E1-2, E2-1, E2-2), and \textit{specificity of the instruction prompt} (E1-4, E1-5, E2-5, E4-5). For each robustness check, we report the deviation from the main result in the baseline experiment (Experiments $\bullet$-0, Sections 5.1-5.3), considering the value of a principal measurement in each result: for Experiments 1-$\bullet$ and 2-$\bullet$ (Task 1, content reproduction), consider the mean of $IE_{\bullet}$ or $EE_{\bullet}$, $\overline{IE}_{\bullet}$ or $\overline{EE}_{\bullet}$ (Section 5.1); for Experiments 3-$\bullet$ (Task 2, content comparison), consider the difference between the obtained $S_{Copeland}$ and the perfect sequence, $\Delta_S$ (Section 5.2); for Experiments 4-$\bullet$ (Task 3, content scoring), consider the mean of the LLM-output score, $\hat{w}_{ave}$ (Section 5.3). We show the percentage deviation of these measurements at corresponding robustness checks; for Experiments 1-$\bullet$ and 2-$\bullet$, we further show the average deviation across the measurements. These deviations are calculated for each article; at each check, we report the mean and variance of the deviation over an ensemble of 50 articles sampled from the whole set. 

\begin{figure}[h!]
    \centering
    \includegraphics[width=0.75\linewidth]{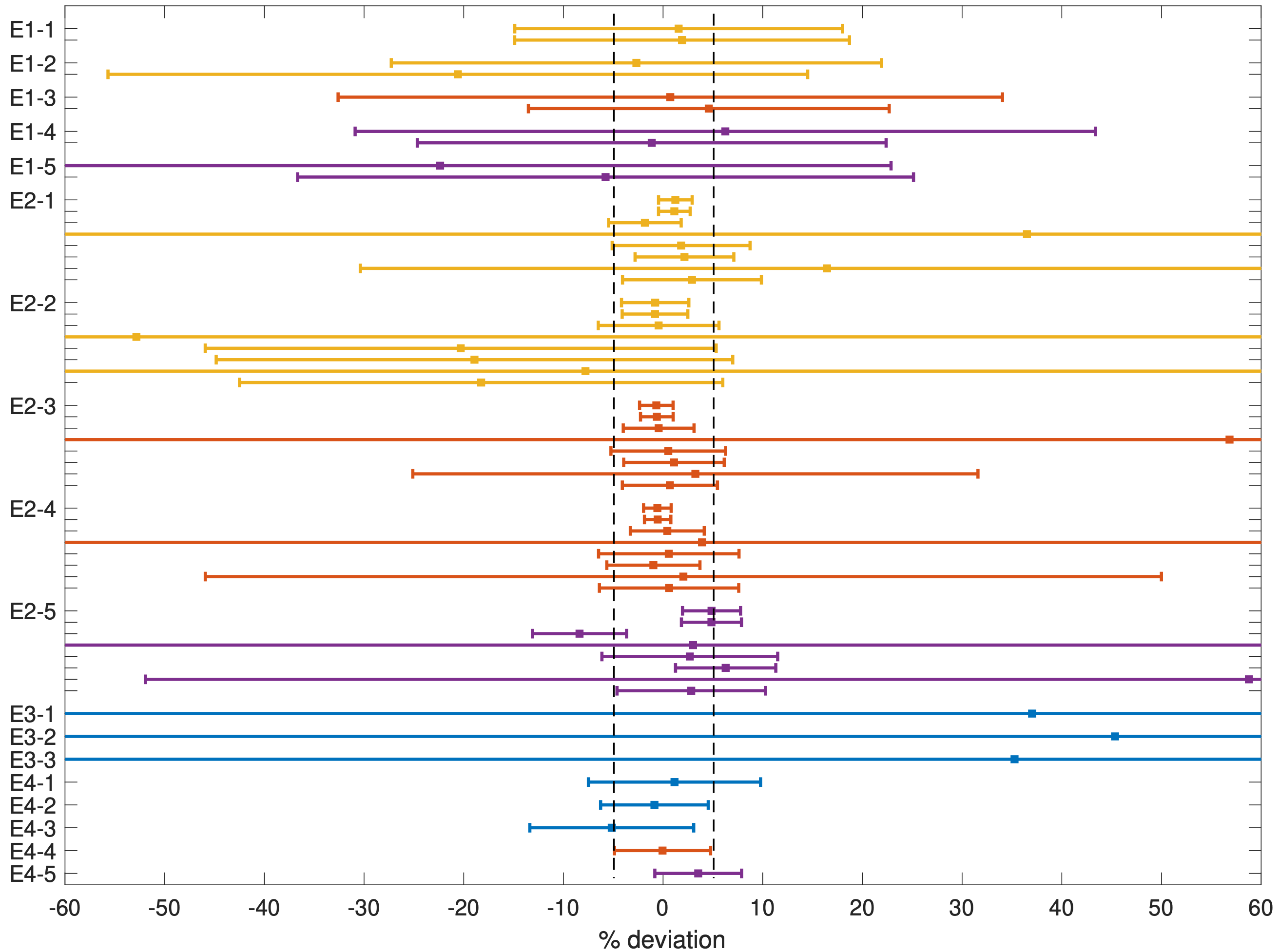}
    \caption{Robustness checks. Percentage deviations (dots and error bars: means and variances) of the measurements in baseline experiments (Experiments $\bullet$-0): $\overline{IE}_{\bullet}/\overline{EE}_{\bullet}$ for Experiments 1-$\bullet$ (two measurements) and 2-$\bullet$ (eight measurements); $\Delta_S$ for Experiments 3-$\bullet$; 
    $\hat{w}_{ave}$ for Experiments 4-$\bullet$. Consider four aspects of the input prompt: semantic robustness of the prompt (blue), richness of the prompt (red), abundance of the data prompt (yellow), and specificity of the instruction prompt (purple). Dash lines: 5$\%$ deviations.}
    \label{fig14}
\end{figure}

Results (Figure \ref{fig14}) show $<5\%$ deviations of these measurements in 39/58 cases, suggesting a strong robustness for our findings. Compared to other experiments, the variance is larger at (1) Experiments 1-$\bullet$, which is consistent with the more fluctuating outcome during keywords reproduction due to its higher output restriction (e.g., compared to abstract reproduction (Section 5.1)), and at (2) Experiments 3-$\bullet$, which is consistent with the substantial error rate of the LLM-output content comparison (Section 5.2). For the four aspects of the input prompt, our results suggest that semantic robustness (blue) and the specificity of the instruction prompt (purple) play a more important role in prompt robustness than the prompt's richness (red) or data abundance (yellow). The results' deviations are larger when the prompt is paraphrased or when the prompt instruction's specificity changes; the richness and data abundance influence the prompt to a lesser extent.

\newpage

\section{Metrics for Text Quality Indication}

\begin{figure}[h!]
    \centering
    \vspace{-0.45cm}
    \includegraphics[width=0.7\linewidth]{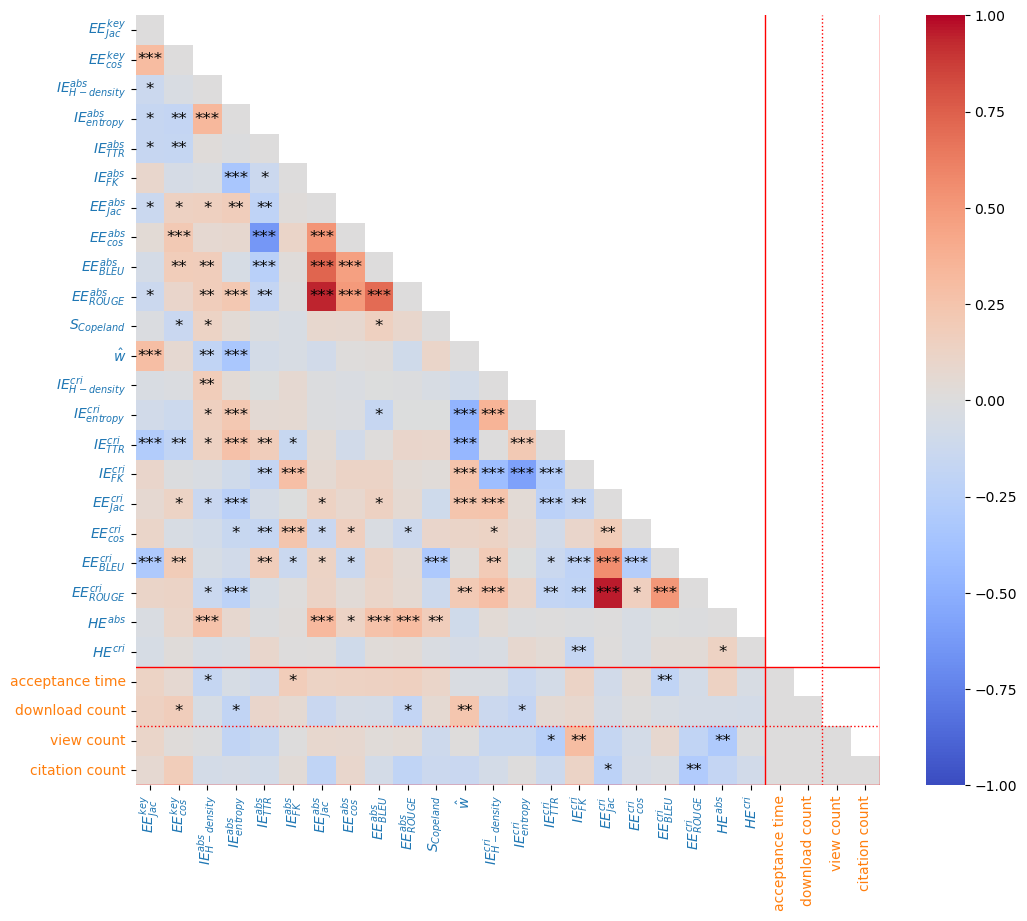}
    \caption{Correlation (Pearson's $r$) between LLM-reliant (blue) and ground-truth (orange) text metrics. $^*/^{**}/^{***}$: $p<0.05/0.01/0.001$.}
    \label{fig15}
\end{figure}

At the four tasks, we obtain metrics on the input text that are constructed based on LLMs' processing of the text: at content reproduction, the $IE_{\bullet}^{abs}$/$EE_{\bullet}^{key/abs}$/$HE^{abs}$ from internal/external/human evaluation of LLM-generated keywords/abstracts; at content comparison, the Copeland score $S_{Copeland}$; at content scoring, the LLM-output text score $\hat{w}$; and at content reflection, the $IE_{\bullet}^{cri}$/$EE_{\bullet}^{cri}$/$HE^{cri}$ from internal/external/human evaluation of LLM-generated critiques. We employ these 22 metrics that measure different aspects of LLMs' processing to indicate text quality. We study the correlations between these LLM-reliant quality metrics and the collected ground-truth quality metrics: articles' acceptance time and download count for ISR and MS articles, view count and citation count for JMIS articles (Section 4.1). We calculate Pearson's $r$ (Figure \ref{fig15}) between these 26 quality metrics and Spearman's $\rho$ and Kendall's $\tau$ (Appendix G) between their ranks.

Results (Figure \ref{fig15}) show that, in many cases, LLM-reliant metrics are significantly correlated (red) or anti-correlated (blue) with each other. By contrast, LLM-reliant metrics (text in blue) demonstrate rare correlation or anti-correlation with ground-truth metrics (text in orange). For example, $IE_{H-density}^{cri}$ (13th-column, bottom-up) has no significant correlation or anti-correlation with the four ground-truth metrics; however, it is significantly correlated with $IE_{entropy}^{cri}$ ($^{***}$), $EE_{jac}^{cri}$($^{***}$), $EE_{cos}^{cri}$($^{*}$), $EE_{BLEU}^{cri}$($^{**}$), and $EE_{ROUGE}^{cri}$($^{***}$), and it is anti-correlated with $IE_{FK}^{cri}$ ($^{***}$). Overall, the four ground-truth metrics show significant ($p<0.05$) correlation with LLM-reliant metrics in 3/22 (acceptance time), 5/22 (download count), 3/22 (view count), and 2/22 (citation count) cases, respectively.

The results are similar at Spearman's $\rho$ and Kendall's $\tau$ (Appendix G). This suggests that the two categories of metrics (LLM-reliant vs. ground-truth) do not indicate text quality along the same dimension. It points to the potential of utilizing LLMs' \textit{imperfect} yet \textit{interdisciplinary} processing of academic content to construct (arguably) less-biased metrics for articles, journals, and scholars, which can complement current diffusion-based metrics that are more or less discipline-specific and inevitably biased by human factors.

\newpage

\section{Concluding Remarks}

The output of AI tools shares features that align with human preferences \citep{MC2024}. LLMs produce ``standard'' responses -- which we desire in many contexts -- relying on their capability of conducting effective information compression \citep[e.g.,][]{S2017,W2019}, an important test for AI systems \citep{M1999} largely considered in the fabrication of LLMs \citep[e.g.,][]{TLet2023}. This capability underlies LLMs' basic text-processing function of literature digest and our expectation of employing them in aiding scientific research. Yet reduction and synthesis trades off elaboration and complexity, and we should not take LLMs' information compression for granted. We need to probe into the limitations of LLMs in this regard and the tradeoffs in their employment, investigating the costs associated with the gains in efficiency. 

In this study, we developed a guided workflow assembling known analyses and employing well-adopted metrics as well as human arbiters to evaluate the potential of recruiting LLMs to process academic text input and construct peer reviews in future scientific practice. We assessed four features of the LLM output in conducting four tasks: reliability at content reproduction, scalability at content comparison, discrimination at content scoring, and insightfulness at content reflection. Overall, we recorded a compromised performance of the tested LLM in each aspect:  the LLM's ability to summarize and paraphrase academic texts is acceptably reliable yet discouraging; its ability to rank texts through pairwise text comparison is faintly scalable; its ability to grade academic texts is prone to poor discrimination; and its qualitative reflection on a given text is self-consistent but hardly insightful in terms of inspiring meaningful research. These phenomena, consistent across metric-based internal evaluation (linguistic assessment), external evaluation (comparing to the ground truth), and human evaluation, are robust to the variation of LLM prompts in semantics, input richness, data abundance, and instruction specificity.

Specifically, we found that: (1) LLM-generated texts can employ a larger vocabulary but its word use is more concentrated than that of ground-truth texts; (2) the quality of the LLM output varies across different input article types; (3) when comparing two texts, the error rate of the LLM-output preference is $22\%-41\%$; (4) LLM-generated critiques have a smaller linguistic complexity than LLM-generated abstracts; (5) the specificity of the instruction influences the prompt more than the prompt's richness or data abundance, and the prompt is most robust to semantic variations; (6) the LLM-generated text balances abundant and frequent word use, and the LLM's abundant word use hinders the reproduction of the ground truth; (7) LLM-reliant and ground-truth text metrics do not indicate input text quality along the same dimension.

Our results suggest that the LLM's performance downgrades along with a task's increasing need for solid scientific understanding to yield desirable solutions. The LLM can reasonably play the role of \textit{oracle}, but it does not do as well when serving as a \textit{judgmental} or \textit{knowledgeable arbiter}, and it insufficiently qualifies as a \textit{collaborator}. This implies that, in the future of science, for aiding research development, LLMs' application potential can be acceptable at the level of literature digest, but it has diminishing value at text ranking and text rating, and it is far from endorsable as a full research assistant. LLMs can work well when summarizing academic content and helping researchers navigate the literature, but they do not perform as well when selecting content, evaluating content, or suggesting research directions. From these experiments, upon the acceptable reliability, unwarranted scalability, skewed discrimination, and bounded insightfulness of their output, we found it appropriate not to recommend, rather than to recommend, an unchecked use of LLMs in conducting academic peer reviews.

\subsection{Research implications}

In summary, our study brings the following insights to the audience in IS and broad managerial sciences.

(1) We organize four individual tasks that CS studies widely employ to evaluate LLMs' capabilities into a guided workflow to assess LLMs' performance in a focal application: process academic texts and output the generated abstract/keywords/critiques (content output) for the input article or a preference/score (decision output). Through a series of tasks, we look into a spectrum of LLMs' capabilities that increasingly demand a deep understanding of scientific content. This investigation goes beyond viewing LLMs as tools and looking for ameliorating their engineering, as is the case in CS studies, and points to discussing their behavior and our acknowledgment of them at the human-machine interaction, in IS studies.

(2) Our evaluation helps determine LLMs' potential to be recruited in aiding academic peer review. We exemplify an effective workflow of the four tasks with detailed instructions on the prompts. The tasks are connected to different components of the peer review and can be adopted by different stakeholders, e.g., editors selecting submissions (content comparison), reviewers summarizing content (content reproduction) or rating submissions (content scoring), and authors developing manuscripts (content reflection). At each task, we demonstrate the analyses on the LLM output in detail to inform future evaluations. These analyses are modular and fully transparent to ensure access and replicability.
    
(3) We use a unique set of text materials to assess the LLM and an abundant set of metrics in the metric-based evaluation of its output. Compared to more specialized content, the diverse content of IS studies with articles' different semantic styles provides a comprehensive testbed for assessing LLMs' text processing. We adopt four metrics each in both the internal evaluation (linguistic assessment) and the external evaluation (comparing to the ground truth) of the LLM-generated texts and compare the measurements at different types of articles (e.g., regular/special issue) at three top journals. These considerations underlie the extensibility of our evaluation. 

(4) Our exemplary analyses suggest an integrated piece of evidence against an endorsement of LLMs' text-processing capabilities. The findings are robust to the variation of LLM prompts, and we thus find it appropriate to avoid an overlooked use of LLMs in aiding academic peer reviews. Compared to other evaluation studies, our discussions heavily emphasize LLMs' limitations, aligning with the conservative view on the development of LLMs and the generative AI landscape, as is primarily adopted by social science researchers standing against the over-optimists and opportunists in engineering fields.

\subsection{Limitations and future directions}

Our study is limited and points to future research directions. Above all, our guided workflow is only one viable tool for assessing LLMs' processing of a particular type of text input, and our evaluation is by no means comprehensive. Alternative tasks, metrics, and analyses can be called for in other evaluation paradigms. 

(1) \textit{Text input/output length limits.} One major limitation of our analysis involves the incomplete input of scientific articles into the LLM due to its input length limit. To include more article content besides the introduction and conclusion sections, we can adopt the latest versions of some LLMs that support longer text input or consider multi-round dialogues to feed in the text by segments. The LLMs' output length is also capped; we currently query LLMs for structured output (keywords/abstracts/critiques), and querying LLMs for unstructured output potentially with a greater output size merits serious exploration.

(2) \textit{Drawbacks of evaluations.} For internal evaluation, metrics that assess the text's linguistic quality based on words can be outrun by metrics based on the text's semantic meaning, e.g., the ``semantic entropy'' \citep{FKet2024} developed in particular for evaluating the LLM output, which employs alternative LLMs to evaluate the output of a focal LLM. For external evaluation, text similarity metrics like BLEU and ROUGE may correlate weakly with human judgment: it is possible that two distinct texts for the same context, while both valid in human eyes, do not share any words in common and thus have zero similarity under these metrics \citep{LLet2016}. Overall, metric-based evaluations are rigid. We conducted a human evaluation to compensate for this deficiency, while its design is subject to elaborate improvements. Our recruited arbiter team is also less academically established than ordinary reviewers (e.g., faculty members).

(3) \textit{Broad empirical tests.} Different LLMs and academic texts from diverse disciplines can be adopted to replicate the analysis. Currently, we use those sections of articles where figures/tables/equations are rare; optical character recognition could allow non-text input with multimodal LLMs \citep{BT2024}. We used the articles from an interdisciplinary field (IS) to include content of different topics and writing styles; the LLM's performance may differ across disciplines with more specialized content. For example, LLMs may perform better in evaluating texts from CS conferences whose open reviews may have been utilized in training LLMs. 

(4) \textit{Enhancing the LLM output.} For first-order results, we focused on online, one-round, and zero-shot human-LLM dialogues with non-pretrained LLMs. LLMs' behavior evolves in offline, multi-round dialogues with few-shot prompting \citep{BMet2020}, and the LLM output may get enhanced through chain-of-thought \citep{WWet2022}, fine-tuning \citep{HSet2021}, contextual calibration \citep{ZWet2021}, text purification \citep{LUet2023}, or model merging \citep{ASet2024}, etc., besides the continued upgrade of the LLM itself due to fierce market competition. These techniques and advanced LLMs can be studied in future experiments. Further, the four tasks can be assembled into an integrated peer review where LLMs are asked to provide a structured response (Appendix H); this integrated LLM output can be studied closely.

(5) \textit{Common pitfalls of LLMs.} Studies have revealed that LLM output is subject to common pitfalls, some of which are inevitable \citep[e.g.,][]{XJet2024}, around the challenges in factuality \citep{ABet2024}. These can become their major obstacles to satisfactorily completing the tasks. As mentioned, these include hallucination \citep{JLet2023} (primarily at Task 1), stochasticity \citep{BGet2021} (primarily at Task 2), tempered objectivity \citep{DMet2024} (primarily at Task 3), non-understanding \citep{M2023} (primarily at Task 4), and deception \citep{H2024} (potentially at all tasks), and it is noted that a small factual lapse may drastically poison the scientific literature \citep{YXet2024} and that in general, using LLMs as judges may engender various types of biases \citep{ZCet2023}. Mapping the pitfalls of LLMs to their employment in different tasks during scientific research assistantship is an important direction.

(6) \textit{Reflexivity of LLMs.} We can investigate the self-reflection of LLMs \citep{Ba2023,RG2024} in our workflow: one may ask the LLM to (i) suggest tasks for evaluating its processing of academic texts and its potential in constructing peer reviews; (ii) generate prompts based on an original prompt and use the LLM-generated prompts; and (iii) explain the logic of its quality preference/quality score. These inquiries are essential for studying LLMs' capability of understanding. Further, we can utilize other LLMs to evaluate the output of the focal LLM (similar to \citet{FKet2024}), parallel to the human evaluation; however, as we showed that the LLM-led reviews are flawed, this self-judging may encounter a system error. 

(7) \textit{Related topics around AI-assisted peer review.} We focused on discussing the benefits of LLMs for input text processing and scientific peer review and didn't consider their abuse in misconducts around science production such as plagiarism \citep{QLet2023}, fake data \citep{TSet2023}, dummy review \citep{Br2024}, or predatory publishing \citep{Kd2024} that are severely detrimental to academic integrity. These problems call for serious discussions to prevent the misuse of LLMs in scientific research. \\

LLMs and generative AI \textit{en gros} are not cooking a free lunch for our future. For them to be put to responsible use, their values ought to align with ours \citep{NH2024}. If there are sacrifices we must make to embrace the hot air for the sky ride, science and the integrity of it should be among the last items that we throw out of the balloon basket. Surrendering scientific exploration and conversation to automatic agents is hardly a \textit{humain} decision; in the pursuit of truth, we can take nothing for granted.

\appendix
\setcounter{table}{0}
\setcounter{figure}{0}
\renewcommand{\thetable}{A\arabic{table}}
\renewcommand{\thefigure}{A\arabic{figure}}

\section*{Appendix A: Relevant studies on LLM evaluation}

Along with the rapidly growing research attention around LLMs, there has been abundant literature on the evaluation of LLMs that employ these four tasks. We summarize a list of key studies (Table \ref{tab:ref}) that focus on (i) assessing LLMs' text output, which primarily employs content reproduction (Task 1) and content reflection (Task 4), via metric-based evaluation or human evaluation; or (ii) assessing LLMs' capabilities as judges, which primarily employ content comparison (Task 2) and content scoring (Task 3)), via pair-wise (as opposed to list-wise, set-wise, etc. \citep{ZZet2024}) comparisons and/or direct or step-wise scoring that possibly lead to content ranking. The list is nonetheless by no means exhaustive, as ongoing research efforts continue to broaden the ground for evaluating the behavior of LLMs and more elaborate AI tools built around them. \\

\begin{table}[h!]
\centering
\resizebox{\textwidth}{!}
{
\begin{tabular*}{\textwidth}{@{\extracolsep{\fill}}lcccc}
\hline
&&&& \\
&
\shortstack{\textbf{Content}\\\textbf{Reproduction}} &
\shortstack{\textbf{Content}\\\textbf{Comparison}} &
\shortstack{\textbf{Content}\\\textbf{Scoring}} &
\shortstack{\textbf{Content}\\\textbf{Reflection}} \\
&&&& \\
\hline
&&&& \\
\citet{CSet2024} & \ding{51}  &  & \ding{51}  & \ding{51} \\
\citet{CWet2023} & & \ding{51}  & \ding{51}  & \\
\citet{DT2024} & & & \ding{51}  & \\
\citet{DWet2024} & \ding{51}  &  & \ding{51}  & \ding{51} \\
\citet{GBet2024} &  &  &  & \ding{51} \\
\citet{GRet2025} &  &  &  & \ding{51} \\
\citet{JZet2024} &  &  & \ding{51}  & \ding{51} \\
\citet{LCet2025} & \ding{51} &  &  & \ding{51} \\
\citet{LLet2025} & \ding{51} &  &  &  \\
\citet{LZet2024a} & &  &  & \ding{51} \\
\citet{LGet2024} & & \ding{51}  &  & \\
\citet{LZet2024b} & & \ding{51}  & \ding{51}  & \\
\citet{LS2023} & & \ding{51}  &  & \ding{51} \\
\citet{LMet2023} & & \ding{51} & \ding{51} & \\
\citet{QJet2023} & & \ding{51} & \ding{51} & \\
\citet{SCet2023} & & \ding{51} & \ding{51}  & \ding{51} \\
\citet{STet2025} & \ding{51}  &  & \ding{51}  & \ding{51} \\
\citet{TOet2024} & \ding{51}  &  &  &  \\
\citet{WZet2020} & \ding{51}  &  & \ding{51}  & \ding{51} \\
\citet{YDet2024} & \ding{51}  &  & \ding{51}  & \ding{51} \\
\citet{YLet2022} & \ding{51}  &  & \ding{51}  & \ding{51} \\
\citet{ZSet2024} & \ding{51}  &  & \ding{51}  & \ding{51} \\
\citet{ZTet2024} & & \ding{51} & & \\
\citet{ZCet2023} & & \ding{51} & \ding{51} & \\
\citet{ZCet2024} &  &  & \ding{51}  & \ding{51} \\
&&&& \\
\hline
\end{tabular*}
}
\caption{Relevant studies on LLM evaluation that employ similar tasks.}
\label{tab:ref}
\end{table}

\newpage

\addcontentsline{toc}{section}{Appendix B: Information on human arbiters}

\section*{Appendix B: Information on human arbiters}

\begin{table}[h]
\centering
\begin{tabular}{c|c|c|c|c|c|c|c}
\hline
\multicolumn{2}{c|}{\textbf{Academic status}} & \multicolumn{2}{c|}{\textbf{Area of study/research}} & \multicolumn{2}{c|}{\textbf{Academic institution}} & \multicolumn{2}{c}{\textbf{Language test score}} \\ \hline 
Year 0 phd & 2  & information systems & 18 & Xiamen U & 35 & IELTS & 39 \\
Year 1 phd & 41 & management (science) & 17 & Peking U & 18 & CET-6 & 38\\
Year 2 phd & 19 & CS/AI-related & 9 & City U of HK & 13 & TOEFL & 6 \\
Year 3 phd & 11 & (management) accounting & 7 & Northeast U (China) & 4 & GRE & 2\\
Year 4 phd & 9  & (corporate) finance & 6 & Tianjin U & 3 & PTE & 1 \\
Year 5 phd & 3  & (energy) economics & 6 & CUHK & 2 & not reported & 1 \\
Undergraduate & 1 & marketing & 6 & Fuzhou U & 2 & & \\
not reported & 1 & operations research-related & 4 & Renmin U & 2 & & \\
& & railway engineering & 1 & BUPT & 1 & & \\
& & technical innovation & 1 & Lanzhou U & 1 & & \\
& & other management fields & 8 & Southeast U (China) & 1 & & \\
& & not reported & 4 & U of Groningen & 1 & & \\
& &  &  & U of Queensland & 1 & & \\
& &  &  & U of Southern California & 1 & & \\
& &  &  & UST Beijing & 1 & & \\
& &  &  & Zhejiang U & 1 & & \\
 
\hline
\end{tabular}
\caption{Statistics of the information on human arbiters.}
\label{tab:human_arbiter}
\end{table}

\begin{table}[h]
\centering
\begin{tabular}{ccccccc}
\hline 
Score & 5 & 4 & 3 & 2 & 1 & All \\ \hline
No. & 4 & 9 & 34 & 35 & 5 & 87 \\
\hline
\end{tabular}
\caption{Human arbiters' self-reported familiarity with LLM outputs.}
\label{tab:familiarity_LLM}
\end{table}

\newpage

\addcontentsline{toc}{section}{Appendix C: Online interface for human evaluation}

\section*{Appendix C: Online interface for human evaluation}

\textit{Forewords and Instructions:}

Welcome to the evaluation of LLM (large language model)-generated texts. Please complete the assignment through this web page. Your input will be utilized for studying LLMs' performance in processing texts.

With sufficient domain knowledge in management studies, you will serve as an expert arbiter and grade the LLM-generated outputs (abstracts/critiques) for an input text which is an article published in a top journal in the field of information systems in management sciences (Information Systems Research/ISR or Journal of Management Information Systems/JMIS; both known for exhibiting interdisciplinary research across social, physical, and managerial sciences and supporting diverse scientific approaches in studying information technologies). The LLM used for generating the output is Google's Gemini Pro 1.0.

You will conduct two types of evaluations. In each instance of evaluation, please submit an integer from 1 (least) to 5 (most) to indicate the LLM output's quality. Non-integer or out-of-range scores will not pass. 

In Evaluation $\#$1, please compare an LLM-generated abstract to the ground-truth article abstract and submit a score to indicate the LLM output's \underline{reliability}. \textbf{You will read 12 such abstract pairs. The time limit is 3 minutes for each evaluation.}

In Evaluation $\#$2, please read the input article sections and the LLM-generated critiques on the article and submit a score to indicate the LLM output's \underline{logicalness}. \textbf{You will read 8 such article-critique pairs. The time limit is 8 minutes for each evaluation.}

There will be multiple breaks between evaluation sessions. The elapsed time will be displayed on the web page. \textbf{You will devote 120 minutes at maximum to the assignment.} Please make careful evaluations. \\

Thanks in advance for your participation. \\

[Pause for 2 minutes; then display ``Click to move on to the next page"]\\

\textit{Information of the Arbiter:}

Please indicate your English reading proficiency with language test scores (e.g., TOEFL 111, GRE 330+4):

Please indicate your current academic status (e.g., Year 2 PhD student, Year 4 undergraduate student):

Please indicate your area of study/research (e.g., information systems, marketing, management):

Please indicate your academic institution affiliation (e.g., CUHK, Arizona State University)

Please indicate your familiarity with LLM-generated content with a score from 1 (least) to 5 (most):

\textbf{Your input will not be publicized or used for commercial purposes.} 

[Pause for 1 minute; then display ``Click to start the evaluation"] \\

\textit{Evaluation $\#$1: LLM-generated article abstracts}

In the following evaluation, please compare an LLM-generated abstract to the ground-truth article abstract and submit a score from 1 (least) to 5 (most) to indicate the LLM output's \underline{reliability}. \textbf{You will read 12 such abstract pairs. The time limit is 3 minutes for each evaluation.}

[Materials 1 - 4] -- \textit{Break:} Please take a break for 1 minute. [Pause for 1 minute]

[Materials 5 - 8] -- \textit{Break:} Please take a break for 1 minute. [Pause for 1 minute]

[Materials 9 - 12] -- \textit{Break:} Please take a break for 1 minute. [Pause for 1 minute]\\

\textit{Evaluation $\#$2: LLM-generated article critiques}

In the following evaluation, please read the input article sections and the LLM-generated critiques on the article and submit a score from 1 (least) to 5 (most) to indicate the LLM output's \underline{logicalness}. \textbf{You will read 8 such article-critique pairs. The time limit is 8 minutes for each evaluation.}

[Materials 1 - 2] -- \textit{Break:} Please take a break for 1 minute. [Pause for 1 minute]

[Materials 3 - 4] -- \textit{Break:} Please take a break for 1 minute. [Pause for 1 minute]

[Materials 5 - 6] -- \textit{Break:} Please take a break for 1 minute. [Pause for 1 minute]

[Materials 7 - 8]\\

\textit{Afterwords:} 

Thank you very much for your participation. Your input is important to our study. 

Your payments will be processed shortly. Please help keep the contents confidential from a third party.

Your input has been successfully recorded. You can now close this webpage. 

\newpage

\addcontentsline{toc}{section}{Appendix D: Probabilities of correct Copeland scores}

\section*{Appendix D: Probabilities of correct Copeland scores}

Suppose the ground-truth outcome of comparing text $i$ and text $j$ is $y_{ij}$. The LLM-output preference $\hat{y}_{ij}$ contains error with probability $\epsilon$:
\begin{equation}
\hat{y}_{ij} = \left\{
\begin{aligned}
&y_{ij}, \ \ P = 1-\epsilon, \\
-&y_{ij}, \ \ P = \epsilon. \\
\end{aligned}
\right.
\end{equation}
Averaging the outcomes $\hat{y}_{ij}$ and $\hat{y}_{ji}$ from the two instances of comparison between $i$ and $j$, we obtain
\begin{equation}
\hat{z}_{ij} = \frac{\hat{y}_{ij}+(-\hat{y}_{ji})}{2},
\end{equation}
which constitutes the matrix $\mathbf{Z}=\{\hat{z}_{ij}\}$. Suppose the ground truth is $y_{ij}=-1$. There are four cases for the outcome at $\hat{z}_{ij}$: (i) when $\hat{y}_{ij}$ is false and $\hat{y}_{ji}$ is true, $\hat{z}_{ij} = 0$; (ii) when $\hat{y}_{ij}$ is true and $\hat{y}_{ji}$ is true, $\hat{z}_{ij} = -1$; (iii) when $\hat{y}_{ij}$ is false and $\hat{y}_{ji}$ is false, $\hat{z}_{ij} = 1$; (iv) when $\hat{y}_{ij}$ is true and $\hat{y}_{ji}$ is false, $\hat{z}_{ij} = 0$. Cases (i) and (iv) yield the same outcome; the probabilities for different values of $\hat{z}_{ij}$ are
\begin{equation}
\left\{
\begin{aligned}
&P(\hat{z}_{ij} = 0) &&= 2\epsilon(1-\epsilon), \\
&P(\hat{z}_{ij} = y_{ij}\ \text{[True outcome (TO)]}) &&= (1-\epsilon)^2, \\
&P(\hat{z}_{ij} = -y_{ij}\ \text{[Inverse outcome (IO)]}) &&= \epsilon^2. \\
\end{aligned}
\right.
\end{equation}
The Copeland ranking is based on the scores that sum on the matrix $\mathbf{Z}$:
\begin{equation}
S_{Copeland}(i) = \sum_j^{N-1} \hat{z}_{ij},
\end{equation}
where $N$ is the number of texts in the ensemble. We analyze the probability of having the correct score $S_{Copeland}(i)$.

First consider the text having the ground-truth ranking $\#1$. This text $R_1$ wins all comparisons: $\hat{y}_{R_1j} = 1$ for any $j\neq R_1$, and $S_{Copeland}(R_1) = N-1$. To realize this score $S_{Copeland}(R_1)$, all elements in the sum need to be true, which has a probability
\begin{equation}
P(R_1,\text{TO}) = \prod_{N-1}P(\hat{z}_{R_1j} = y_{R_1j}\ \text{[True outcome (TO)]}) = (1-\epsilon)^{2(N-1)}.
\end{equation}
Next consider the text having the ground-truth ranking $\#2$. This text $R_2$ wins all comparisons except the comparison with text $R_1$: $\hat{y}_{R_2R_1} = -1$, and $\hat{y}_{R_2j} = 1$ for any $j\neq R_1, j\neq R_2$, and $S_{Copeland}(R_1) = N-3$. To realize this score $S_{Copeland}(R_2)$, either (i) all elements in the sum are true, or (ii) $\hat{z}_{R_2R_1}$ and another $\hat{z}_{R_2j}$ for an arbitrary $j$ are false, and the rest $N-3$ elements are true. This has a probability
\begin{equation}
P(R_2,\text{TO}) = (1-\epsilon)^{2(N-1)} + (1-\epsilon)^{2(N-3)}C_1^1C_{N-2}^1\epsilon^{2\times2}.
\end{equation}
Similarly, for $S_{Copeland}(R_3)$ to be true, there are three possible cases, and their combined probability is 
\begin{equation}
P(R_3,\text{TO}) = (1-\epsilon)^{2(N-1)} + (1-\epsilon)^{2(N-3)}C_2^1C_{N-3}^1\epsilon^{2\times2} + (1-\epsilon)^{2(N-5)}C_2^2C_{N-3}^2\epsilon^{2\times4}.
\end{equation}
For $S_{Copeland}(R_m)$ to be true (text ranked $\#m$), there are $m$ possible cases, and the combined probability is 
\begin{equation}
\begin{aligned}
P(R_m,\text{TO}) = (1-\epsilon)^{2(N-1)} + (1-\epsilon)^{2(N-3)}C_{m-1}^1C_{N-m}^1\epsilon^{2\times2} + (1-\epsilon)^{2(N-5)}C_{m-1}^2C_{N-m}^2\epsilon^{2\times4} + ... \\
+ (1-\epsilon)^{2[N-1-2(m-1)]}C_{m-1}^{m-1}C_{N-m}^{m-1}\epsilon^{2\times2(m-1)}. 
\end{aligned}
\end{equation}
This ends halfway at the rank: $N-m \geq m-1 \Rightarrow m \leq (N+1)/2$. For the second half of the rank, $m > (N+1)/2$, the probabilities mirror the first half, with the substitution $m-1 \rightarrow N-m$. Then
\begin{equation}
\begin{aligned}
P(R_m,\text{TO}) = (1-\epsilon)^{2(N-1)} + (1-\epsilon)^{2(N-3)}C_{N-m}^1C_{m-1}^1\epsilon^{2\times2} + (1-\epsilon)^{2(N-5)}C_{N-m}^2C_{m-1}^2\epsilon^{2\times4} + ... \\
+ (1-\epsilon)^{2[N-1-2(N-m)]}C_{N-m}^{N-m}C_{m-1}^{N-m}\epsilon^{2\times2(N-m)},
\end{aligned}
\end{equation}
and, for example,
\begin{equation}
\begin{aligned}
P(R_{N-1},\text{TO}) &= P(R_2,\text{TO}) = (1-\epsilon)^{2(N-1)} + (1-\epsilon)^{2(N-3)}C_1^1C_{N-2}^1\epsilon^{2\times2}, \\
P(R_{N},\text{TO}) &= P(R_1,\text{TO}) = (1-\epsilon)^{2(N-1)}.
\end{aligned}
\end{equation}
To investigate the scalability of this text ranking through content comparison, we analyze the change of the probability of obtaining the correct score $S_{Copeland}(R_{\bullet})$, $P(R_{\bullet},\text{TO})$, as $N$ goes up. For the $\#1$ text $R_1$, 
\begin{equation}
\frac{dP(R_1,\text{TO})}{dN} = \frac{d(1-\epsilon)^{2(N-1)}}{dN} < 0;
\end{equation}
for all values of $\epsilon$, the probability of obtaining the correct score for text $R_1$ always decreases as $N$ goes up.

For the $\#2$ text $R_2$, 
\begin{equation}
\begin{aligned}
\frac{dP(R_2,\text{TO})}{dN} &= \frac{d(1-\epsilon)^{2(N-1)} + (1-\epsilon)^{2(N-3)}C_1^1C_{N-2}^1\epsilon^{2\times2}}{dN} \\
&= (1-\epsilon)^{2(N-3)}\{\epsilon^4 + 2ln(1-\epsilon)[(1-\epsilon)^4+(N-2)\epsilon^4]\}. \\
\end{aligned}
\end{equation}
Then
\begin{equation}
\begin{aligned}
\frac{dP(R_2,\text{TO})}{dN} > 0 \Rightarrow & \epsilon^4 + 2ln(1-\epsilon)[(1-\epsilon)^4+(N-2)\epsilon^4] > 0 \\
\Rightarrow & N - 2 < -\frac{1}{2ln(1-\epsilon)} - (\frac{1-\epsilon}{\epsilon})^4 < 0, \ \ \text{for $\epsilon \in [0,1)$}.
\end{aligned}
\end{equation}
Such a condition cannot be satisfied when $N$ is positive, which means that $dP(R_2,\text{TO})/dN < 0$ uniformly for all $\epsilon$, same as the $R_1$ case.

Similar derivations establish that $dP(R_m,\text{TO})/dN < 0$ for all $\epsilon$ and $m$, which can be visualized with simulation results (Figure 7a). This means that for a text of any ground-truth ranking, the probability of obtaining its correct score $S_{Copeland}$ decreases uniformly as $N$ increases, i.e., the content comparison is \textit{not scalable}, in the sense of when using it to construct the text ranking.

The output $S_{Copeland}$ list (Figure 6b) does not follow the perfect sequence $N-1$, $N-3$, $N-5$, ..., $1-N$. The deviation from this perfect sequence can be used to measure the loss in scalability. We use the subset of $\mathbf{Z}$ to compute $S_{Copeland}$ and such a deviation, and show this scalability loss as $N$ increases (Figure 7b).

Results show that as the number of texts $N$ increases, $P(R_{m},\text{TO})$ decreases for all $\epsilon$ and $m$. Note that $P(R_{\bullet},TO)$ approaches $P(R_{1},TO)$ when $\epsilon$ is small, as the terms with the $\epsilon$ multipliers vanish; the difference between $P(R_{1},TO)$ and $P(R_{2},TO)$ can be seen, e.g., at $\epsilon=0.5$.

\newpage

\addcontentsline{toc}{section}{Appendix E: Grouped Top-25 critique subjects generated by the LLM}

\section*{Appendix E: Grouped top-25 LLM-generated critique subjects}

{
\setlength{\baselineskip}{9pt}

(1) lack of empirical evidence/limited empirical evidence/insufficient empirical evidence/limited empirical validation/lack of empirical evaluation/lack of empirical data/lack of empirical support/lack of real-world data/insufficient evidence/lack of evidence/lack of supporting evidence

(2) limited generalizability/generalizability concerns/lack of generalizability/of findings/of limitations/of results/limited generalizability to other contexts/of findings/limited generalizability discussion/limited generalizability discussed only in conclusion

(3) narrow focus/scope limitations/narrow scope/limited scope of the study/of conclusion/of analysis

(4) limited discussion of limitations/insufficient discussion of limitations/incomplete discussion of limitations/limited discussion of limitations and future research directions/lack of discussion on the limitations of the study/limited discussion of limitations impact

(5) lack of a clear research question/of specific/explicit research question/lack of a clear statement of the research question/lack of clear research questions and hypotheses/lack of explicit research hypotheses

(6) limited discussion of prior research/of existing research/of related work/lack of literature review/insufficient literature review/lack of citations/limited discussion of existing literature/limited discussion of existing methods/lack of a comprehensive literature review/insufficient discussion of prior research

(7) limited theoretical foundation/lack of theoretical foundation/lack of theoretical grounding/insufficient explanation of the theoretical framework/limited discussion of methodology/lack of clear theoretical contribution in introduction/lack of clear theoretical contribution in conclusion

(8) limited practical implications/lack of practical implications/absence of practical implications/lack of discussion on the broader implications of the findings/insufficient discussion of practical implications/lack of clarity on practical implications/insufficient practical implications/lack of discussion on practical implications

(9) limited discussion of alternative explanations/alternative perspectives/limited exploration of alternative explanations/lack of consideration for alternative explanations

(10) lack of clarity in the research question/lack of clear research question/lack of specificity in research questions/lack of clarity in problem definition

(11) limited discussion of ethical considerations/insufficient consideration of ethical implications/ethical concerns/lack of discussion on the ethical implications/limited consideration of ethical implications

(12) overgeneralization/overgeneralization of findings/overgeneralized introduction

(13) oversimplification/overly simplistic model

(14) methodological limitations/insufficient detail on methodology

(15) lack of contextualization/lack of real-world context/lack of context

(16) overemphasis on novelty/overstated novelty

(17) incomplete conclusion/overly broad conclusion

(18) overly broad introduction

(19) lack of specific examples/case studies

(20) lack of clear research gap

(21) lack of clarity in the introduction

(22) overreliance on previous studies/overreliance on previous citations

(23) confounding factors/confounding variables/potential for unobserved confounding factors

(24) limited external validity/overreliance on self-reported data

(25) repetitive and unfocused conclusion
}

\newpage

\addcontentsline{toc}{section}{Appendix F: Details of human evaluation results}

\section*{Appendix F: Details of human evaluation results}

We check whether human arbiters' background characteristics influence their scoring. We employ ordinary least squares (OLS) regression, incorporating HC3-robust standard errors to ensure valid inference under heteroskedasticity, i.e., non-constant error variance across covariates. We consider four independent variables: arbiters' academic status, area of study, familiarity with LLMs, and English language proficiency. We check if these characteristics can predict arbiters' delivered $HE^{abs}$ and $HE^{cri}$. Separate OLS models are estimated for $HE^{abs}$ and $HE^{cri}$. We further control the average time of the arbiters' evaluation.

The background characteristics are pre-processed as follows:

\begin{itemize}

\item \textit{Academic status.} We consider the number of years (0-5) each arbiter has been enrolled in their Ph.D. program. We drop the one undergraduate and the one not-reported.

\item \textit{Area of study.} Arbiters' self-reported areas of study/research are coded into six categories: Economics (n=7), Finance \& Accounting (n=13), Information Systems (IS) \& Computer Sciences (CS) (n=27), Management \& Marketing (n=28), Operations Research \& Supply Chain Management (SCM) (n=5), and Others (n=7). IS \& CS is used as the reference category for the area of study.

\item \textit{Familiarity with LLMs.} No pre-processing. 

\item \textit{English language proficiency.} We use the TOEFL score as the indicator for English language proficiency. IELTS scores are converted to their TOEFL score equivalents using the official ETS concordance table\footnote{https://www.ets.org/toefl/institutions/ibt/compare-scores.html}. College English Test 6 (CET-6) scores are first translated to equivalent IELTS scores based on the guidelines from Birmingham City University's qualification table\footnote{https://www.bcu.ac.uk/international/your-application/english-language-and-english-tests/accepted-qualifications} and then converted to TOEFL scores. GRE/PTE/not-reported are dropped.

\end{itemize}

Table \ref{tab:regression_results} shows the results. Overall, arbiters' background characteristics do not demonstrate statistically significant associations with their scoring.

\begin{table}[htbp]
\centering
\begin{tabular}{l cc cc}
\hline
& \multicolumn{2}{c}{$HE^{abs}$} & \multicolumn{2}{c}{$HE^{cri}$} \\
\cmidrule(lr){2-3} \cmidrule(lr){4-5}
Variable & Coefficient & (Std. Error) & Coefficient & (Std. Error) \\
\hline
Intercept                                       & $3.278^{***}$ & $(0.545)$ & $4.424^{***}$ & $(0.952)$ \\
\addlinespace 
\textit{Area of Study (Reference: IS \& CS)} & & & & \\
\quad Economics                                 & $-0.026$      & $(0.256)$ & $-0.037$      & $(0.262)$ \\
\quad Finance \& Accounting                     & $-0.104$      & $(0.170)$ & $-0.216$      & $(0.182)$ \\
\quad Management \& Marketing                    & $-0.045$      & $(0.154)$ & $0.141$       & $(0.170)$ \\
\quad Operations Research \& SCM                & $-0.127$      & $(0.175)$ & $-0.172$      & $(0.323)$ \\
\quad Other Area                                & $-0.151$      & $(0.251)$ & $-0.345$      & $(0.227)$ \\
\addlinespace 
Program Seniority (Status)                      & $-0.036$      & $(0.049)$ & $-0.036$      & $(0.050)$ \\
Log(Time Spent)                                 & $0.018$       & $(0.097)$ & $-0.133$      & $(0.162)$ \\
Familiarity with LLMs                           & $0.026$       & $(0.058)$ & $0.050$       & $(0.074)$ \\
English Proficiency (TOEFL equiv.)              & $0.002$       & $(0.003)$ & $-0.006^{*}$  & $(0.003)$ \\
\hline
\addlinespace 
R-squared                                       & \multicolumn{2}{c}{0.035} & \multicolumn{2}{c}{0.129} \\
Adjusted R-squared                              & \multicolumn{2}{c}{-0.078} & \multicolumn{2}{c}{0.027} \\
F-statistic                                     & \multicolumn{2}{c}{0.350}  & \multicolumn{2}{c}{1.423} \\
Prob (F-statistic)                              & \multicolumn{2}{c}{0.955}  & \multicolumn{2}{c}{0.193} \\
Observations                                    & \multicolumn{2}{c}{87}     & \multicolumn{2}{c}{87} \\
\addlinespace 
\hline
\multicolumn{5}{l}{\textit{Notes:} Standard errors (HC3-robust) are in parentheses. $^{\dagger} p < 0.10$, $^{*} p < 0.05$, $^{**} p < 0.01$, $^{***} p < 0.001$.} \\
\end{tabular}
\caption{OLS Regression: Arbiters' background characteristics on their delivered $HE^{abs}$ and $HE^{cri}$.}
\label{tab:regression_results}
\end{table}

\newpage

\addcontentsline{toc}{section}{Appendix G: Correlation between LLM-reliant and ground-truth text metrics}

\section*{Appendix G: LLM-reliant/ground-truth text metrics correlations}

We calculate Pearson's $r$ (Figure 15 in the main text) between LLM-reliant and ground-truth text quality metrics and Spearman's $\rho$ and Kendall's $\tau$ between their ranks (Tables A5, A6).

\textit{LLM-reliant text quality metrics:} at Task 1, the $IE_{\bullet}^{abs}$/$EE_{\bullet}^{key/abs}$/$HE^{abs}$ from internal/external/human evaluation of LLM-generated keywords/abstracts; at Task 2, the Copeland score $S_{Copeland}$; at Task 3, the LLM-output text score $\hat{w}$; at Task 4, the $IE_{\bullet}^{cri}$/$EE_{\bullet}^{cri}$/$HE^{cri}$ from internal/external/human evaluation of LLM-generated critiques (22 in total). \textit{Ground-truth text quality metrics:} for ISR and MS articles, acceptance time and download count (Table A5); for JMIS articles, view count and citation count (Table A6).

\begin{table}[h]
    \centering
    \resizebox{0.68\textwidth}{!}
    {
    \begin{tabular}{lcccccc}
        \toprule
        & \multicolumn{3}{c}{acceptance time} & \multicolumn{3}{c}{download count} \\
        \cmidrule(lr){2-4} \cmidrule(lr){5-7}
        & Pearson’s $r$ & Spearman's $\rho$ & Kendall's $\tau$ & Pearson’s $r$ & Spearman's $\rho$ & Kendall's $\tau$ \\
        \midrule
        $EE_{Jac}^{key}$ & 0.1264 & 0.1281 & 0.0862 & 0.1448 & 0.2822 & 0.1884 \\
        $EE_{cos}^{key}$ & 0.0702 & 0.1152 & 0.0802 & \textbf{0.1751}* & \textbf{0.2780}* & \textbf{0.1934}*  \\
        $IE_{H-density}^{abs}$ & \textbf{-0.1694}* & \textbf{-0.1837}* & \textbf{-0.1193}* &-0.0538 & -0.0471 & -0.0307 \\
        $IE_{entropy}^{abs}$ & -0.0498 & -0.0841 & -0.0573 & \textbf{-0.1930}* & \textbf{-0.1542}* & \textbf{-0.1064}* \\
        $IE_{TTR}^{abs}$ & -0.0932 & -0.0877 & -0.0629 & 0.0961 & -0.1089 & -0.0748 \\
        $IE_{FK}^{abs}$ & \textbf{0.1872}* & \textbf{0.1792}* & \textbf{0.1240}* & 0.0580 & 0.0149 & 0.0128 \\
        $EE_{Jac}^{abs}$ & 0.1271 & 0.1386 & 0.0916 & -0.1476 & -0.0178 & -0.0159 \\
        $EE_{cos}^{abs}$ & 0.1383 & 0.1422 & 0.0966 & -0.0551 & 0.1636 & 0.1104 \\ 
        $EE_{BLEU}^{abs}$ & 0.1426 & 0.1162 & 0.0814 & -0.0663 & -0.0195 & -0.0103 \\
        $EE_{ROUGE-L}^{abs}$ & 0.1503 & 0.1478 & 0.0938 & \textbf{-0.1829} & \textbf{-0.0552}* & \textbf{-0.0379}* \\
        $S_{Copeland}$ & 0.1063 & 0.0566 & 00.0361 & 0.0619 & 0.0828 & 0.0542 \\
        $\hat{w}$ & -0.0296 & -0.0444 & -0.0274 & \textbf{0.2357}** & \textbf{0.1731}** & \textbf{0.1243}** \\
        $IE_{H-density}^{cri}$ & -0.0295 & -0.0429 & -0.0279 & -0.1292 & -0.1215 & -0.0823 \\
        $IE_{entropy}^{cri}$ & -0.1347 & -0.0836 & -0.0591 & \textbf{-0.1764}* & \textbf{-0.2519}* & \textbf{-0.1728}* \\
        $IE_{TTR}^{cri}$ & -0.0773 & -0.0826 & -0.0552 & 0.0542 & -0.1173 & -0.0780 \\
        $IE_{FK}^{cri}$ & 0.1217 & 0.1287 & 0.0870 & 0.0848 & 0.1473 & 0.0992\\
        $EE_{Jac}^{cri}$ & -0.0929 & -0.1339 & -0.080 & -0.0570 & -0.0220 & -0.0133 \\
        $EE_{cos}^{cri}$ & 0.0380 & 0.0154 & 0.0120 & 0.0060 & 0.0242 & 0.0169 \\
        $EE_{BLEU}^{cri}$ & \textbf{-0.2029}** & \textbf{-0.1624}** & \textbf{-0.1033}** & -0.0419 & -0.1604 & -0.1073 \\
        $EE_{ROUGE-L}^{cri}$ & -0.0761 & -0.1172 & -0.0747 & -0.0697 & -0.0408 & -0.0231 \\
        $HE^{abs}$ & 0.1357 & 0.1059 & 0.0747 & -0.0669 & 0.0183 & 0.0154 \\
        $HE^{cri}$ & -0.0442 & 0.0088 & 0.0069 & -0.0764 & -0.1071 & -0.0732 \\
        \bottomrule
    \end{tabular}
    }
    \caption{Correlation between LLM-reliant and ground-truth text metrics (for ISR and MS articles).}
    \label{tab:correlations_transposed1}
\end{table}

\begin{table}[h]
    \centering
    \resizebox{0.68\textwidth}{!}
    {
    \begin{tabular}{lcccccc}
        \toprule
        & \multicolumn{3}{c}{view count} & \multicolumn{3}{c}{citation count} \\
        \cmidrule(lr){2-4} \cmidrule(lr){5-7}
        & Pearson’s $r$ & Spearman's $\rho$ & Kendall's $\tau$ & Pearson’s $r$ & Spearman's $\rho$ & Kendall's $\tau$ \\
        \midrule
        $EE_{Jac}^{key}$ & 0.1071 & 0.0949 & 0.0677 & 0.0674 & -0.0127 & -0.0089 \\
        $EE_{cos}^{key}$ & 0.0213 & 0.0480 & 0.0408 & 0.1836 & 0.2175 & 0.1684 \\
        $IE_{H-density}^{abs}$ & -0.0115 & -0.1637 & -0.1073 & -0.0777 & -0.1483 & -0.1073 \\
        $IE_{entropy}^{abs}$ & -0.1922 & -0.1634 & -0.1200 & -0.0601 & -0.1852 & -0.1494 \\
        $IE_{TTR}^{abs}$ & -0.1502 & -0.1244 & -0.0896 & -0.0857 & -0.2141 & -0.1717 \\
        $IE_{FK}^{abs}$ & 0.0049 & -0.0645 & -0.0472 & 0.0333 & 0.0770 & 0.0553 \\
        $EE_{Jac}^{abs}$ & 0.0788 & 0.0963 & 0.0756 & -0.1969 & -0.1584 & -0.1246 \\
        $EE_{cos}^{abs}$ & 0.0733 & 0.0782 & 0.0548 & 0.0869 & 0.1970 & 0.1477 \\
        $EE_{BLEU}^{abs}$ & 0.0290 & 0.0513 & 0.0358 & -0.0724 & -0.0464 & -0.0363 \\
        $EE_{ROUGE-L}^{abs}$ & 0.0448 & 0.0279 & 0.0364 & -0.2031 & -0.1065 & -0.0792 \\
        $S_{Copeland}$ & -0.1095 & -0.0706 & -0.0456 & -0.1318 & -0.1531 & -0.1234 \\
        $\hat{w}$ & 0.0078 & 0.0183 & 0.0151 & -0.1412 & -0.1914 & -0.1587 \\
        $IE_{H-density}^{cri}$ & -0.1498 & -0.2141 & -0.1485 & -0.0743 & -0.1207 & -0.0946 \\
        $IE_{entropy}^{cri}$ & -0.1516 & -0.1340 & -0.0965 & 0.0011 & -0.0319 & -0.0289 \\
        $IE_{TTR}^{cri}$ & \textbf{-0.2559}* & \textbf{0.0096}* & \textbf{0.0054}* & -0.1199 & 0.0257 & 0.0165 \\
        $IE_{FK}^{cri}$ & \textbf{0.2997}** & \textbf{0.1557}** & \textbf{0.1073}** & 0.1203 & 0.1752 & 0.1378 \\
        $EE_{Jac}^{cri}$ & -0.1650 & -0.1916 & -0.1149 & \textbf{-0.2206}* & \textbf{-0.1857}* & \textbf{-0.1411}* \\
        $EE_{cos}^{cri}$ & -0.0712 & -0.0315 & -0.0250 & -0.0594 & 0.0402 & 0.0355 \\
        $EE_{BLEU}^{cri}$ & 0.0729 & 0.0192 & 0.0092 & -0.0299 & -0.0856 & -0.0652 \\
        $EE_{ROUGE-L}^{cri}$ & -0.1885 & -0.2319 & -0.1491 & \textbf{-0.2920}** & \textbf{-0.2984}** & \textbf{-0.2319}* \\
        $HE^{abs}$ & \textbf{-0.2979}* & \textbf{-0.2477}** & \textbf{-0.1793}* & -0.1733 & -0.2544 & -0.1973 \\
        $HE^{cri}$ & -0.0113 & -0.0467 & -0.0372 & -0.0921 & -0.1432 & -0.1161 \\
        \bottomrule
    \end{tabular}
    }
    \caption{Correlation between LLM-reliant and ground-truth text metrics (for JMIS articles).}
    \label{tab:correlations_transposed2}
\end{table}

Except for rare cases, the correlations from the three measures agree with one another, and the cases of significance concur. Overall, LLM-reliant metrics demonstrate rare correlation or anti-correlation with ground-truth metrics, measured by Pearson's $r$, Spearman's $\rho$, or Kendall's $\tau$.

\newpage

\addcontentsline{toc}{section}{Appendix H: Integrated prompt for peer review}

\section*{Appendix H: Integrated prompt for peer review}

We can aggregate the four tasks in a holistic workflow and construct an integrated prompt for the peer review of academic manuscripts (Figure A1). The prompt consists of four parts: role assumption (instruction prompt specifying the LLM role of a professional reviewer), text input (data prompt), task specification (instruction prompt specifying peer review deliverables), and response collection (instruction prompt soliciting the peer review output). The four tasks (in red) constitute the body of the instruction prompt; elements of the investigated robustness checks (in blue) are embodied in text input and task specification.

\begin{figure}[h!]
    \centering
    \includegraphics[width=\linewidth]{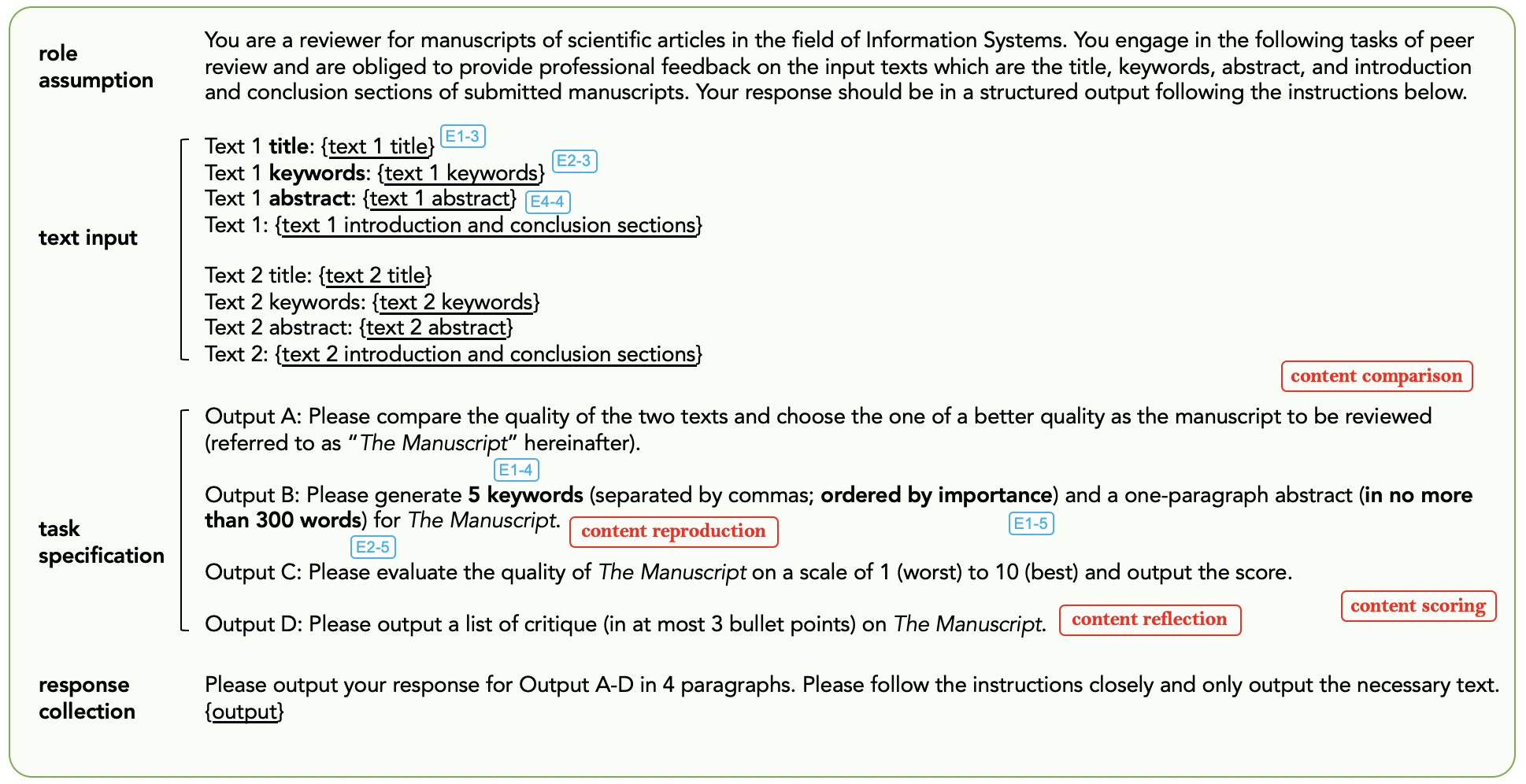}
    \caption{Integrated prompt for the peer review of academic manuscripts.}
    \label{fig:integrated prompt}
\end{figure}

We can ask human arbiters to evaluate the LLM response in this structured peer review. For example, human arbiters can answer four questions regarding Outputs A-D of the LLM response and submit four scores: 

At Output A/B/C/D, to what extent does the 

(1) LLM-output text preference indicate a good content comparison?

(2) LLM-generated text summary indicate a good content reproduction?

(3) LLM-output text grade indicate a good content scoring?

(4) LLM-generated text critique indicate a good content reflection? \\

\newpage


\begin{thebibliography}{26}

\bibitem[{\textit{Abbasi and Chen}(2008)}]{AC2008}
Abbasi, A., $\&$ Chen, H. (2008), CyberGate: A design framework and system for text analysis of computer-mediated communication, \textit{MIS Quarterly}, 811-837.

\bibitem[{\textit{Abbasi et al.}(2024)}]{APet2024}
Abbasi, A., Parsons, J., Pant, G., Sheng, O. R. L., $\&$ Sarker, S. (2024), Pathways for Design Research on Artificial Intelligence, \textit{Information Systems Research}.

\bibitem[{\textit{Agrawal et al.}(2024)}]{AMet2024}
Agrawal, A., McHale, J., $\&$ Oettl, A. (2024), Artificial intelligence and scientific discovery: A model of prioritized search, \textit{Research Policy}, 53(5), 104989.

\bibitem[{\textit{Akiba et al.}(2024)}]{ASet2024}
Akiba, T., Shing, M., Tang, Y., Sun, Q., $\&$ Ha, D. (2024), Evolutionary optimization of model merging recipes, arXiv:2403.13187.

\bibitem[{\textit{Allen}(2025)}]{A2025}
Allen, K. (2025), Measure the deeds that
make academic life fulfilling, \textit{Nature}, 638, 861.

\bibitem[{\textit{Alvarez et al.}(2024)}]{ACet2024}
Alvarez, A., Caliskan, A., Crockett, M. J., Ho, S. S., Messeri, L., $\&$ West, J. (2024), Science communication with generative AI, \textit{Nature Human Behaviour}, 1-3.

\bibitem[{\textit{Andersen}(2020)}]{A2020}
Andersen, M. (2020), Easing the burden of peer review, \textit{Nature Astronomy}, 4(7), 646-647.

\bibitem[{\textit{Angelopoulos et al.}(2024)}]{ACAet2024}
Angelopoulos, A., Cahoon, J. F., $\&$ Altrovitz, R. (2024), Transforming science labs into automated factories of discovery, \textit{Science Robotics}, 9(95).

\bibitem[{\textit{Aria and Cuccurullo}(2017)}]{AC2017}
Aria, M., $\&$ Cuccurullo, C. (2017), bibliometrix: An R-tool for comprehensive science mapping analysis, \textit{Journal of Informetrics}, 11(4), 959-975.

\bibitem[{\textit{Arora et al.}(2024)}]{ACNet2024}
Arora, N., Chakraborty, I., $\&$ Nishimura, Y. (2024), AI-Human Hybrids for Marketing Research: Leveraging LLMs as Collaborators, \textit{Journal of Marketing}, 00222429241276529.

\bibitem[{\textit{Augenstein et al.}(2024)}]{ABet2024}
Augenstein, I., Baldwin, T., Cha, M., Chakraborty, T., Ciampaglia, G. L., Corney, D., ... $\&$ Zagni, G. (2024), Factuality challenges in the era of large language models and opportunities for fact-checking, \textit{Nature Machine Intelligence}, 1-12.

\bibitem[{\textit{Balmer}(2023)}]{Ba2023}
Balmer, A. (2023), A sociological conversation with ChatGPT about AI ethics, affect and reflexivity, \textit{Sociology}, 57(5), 1249-1258.

\bibitem[{\textit{Benbasat and Weber}(1996)}]{BW1996}
Benbasat, I., $\&$ Weber, R. (1996), Research commentary: Rethinking “diversity” in information systems research, \textit{Information Systems Research}, 7(4), 389-399.

\bibitem[{\textit{Bender et al.}(2021)}]{BGet2021}
Bender, E. M., Gebru, T., McMillan-Major, A., $\&$ Shmitchell, S. (2021, March). On the dangers of stochastic parrots: Can language models be too big? In \textit{Proceedings of the 2021 ACM Conference on Fairness, Accountability, and Transparency} (pp. 610-623).

\bibitem[{\textit{Bhargava et al.}(2025)}]{BBet2025}
Bhargava, H. K., Brown, S., Ghose, A., Gupta, A., Leidner, D., $\&$ Wu, D. J. (2025), Exploring generative AI’s impact on research: Perspectives from senior scholars in management information systems, \textit{ACM Transactions on Management Information Systems}, 16(2), 1-9.

\bibitem[{\textit{Binz et al.}(2025)}]{BAet2025}
Binz, M., Alaniz, S., Roskies, A., Aczel, B., Bergstrom, C. T., Allen, C., ... $\&$ Schulz, E. (2025), How should the advancement of large language models affect the practice of science? \textit{Proceedings of the National Academy of Sciences}, 122(5), e2401227121.

\bibitem[{\textit{Biswas and Talukdar}(2024)}]{BT2024}
Biswas, A., $\&$ Talukdar, W. (2024), Robustness of Structured Data Extraction from In-Plane Rotated Documents Using Multi-Modal Large Language Models (LLM), \textit{Journal of Artificial Intelligence Research}.


\bibitem[{\textit{Bornmann et al.}(2021)}]{BHet2021}
Bornmann, L., Haunschild, R., $\&$ Mutz, R. (2021), Growth rates of modern science: a latent piecewise growth curve approach to model publication numbers from established and new literature databases, \textit{Humanities and Social Sciences Communications}, 8(1), 1-15.

\bibitem[{\textit{Brainard}(2023)}]{B2023}
Brainard, J. (2023), Can AI help scientists surf a paper flood? \textit{Science}, 382(6673), 866-867.

\bibitem[{\textit{Brainard}(2024)}]{Br2024}
Brainard, J. (2024), Duplicated phrases in peer review draw scrutiny, \textit{Science}, 385(6714), 1150.

\bibitem[{\textit{Brown}(2024)}]{B2024}
Brown, S. (2024), Community Building through Virtuous Reviewing, \textit{MIS Quarterly}, 48(2).

\bibitem[{\textit{Brown et al.}(2020)}]{BMet2020}
Brown, T., Mann, B., Ryder, N., Subbiah, M., Kaplan, J. D., Dhariwal, P., ... $\&$ Amodei, D. (2020), Language models are few-shot learners, \textit{Advances in Neural Information Processing Systems}, 33, 1877-1901.

\bibitem[{\textit{Callaway}(2023)}]{C2023}
Callaway, E. (2023), AI writes summaries of preprints in bioRxiv trial, \textit{Nature}.

\bibitem[{\textit{Caro et al.}(2022)}]{CCet2022}
Caro, F., Colliard, J. E., Katok, E., Ockenfels, A., Stier-Moses, N., Tucker, C., $\&$ Wu, D. J. (2022). Management Science Special Issue on the Human-Algorithm Connection, \textit{Management Science}, 68(1), 7-8.

\bibitem[{\textit{Chang and Woo}(1994)}]{CW1994}
Chang, M. K., $\&$ Woo, C. C. (1994), A speech-act-based negotiation protocol: design, implementation, and test use, \textit{ACM Transactions on Information Systems} (TOIS), 12(4), 360-382.

\bibitem[{\textit{Checco et al.}(2021)}]{CBet2021}
Checco, A., Bracciale, L., Loreti, P., Pinfield, S., $\&$ Bianchi, G. (2021), AI-assisted peer review, \textit{Humanities and Social Sciences Communications}, 8(1), 1-11.

\bibitem[{\textit{Chen and Chan}(2024)}]{CC2024}
Chen, Z., $\&$ Chan, J. (2024), Large language model in creative work: The role of collaboration modality and user expertise, \textit{Management Science}.

\bibitem[{\textit{Copeland}(1951)}]{C1951}
Copeland, A. H. (1951), A reasonable social welfare function, In \textit{University of Michigan Seminar on Applications of Mathematics to the Social Sciences}.

\bibitem[{\textit{Crabtree}(2020)}]{C2020}
Crabtree, G. (2020), Self-driving laboratories coming of age, \textit{Joule}, 4(12), 2538-2541.

\bibitem[{\textit{Dai et al.}(2024)}]{DVet2024}
Dai, T., Vijayakrishnan, S., Szczypiński, F. T., Ayme, J. F., Simaei, E., Fellowes, T., ... $\&$ Cooper, A. I. (2024), Autonomous mobile robots for exploratory synthetic chemistry, \textit{Nature}, 1-8.

\bibitem[{\textit{D'Arcy et al.}(2024)}]{DHet2024}
D'Arcy, M., Hope, T., Birnbaum, L., $\&$ Downey, D. (2024), Marg: Multi-agent review generation for scientific papers, arXiv:2401.04259.

\bibitem[{\textit{Denison et al.}(2024)}]{DMet2024}
Denison, C.E., MacDiarmid, M.S., Barez, F., Duvenaud, D.K., Kravec, S., Marks, S., ... $\&$ Hubinger, E. (2024), Sycophancy to Subterfuge: Investigating Reward-Tampering in Large Language Models, arXiv:2406.10162.

\bibitem[{\textit{De Regt and Dieks}(2005)}]{DD2005}
De Regt, H. W., $\&$ Dieks, D. (2005), A contextual approach to scientific understanding, \textit{Synthese}, 144, 137-170.

\bibitem[{\textit{Drori and Te'eni}(2024)}]{DT2024}
Drori, I., $\&$ Te'eni, D. (2024), Human-in-the-Loop AI Reviewing: Feasibility, Opportunities, and Risks, \textit{Journal of the Association for Information Systems}, 25(1), 98-109.

\bibitem[{\textit{Du et al.}(2024)}]{DWet2024}
Du, J., Wang, Y., Zhao, W., Deng, Z., Liu, S., Lou, R., ... $\&$ Yin, W. (2024), Llms assist nlp researchers: Critique paper (meta-) reviewing, arXiv:2406.16253.

\bibitem[{\textit{Eloundou et al.}(2024)}]{EMet2024}
Eloundou, T., Manning, S., Mishkin, P., $\&$ Rock, D. (2024), GPTs are GPTs: Labor market impact potential of LLMs, \textit{Science}, 384, 1306-1308.

\bibitem[{\textit{Extance}(2018)}]{E2018}
Extance, A. (2018), How AI technology can tame the scientific literature, \textit{Nature}, 561(7722), 273-275.

\bibitem[{\textit{Fang et al.}(2024)}]{FCet2024}
Fang, X., Che, S., Mao, M., Zhang, H., Zhao, M., $\&$ Zhao, X. (2024), Bias of AI-generated content: an examination of news produced by large language models, \textit{Scientific Reports}, 14(1), 5224.

\bibitem[{\textit{Farquhar et al.}(2024)}]{FKet2024}
Farquhar, S., Kossen, J., Kuhn, L., $\&$ Gal, Y. (2024), Detecting hallucinations in large language models using semantic entropy, \textit{Nature}, 630(8017), 625-630.

\bibitem[{\textit{Frieder et al.}(2024)}]{FPet2024}
Frieder, S., Pinchetti, L., Griffiths, R. R., Salvatori, T., Lukasiewicz, T., Petersen, P., $\&$ Berner, J. (2024), Mathematical capabilities of chatgpt, \textit{Advances in Neural Information Processing Systems}, 36.

\bibitem[{\textit{Frith}(2020)}]{F2020}
Frith, U. (2020), Fast lane to slow science, \textit{Trends in Cognitive Sciences}, 24(1), 1-2.

\bibitem[{\textit{Gao et al.}(2024)}]{GLet2024}
Gao, Y., Lee, D., Burtch, G., $\&$ Fazelpour, S. (2024), Take Caution in Using LLMs as Human Surrogates: Scylla Ex Machina, arXiv:2410.19599.

\bibitem[{\textit{Gibney}(2024a)}]{Gia2024}
Gibney, E. (2024), Not all "open source" AI models are actually open: here's a ranking, \textit{Nature}.

\bibitem[{\textit{Gibney}(2024b)}]{Gib2024}
Gibney, E. (2024), Has your paper been used to train an AI model? Almost certainly, \textit{Nature}, 632, 715-716.

\bibitem[{\textit{Gil et al.}(2014)}]{GGet2014}
Gil, Y., Greaves, M., Hendler, J., $\&$ Hirsh, H. (2014), Amplify scientific discovery with artificial intelligence, \textit{Science}, 346(6206), 171-172.

\bibitem[{\textit{Gneiting et al.}(2007)}]{GBet2007}
Gneiting, T., Balabdaoui, F., $\&$ Raftery, A. E. (2007), Probabilistic forecasts, calibration and sharpness, \textit{Journal of the Royal Statistical Society Series B: Statistical Methodology}, 69(2), 243-268.

\bibitem[{\textit{Gregor and Hevner}(2013)}]{GnH2013}
Gregor, S., $\&$ Hevner, A. R. (2013), Positioning and presenting design science research for maximum impact, \textit{MIS Quarterly}, 337-355.

\bibitem[{\textit{Gruda}(2024)}]{G2024}
Gruda, D. (2024), Three ways ChatGPT helps me in my academic writing, \textit{Nature}.

\bibitem[{\textit{Hagendorff}(2024)}]{H2024}
Hagendorff, T. (2024), Deception abilities emerged in large language models, \textit{Proceedings of the National Academy of Sciences}, 121(24), e2317967121.

\bibitem[{\textit{Halliday}(1985)}]{H1985}
Halliday, M. A. K. (1985). \textit{Spoken and written language}, Victoria Deakin University Press, pp. 61-64.

\bibitem[{\textit{He}(2024)}]{He2024}
He, Y. H. (2024), AI-driven research in pure mathematics and theoretical physics, \textit{Nature Reviews Physics}, 1-8.

\bibitem[{\textit{Hevner et al.}(2004)}]{HMet2004}
Hevner, A. R., March, S. T., Park, J., $\&$ Ram, S. (2004), Design science in information systems research, \textit{MIS Quarterly}, 75-105.

\bibitem[{\textit{Hicks et al.}(2015)}]{HWet2015}
Hicks, D., Wouters, P., Waltman, L., De Rijcke, S., $\&$ Rafols, I. (2015), Bibliometrics: the Leiden Manifesto for research metrics, \textit{Nature}, 520(7548), 429-431.

\bibitem[{\textit{Hitsuwari et al.}(2023)}]{HUet2023}
Hitsuwari, J., Ueda, Y., Yun, W., $\&$ Nomura, M. (2023), Does human–AI collaboration lead to more creative art? Aesthetic evaluation of human-made and AI-generated haiku poetry, \textit{Computers in Human Behavior}, 139, 107502.

\bibitem[{\textit{Hu et al.}(2021)}]{HSet2021}
Hu, E. J., Shen, Y., Wallis, P., Allen-Zhu, Z., Li, Y., Wang, S., ... $\&$ Chen, W. (2021), Lora: Low-rank adaptation of large language models, arXiv:2106.09685.

\bibitem[{\textit{Huynh et al.}(2023)}]{HJet2023}
Huynh, J., Jiao, C., Gupta, P., Mehri, S., Bajaj, P., Chaudhary, V., $\&$ Eskenazi, M. (2023), Understanding the effectiveness of very large language models on dialog evaluation, arXiv:2301.12004.

\bibitem[{\textit{Iten et al.}(2020)}]{IMet2020}
Iten, R., Metger, T., Wilming, H., Del Rio, L., $\&$ Renner, R. (2020), Discovering physical concepts with neural networks, \textit{Physical Review Letters}, 124(1), 010508.

\bibitem[{\textit{Jablonka et al.}(2023)}]{JAet2023}
Jablonka, K. M., Ai, Q., Al-Feghali, A., Badhwar, S., Bocarsly, J. D., Bran, A. M., ... $\&$ Blaiszik, B. (2023), 14 examples of how LLMs can transform materials science and chemistry: a reflection on a large language model hackathon, \textit{Digital Discovery}, 2(5), 1233-1250.

\bibitem[{\textit{Jain et al.}(2021)}]{JPet2021}
Jain, H., Padmanabhan, B., Pavlou, P. A., $\&$ Raghu, T. S. (2021), Editorial for the special section on humans, algorithms, and augmented intelligence: The future of work, organizations, and society, \textit{Information Systems Research}, 32(3), 675-687.

\bibitem[{\textit{Jamieson and Nowak}(2011)}]{JN2011}
Jamieson, K. G., $\&$ Nowak, R. (2011), Active ranking using pairwise comparisons, \textit{Advances in NIPS}, 24.

\bibitem[{\textit{Ji et al.}(2023)}]{JLet2023}
Ji, Z., Lee, N., Frieske, R., Yu, T., Su, D., Xu, Y., ... $\&$ Fung, P. (2023), Survey of hallucination in natural language generation, \textit{ACM Computing Surveys}, 55(12), 1-38.

\bibitem[{\textit{Jia et al.}(2024)}]{JKet2024}
Jia, J., Komma, A., Leffel, T., Peng, X., Nagesh, A., Soliman, T., ... $\&$ Kumar, A. (2024), Leveraging LLMs for dialogue quality measurement. 

\bibitem[{\textit{Jin et al.}(2024)}]{JZet2024}
Jin, Y., Zhao, Q., Wang, Y., Chen, H., Zhu, K., Xiao, Y., $\&$ Wang, J. (2024), AgentReview: Exploring Peer Review Dynamics with LLM Agents, arXiv:2406.12708.

\bibitem[{\textit{Kapoor and Narayanan}(2023)}]{KN2023}
Kapoor, S., $\&$ Narayanan, A. (2023), Leakage and the reproducibility crisis in machine-learning-based science, \textit{Patterns}, 4(9).

\bibitem[{\textit{Kendall and da Silva}(2024)}]{Kd2024}
Kendall, G., $\&$ da Silva, J. A. T. (2024), Risks of abuse of large language models, like ChatGPT, in scientific publishing: Authorship, predatory publishing, and paper mills, \textit{Learned Publishing}, 37(1), 55-62.

\bibitem[{\textit{Kerzendorf et al.}(2020)}]{KPet2020}
Kerzendorf, W. E., Patat, F., Bordelon, D., van de Ven, G., $\&$ Pritchard, T. A. (2020), Distributed peer review enhanced with natural language processing and machine learning, \textit{Nature Astronomy}, 4(7), 711-717.

\bibitem[{\textit{Kincaid et al.}(1975)}]{KFet1975}
Kincaid, J. P., Fishburne Jr, R. P., Rogers, R. L., $\&$ Chissom, B. S. (1975), Derivation of new readability formulas (automated readability index, fog count and flesch reading ease formula) for navy enlisted personnel.

\bibitem[{\textit{King et al.}(2004)}]{KWet2004}
King, R. D., Whelan, K. E., Jones, F. M., Reiser, P. G., Bryant, C. H., Muggleton, S. H., ... $\&$ Oliver, S. G. (2004), Functional genomic hypothesis generation and experimentation by a robot scientist, \textit{Nature}, 427(6971), 247-252.

\bibitem[{\textit{King et al.}(2009)}]{KRet2009}
King, R. D., Rowland, J., Oliver, S. G., Young, M., Aubrey, W., Byrne, E., ... $\&$ Clare, A. (2009), The automation of science, \textit{Science}, 324(5923), 85-89.

\bibitem[{\textit{Kitano}(2021)}]{K2021}
Kitano, H. (2021), Nobel Turing Challenge: creating the engine for scientific discovery, \textit{NPJ Systems Biology and Applications}, 7(1), 29.

\bibitem[{\textit{Kousha and Thelwall}(2024)}]{KT2024}
Kousha, K., $\&$ Thelwall, M. (2024), Artificial intelligence to support publishing and peer review: A summary and review, \textit{Learned Publishing}, 37(1), 4-12.

\bibitem[{\textit{Kovanis et al.}(2016)}]{KPet2016}
Kovanis, M., Porcher, R., Ravaud, P., $\&$ Trinquart, L. (2016), The global burden of journal peer review in the biomedical literature: Strong imbalance in the collective enterprise, \textit{PloS One}, 11(11), e0166387.

\bibitem[{\textit{Krenn et al.}(2022)}]{KPet2022}
Krenn, M., Pollice, R., Guo, S. Y., Aldeghi, M., Cervera-Lierta, A., Friederich, P., ... $\&$ Aspuru-Guzik, A. (2022), On scientific understanding with artificial intelligence, \textit{Nature Reviews Physics}, 4(12), 761-769.

\bibitem[{\textit{Kuznetsov et al.}(2024)}]{KAet2024}
Kuznetsov, I., Afzal, O. M., Dercksen, K., Dycke, N., Goldberg, A., Hope, T., ... $\&$ Gurevych, I. (2024), What Can Natural Language Processing Do for Peer Review? arXiv:2405.06563.

\bibitem[{\textit{Kyle}(2019)}]{K2019}
Kyle, K. (2019), Measuring lexical richness, \textit{The Routledge Handbook of Vocabulary Studies}, 454-475.

\bibitem[{\textit{Landhuis}(2016)}]{L2016}
Landhuis, E. (2016), Scientific literature: Information overload, \textit{Nature}, 535(7612), 457-458.

\bibitem[{\textit{Lahitani et al.}(2016)}]{LPet2016}
Lahitani, A. R., Permanasari, A. E., Setiawan, N. A. (2016), Cosine similarity to determine similarity measure: Study case in online essay assessment, \textit{International Conference on Cyber and IT Service Management}.

\bibitem[{\textit{Lee et al.}(2013)}]{LSet2013}
Lee, C. J., Sugimoto, C. R., Zhang, G., $\&$ Cronin, B. (2013), Bias in peer review, \textit{Journal of the American Society for Information Science and Technology}, 64(1), 2-17.

\bibitem[{\textit{Leibbrandt et al.}(2018)}]{LWet2018}
Leibbrandt, A., Wang, L. C., $\&$ Foo, C. (2018), Gender quotas, competitions, and peer review: Experimental evidence on the backlash against women, \textit{Management Science}, 64(8), 3501-3516.

\bibitem[{\textit{Letzing}}(2024)]{Le2024}
Letzing, J. (2024), To Fully Appreciate AI Expectations, Look to the Trillions Being Invested, World Economic Forum.

\bibitem[{\textit{Li et al.}(2024a)}]{LLet2024a}
Li, Z., Liang, C., Peng, J., $\&$ Yin, M. (2024, May), The value, benefits, and concerns of generative ai-powered assistance in writing, In \textit{2024 CHI Conference on Human Factors in Computing Systems} (pp. 1-25).

\bibitem[{\textit{Li et al.}(2024b)}]{LLet2024b}
Li, Z., Liang, C., Peng, J., $\&$ Yin, M. (2024), How Does the Disclosure of AI Assistance Affect the Perceptions of Writing? arXiv:2410.04545.

\bibitem[{\textit{Liang et al.}(2024)}]{LIet2024}
Liang, W., Izzo, Z., Zhang, Y., Lepp, H., Cao, H., Zhao, X., ... $\&$ Zou, J. Y. (2024), Monitoring AI-Modified Content at Scale: A Case Study on the Impact of ChatGPT on AI Conference Peer Reviews, arXiv:2403.07183.

\bibitem[{\textit{Liang et al.}(2024)}]{LZet2024}
Liang, W., Zhang, Y., Cao, H., Wang, B., Ding, D., Yang, X., ... $\&$ Zou, J. (2023). Can large language models provide useful feedback on research papers? A large-scale empirical analysis, arXiv:2310.01783.

\bibitem[{\textit{Liao and Zhang}(2024)}]{LZ2024}
Liao, Z. $\&$ Zhang, C. (2024), Generative AI makes for better scientific writing — but beware the pitfalls, \textit{Nature}, 631(505).

\bibitem[{\textit{Lin}}(2004)]{L2004}
Lin, C. Y. (2004), Rouge: A package for automatic evaluation of summaries, In \textit{Text Summarization Branches Out} (pp. 74-81).

\bibitem[{\textit{Lin}(2024)}]{L2024}
Lin, Z. (2024), How to write effective prompts for large language models, \textit{Nature Human Behaviour}, 1-5.

\bibitem[{\textit{Lindsay et al.}(1993)}]{LBet1993}
Lindsay, R. K., Buchanan, B. G., Feigenbaum, E. A., $\&$ Lederberg, J. (1993), DENDRAL: a case study of the first expert system for scientific hypothesis formation, \textit{Artificial Intelligence}, 61(2), 209-261.

\bibitem[{\textit{Liu et al.}(2016)}]{LLet2016}
Liu, C. W., Lowe, R., Serban, I. V., Noseworthy, M., Charlin, L., $\&$ Pineau, J. (2016), How not to evaluate your dialogue system: An empirical study of unsupervised evaluation metrics for dialogue response generation, arXiv:1603.08023.

\bibitem[{\textit{Liu and Shah}(2023)}]{LS2023}
Liu, R., $\&$ Shah, N. B. (2023), Reviewergpt? an exploratory study on using large language models for paper reviewing, arXiv:2306.00622.

\bibitem[{\textit{Lu et al.}(2024)}]{LLet2024}
Lu, C., Lu, C., Lange, R. T., Foerster, J., Clune, J., $\&$ Ha, D. (2024), The AI Scientist: Towards Fully Automated Open-Ended Scientific Discovery, arXiv:2408.06292.

\bibitem[{\textit{Lu et al.}(2024)}]{LCet2024}
Lu, M. Y., Chen, B., Williamson, D. F., Chen, R. J., Zhao, M., Chow, A. K., ... $\&$ Mahmood, F. (2024), A Multimodal Generative AI Copilot for Human Pathology, \textit{Nature}, 1-3.

\bibitem[{\textit{Lucas et al.}(2023)}]{LUet2023}
Lucas, J., Uchendu, A., Yamashita, M., Lee, J., Rohatgi, S., $\&$ Lee, D. (2023), Fighting fire with fire: The dual role of llms in crafting and detecting elusive disinformation, arXiv:2310.15515.

\bibitem[{\textit{Mahoney}(1999)}]{M1999}
Mahoney, M. V. (1999), Text compression as a test for artificial intelligence, \textit{AAAI/IAAI}, 970.

\bibitem[{\textit{Malički}(2024)}]{M2024}
Malički, M. (2024), Structure peer review to make it more robust, \textit{Nature}, 631(483).

\bibitem[{\textit{Messeri and Crockett}(2024)}]{MC2024}
Messeri, L., $\&$ Crockett, M. J. (2024), Artificial intelligence and illusions of understanding in scientific research, \textit{Nature}, 627(8002), 49-58.

\bibitem[{\textit{Miller et al.}(2023)}]{MSet2023}
Miller, E. J., Steward, B. A., Witkower, Z., Sutherland, C. A., Krumhuber, E. G., $\&$ Dawel, A. (2023), AI hyperrealism: Why AI faces are perceived as more real than human ones, \textit{Psychological Science}, 34(12), 1390-1403.

\bibitem[{\textit{Mitchell}(2023)}]{M2023}
Mitchell, M. (2023), AI’s challenge of understanding the world, \textit{Science}, 382(6671), eadm8175.

\bibitem[{\textit{Naddaf}(2025)}]{N2025}
Naddaf, M. (2025), Will AI take over peer review?, \textit{Nature}, 639, 852-854.

\bibitem[{\textit{Narayanan and Kapoor}(2025)}]{NK2025}
Narayanan, A., $\&$ Kapoor, S. (2025), Why an overreliance on AI-driven modelling is bad for science, \textit{Nature}, 640(8058), 312-314.

\bibitem[{\textit{Negahban et al.}(2012)}]{NOet2012}
Negahban, S., Oh, S., $\&$ Shah, D. (2012), Iterative ranking from pair-wise comparisons, \textit{Advances in Neural Information Processing Systems}, 25.

\bibitem[{\textit{Nightingale and Farid}(2022)}]{NF2022}
Nightingale, S. J., $\&$ Farid, H. (2022), AI-synthesized faces are indistinguishable from real faces and more trustworthy, \textit{Proceedings of the National Academy of Sciences}, 119(8), e2120481119.

\bibitem[{\textit{Norhashim and Hahn}(2024)}]{NH2024}
Norhashim, H., $\&$ Hahn, J. (2024, October), Measuring human-ai value alignment in large language models, In \textit{Proceedings of the AAAI/ACM Conference on AI, Ethics, and Society} (Vol. 7, pp. 1063-1073).

\bibitem[{\textit{Nowogrodzki}(2024)}]{N2024}
Nowogrodzki, J. (2024), ChatGPT for science: how to talk to your data, \textit{Nature}, 631, 924-925.

\bibitem[{\textit{Oketch et al.}(2025)}]{OLet2025}
Oketch, K., Lalor, J. P., Yang, Y., $\&$ Abbasi, A. (2025), Bridging the LLM accessibility divide? performance, fairness, and cost of closed versus open llms for automated essay scoring, arXiv:2503.11827.

\bibitem[{\textit{Padmanabhan et al.}}(2022)]{PFet2022}
Padmanabhan, B., Fang, X., Sahoo, N., $\&$ Burton-Jones, A. (2022), Machine Learning in Information Systems Research, \textit{MIS Quarterly}, 46(1).

\bibitem[{\textit{Papineni et al.}}(2002)]{PRet2002}
Papineni, K., Roukos, S., Ward, T., $\&$ Zhu, W. J. (2002). Bleu: a method for automatic evaluation of machine translation, In \textit{Proceedings of the 40th Annual Meeting of ACL} (pp. 311-318).

\bibitem[{\textit{Peffers et al.}}(2007)]{PTet2007}
Peffers, K., Tuunanen, T., Rothenberger, M. A., $\&$ Chatterjee, S. (2007), A design science research methodology for information systems research, \textit{Journal of Management Information Systems}, 24(3), 45-77.

\bibitem[{\textit{Prat et al.}}(2015)]{PCet2015}
Prat, N., Comyn-Wattiau, I., $\&$ Akoka, J. (2015), A taxonomy of evaluation methods for information systems artifacts, \textit{Journal of Management Information Systems}, 32(3), 229-267.

\bibitem[{\textit{Prillaman}(2024)}]{P2024}
Prillaman, M. (2024), Is ChatGPT making scientists hyper-productive? The highs and lows of using AI, \textit{Nature}.

\bibitem[{\textit{Quidwai et al.}(2023)}]{QLet2023}
Quidwai, M. A., Li, C., $\&$ Dube, P. (2023), Beyond black box ai-generated plagiarism detection: From sentence to document level, arXiv:2306.08122.

\bibitem[{\textit{Rahwan et al.}(2019)}]{RCet2019}
Rahwan, I., Cebrian, M., Obradovich, N., Bongard, J., Bonnefon, J. F., Breazeal, C., ... $\&$ Wellman, M. (2019), Machine behaviour, \textit{Nature}, 568(7753), 477-486.

\bibitem[{\textit{Rai}(2016)}]{R2016}
Rai, A. (2016), Editor's comments: writing a virtuous review, \textit{MIS Quarterly}, 40(3), iii-x.

\bibitem[{\textit{Rai et al.}(2017)}]{RBet2017}
Rai, A., Burton-Jones, A., Chen, H., Gupta, A., Hevner, A. R., Ketter, W., ... \& Yoo, Y. (2017). Editor’s comments: Diversity of design science research, \textit{MIS Quarterly}, 41(1), iii-xviii.


\bibitem[{\textit{Renze and Guven}(2024)}]{RG2024}
Renze, M., $\&$ Guven, E. (2024), Self-Reflection in LLM Agents: Effects on Problem-Solving Performance, arXiv:2405.06682.

\bibitem[{\textit{Richards}(1987)}]{R1987}
Richards, B. (1987), Type/token ratios: What do they really tell us? \textit{Journal of Child Language}, 14(2), 201-209.

\bibitem[{\textit{Robey}(1996)}]{R1996}
Robey, D. (1996), Research commentary: diversity in information systems research: threat, promise, and responsibility, \textit{Information Systems Research}, 7(4), 400-408.

\bibitem[{\textit{Roberson}(2023)}]{R2023}
Robertson, Z. (2023), Gpt4 is slightly helpful for peer-review assistance: A pilot study, arXiv:2307.05492.

\bibitem[{\textit{Sarker et al.}(2023)}]{SWet2023}
Sarker, S., Whitley, E. A., Goh, K. Y., Hong, Y., Mähring, M., Sanyal, P., ... $\&$ Zhao, H. (2023), Some thoughts on reviewing for Information Systems Research and other leading information systems journals, \textit{Information Systems Research}, 34(4), 1321-1338.

\bibitem[{\textit{Sayood}(2017)}]{S2017}
Sayood, K. (2017), \textit{Introduction to Data Compression}, Morgan Kaufmann.

\bibitem[{\textit{Schulhoff et al.}(2024)}]{SIet2024}
Schulhoff, S., Ilie, M., Balepur, N., Kahadze, K., Liu, A., Si, C., ... $\&$ Resnik, P. (2024), The Prompt Report: A Systematic Survey of Prompting Techniques, arXiv:2406.06608.

\bibitem[{\textit{Shah}(2022)}]{S2022}
Shah, N. B. (2022), Challenges, experiments, and computational solutions in peer review, \textit{Communications of the ACM}, 65(6), 76-87.

\bibitem[{\textit{Shah and Wainwright}(2018)}]{SW2018}
Shah, N. B., $\&$ Wainwright, M. J. (2018), Simple, robust and optimal ranking from pairwise comparisons, \textit{Journal of Machine Learning Research}, 18(199), 1-38.

\bibitem[{\textit{Shen et al.}(2024)}]{STet2024}
Shen, J., Tenenholtz, N., Hall, J. B., Alvarez-Melis, D., $\&$ Fusi, N. (2024), Tag-LLM: Repurposing General-Purpose LLMs for Specialized Domains, arXiv:2402.05140.

\bibitem[{\textit{Shi and Lei}(2022)}]{SL2022}
Shi, Y., $\&$ Lei, L. (2022), Lexical richness and text length: An entropy-based perspective, \textit{Journal of Quantitative Linguistics}, 29(1), 62-79.

\bibitem[{\textit{Shmueli et al.}(2023)}]{SMet2023}
Shmueli, G., Maria Colosimo, B., Martens, D., Padman, R., Saar-Tsechansky, M., R. Liu Sheng, O., ... $\&$ Tsui, K. L. (2023), How can IJDS authors, reviewers, and editors use (and misuse) generative AI? \textit{INFORMS Journal on Data Science}, 2(1), 1-9.

\bibitem[{\textit{Sourati and Evans}(2023)}]{SE2023}
Sourati, J., $\&$ Evans, J. A. (2023), Accelerating science with human-aware artificial intelligence, \textit{Nature Human Behaviour}, 7(10), 1682-1696.

\bibitem[{\textit{Spier}(2002)}]{S2002}
Spier, R. (2002), The history of the peer-review process, \textit{TRENDS in Biotechnology}, 20(8), 357-358.

\bibitem[{\textit{Steyn}(2025)}]{S2025}
Steyn, B. (2025), Let’s find ways to measure novelty in science, \textit{Nature}, 642, 543.
 
\bibitem[{\textit{Suarez et al.}(2024)}]{SGet2024}
Suarez, D., Gomez, C., Medaglia, A. L., Akhavan-Tabatabaei, R., $\&$ Grajales, S. (2024), Integrated decision support for disaster risk management: Aiding preparedness and response decisions in wildfire management, \textit{Information Systems Research}, 35(2), 609-628.

\bibitem[{\textit{Sun et al.}(2024)}]{SCet2024}
Sun, L., Chan, A., Chang, Y. S., $\&$ Dow, S. P. (2024, March), ReviewFlow: Intelligent Scaffolding to Support Academic Peer Reviewing, In \textit{International Conference on Intelligent User Interfaces} (pp. 120-137).

\bibitem[{\textit{Sunahara et al.}(2021)}]{SPet2021}
Sunahara, A. S., Perc, M., $\&$ Ribeiro, H. V. (2021), Association between productivity and journal impact across disciplines and career age, \textit{Physical Review Research}, 3(3), 033158.

\bibitem[{\textit{Susarla et al.}(2023)}]{SGet2023}
Susarla, A., Gopal, R., Thatcher, J. B., $\&$ Sarker, S. (2023), The Janus Effect of Generative AI: Charting the Path for Responsible Conduct of Scholarly Activities in Information Systems, \textit{Information Systems Research}, 34(2), 399-408.

\bibitem[{\textit{Taloni et al.}(2023)}]{TSet2023}
Taloni, A., Scorcia, V., $\&$ Giannaccare, G. (2023), Large language model advanced data analysis abuse to create a fake data set in medical research, \textit{JAMA Ophthalmology}, 141(12), 1174-1175.

\bibitem[{\textit{Tao et al.}(2024)}]{TOet2024}
Tao, K., Osman, Z. A., Tzou, P. L., Rhee, S. Y., Ahluwalia, V., $\&$ Shafer, R. W. (2024), GPT-4 performance on querying scientific publications: reproducibility, accuracy, and impact of an instruction sheet, \textit{BMC Medical Research Methodology}, 24(1), 139.

\bibitem[{\textit{Tennant et al.}(2017)}]{TDet2017}
Tennant, J. P., Dugan, J. M., Graziotin, D., Jacques, D. C., Waldner, F., Mietchen, D., ... $\&$ Colomb, J. (2017), A multi-disciplinary perspective on emergent and future innovations in peer review, \textit{F1000Research}.

\bibitem[{\textit{To et al.}(2013)}]{TFet2013}
To, V., Fan, S., $\&$ Thomas, D. (2013). Lexical density and readability: A case study of English textbooks. \textit{Internet Journal of Language, Culture and Society}, 37(37), 61-71.

\bibitem[{\textit{Touvron et al.}(2023)}]{TLet2023}
Touvron, H., Lavril, T., Izacard, G., Martinet, X., Lachaux, M. A., Lacroix, T., ... $\&$ Lample, G. (2023), Llama: Open and efficient foundation language models, arXiv:2302.13971.

\bibitem[{\textit{Van Dinter et al.}(2021)}]{VTet2021}
Van Dinter, R., Tekinerdogan, B., $\&$ Catal, C. (2021), Automation of systematic literature reviews: A systematic literature review, \textit{Information and Software Technology}, 136, 106589.

\bibitem[{\textit{Vert}(2023)}]{V2023}
Vert, J. P. (2023), How will generative AI disrupt data science in drug discovery? \textit{Nature Biotechnology}, 41(6), 750-751.

\bibitem[{\textit{Villalobos et al.}(2024)}]{VHet2024}
Villalobos, P., Ho, A., Sevilla, J., Besiroglu, T., Heim, L., $\&$ Hobbhahn, M. (2024, July), Position: Will we run out of data? Limits of LLM scaling based on human-generated data, In \textit{Forty-first International Conference on Machine Learning}.

\bibitem[{\textit{Wagner et al.}(2022)}]{WLet2022}
Wagner, G., Lukyanenko, R., $\&$ Paré, G. (2022), Artificial intelligence and the conduct of literature reviews, \textit{Journal of Information Technology}, 37(2), 209-226.

\bibitem[{\textit{Waltman}(2016)}]{W2016}
Waltman, L. (2016), A review of the literature on citation impact indicators, \textit{Journal of Informetrics}, 10(2), 365-391.

\bibitem[{\textit{Waltz and Buchanan}(2009)}]{WB2009}
Waltz, D., $\&$ Buchanan, B. G. (2009), Automating science, \textit{Science}, 324(5923), 43-44.

\bibitem[{\textit{Wang et al.}(2023)}]{WFet2023}
Wang, H., Fu, T., Du, Y., Gao, W., Huang, K., Liu, Z., ... $\&$ Zitnik, M. (2023), Scientific discovery in the age of artificial intelligence, \textit{Nature}, 620(7972), 47-60.

\bibitem[{\textit{Wang et al.}(2023)}]{WPet2023}
Wang, W., Pei, S., $\&$ Sun, T. (2023), Unraveling generative ai from a human intelligence perspective: A battery of experiments, Available at SSRN 4543351.

\bibitem[{\textit{Wang et al.}(2013)}]{WJet2013}
Wauthier, F., Jordan, M., $\&$ Jojic, N. (2013, May), Efficient ranking from pairwise comparisons, In \textit{International Conference on Machine Learning} (pp. 109-117). PMLR.

\bibitem[{\textit{Wei et al.}(2022)}]{WWet2022}
Wei, J., Wang, X., Schuurmans, D., Bosma, M., Xia, F., Chi, E., ... $\&$ Zhou, D. (2022), Chain-of-thought prompting elicits reasoning in large language models, \textit{NIPS}, 35, 24824-24837.


\bibitem[{\textit{Wolff}(2019)}]{W2019}
Wolff, J. G. (2019), Information compression as a unifying principle in human learning, perception, and cognition, \textit{Complexity}, 2019(1), 1879746.

\bibitem[{\textit{Xu et al.}(2024)}]{XJet2024}
Xu, Z., Jain, S., $\&$ Kankanhalli, M. (2024), Hallucination is inevitable: An innate limitation of large language models, arXiv:2401.11817.

\bibitem[{\textit{Yang et al.}(2024)}]{YXet2024}
Yang, J., Xu, H., Mirzoyan, S., Chen, T., Liu, Z., Liu, Z., ... $\&$ Wang, S. (2024), Poisoning medical knowledge using large language models, \textit{Nature Machine Intelligence}, 1-13.

\bibitem[{\textit{Yoo et al.}(2024)}]{YHet2024}
Yoo, Y., Henfridsson, O., Kallinikos, J., Gregory, R., Burtch, G., Chatterjee, S., $\&$ Sarker, S. (2024). The next frontiers of digital innovation research, \textit{Information Systems Research}, 35(4), 1507-1523.

\bibitem[{\textit{Yuksekgonul et al.}(2025)}]{YBet2025}
Yuksekgonul, M., Bianchi, F., Boen, J., Liu, S., Lu, P., Huang, Z., ... $\&$ Zou, J. (2025), Optimizing generative AI by backpropagating language model feedback, \textit{Nature}, 639(8055), 609-616.

\bibitem[{\textit{Zdeborová}(2017)}]{Z2017}
Zdeborová, L. (2017), New tool in the box, \textit{Nature Physics}, 13(5), 420-421.

\bibitem[{\textit{Zhao et al.}(2021)}]{ZWet2021}
Zhao, Z., Wallace, E., Feng, S., Klein, D., $\&$ Singh, S. (2021, July), Calibrate before use: Improving few-shot performance of language models, In \textit{ICML} (pp. 12697-12706), PMLR.

\bibitem[{\textit{Zheng et al.}(2023)}]{ZCet2023}
Zheng, L., Chiang, W. L., Sheng, Y., Zhuang, S., Wu, Z., Zhuang, Y., ... $\&$ Stoica, I. (2023), Judging llm-as-a-judge with mt-bench and chatbot arena, \textit{Advances in NIPS}, 36, 46595-46623.

\bibitem[{\textit{Zhou et al.}(2024)}]{ZHet2024}
Zhou, B., Hu, Y., Weng, X., Jia, J., Luo, J., Liu, X., ... $\&$ Huang, L. (2024), TinyLLaVA: A Framework of Small-scale Large Multimodal Models, arXiv:2402.14289.

\bibitem[{\textit{Zhuang et al.}(2024)}]{ZZet2024}
Zhuang, S., Zhuang, H., Koopman, B., $\&$ Zuccon, G. (2024, July), A setwise approach for effective and highly efficient zero-shot ranking with large language models, In \textit{Proceedings of the 47th International ACM SIGIR Conference on Research and Development in Information Retrieval} (pp. 38-47).

\bibitem[{\textit{Zou}(2024)}]{Z2024}
Zou, J. (2024), ChatGPT is transforming peer review — how can we use it responsibly? \textit{Nature}, 635, 10.

\end{thebibliography}

\begin{thebibliography}{99}

\bibitem[{\textit{Chamoun et al.}(2024)}]{CSet2024}
Chamoun, E., Schlichtkrull, M., $\&$ Vlachos, A. (2024, August), Automated Focused Feedback Generation for Scientific Writing Assistance, In \textit{Findings of the Association for Computational Linguistics ACL 2024} (pp. 9742-9763).

\bibitem[{\textit{Chen et al.}(2023)}]{CWet2023}
Chen, Y., Wang, R., Jiang, H., Shi, S., $\&$ Xu, R. (2023), Exploring the use of large language models for reference-free text quality evaluation: An empirical study, arXiv:2304.00723.

\bibitem[{\textit{Drori and Te'eni}(2024)}]{DT2024}
Drori, I., $\&$ Te'eni, D. (2024), Human-in-the-Loop AI Reviewing: Feasibility, Opportunities, and Risks, \textit{Journal of the Association for Information Systems}, 25(1), 98-109.

\bibitem[{\textit{Du et al.}(2024)}]{DWet2024}
Du, J., Wang, Y., Zhao, W., Deng, Z., Liu, S., Lou, R., ... $\&$ Yin, W. (2024), Llms assist nlp researchers: Critique paper (meta-) reviewing, arXiv:2406.16253.

\bibitem[{\textit{Gao et al.}(2024)}]{GBet2024}
Gao, Z., Brantley, K., $\&$ Joachims, T. (2024), Reviewer2: Optimizing review generation through prompt generation, arXiv:2402.10886.

\bibitem[{\textit{Gao et al.}(2025)}]{GRet2025}
Gao, X., Ruan, J., Gao, J., Liu, T., $\&$ Fu, Y. (2025), Reviewagents: Bridging the gap between human and ai-generated paper reviews, arXiv:2503.08506.

\bibitem[{\textit{Jin et al.}(2024)}]{JZet2024}
Jin, Y., Zhao, Q., Wang, Y., Chen, H., Zhu, K., Xiao, Y., $\&$ Wang, J. (2024), AgentReview: Exploring Peer Review Dynamics with LLM Agents, arXiv:2406.12708.

\bibitem[{\textit{Lee et al.}(2025)}]{LCet2025}
Lee, Y., Cho, W., $\&$ Kim, J. (2024), Checkeval: A reliable llm-as-a-judge framework for evaluating text generation using checklists, arXiv:2403.18771.

\bibitem[{\textit{Li et al.}(2025)}]{LLet2025}
Li, M., Li, H., $\&$ Tan, C. (2025), HypoEval: Hypothesis-Guided Evaluation for Natural Language Generation, arXiv:2504.07174.

\bibitem[{\textit{Liang et al.}(2024)}]{LZet2024a}
Liang, W., Zhang, Y., Cao, H., Wang, B., Ding, D., Yang, X., ... $\&$ Zou, J. (2023). Can large language models provide useful feedback on research papers? A large-scale empirical analysis, arXiv:2310.01783.

\bibitem[{\textit{Liu et al.}(2024a)}]{LGet2024}
Liu, Y., Guo, Z., Liang, T., Shareghi, E., Vulić, I., $\&$ Collier, N. (2024), Aligning with logic: Measuring, evaluating and improving logical consistency in large language models, arXiv:2410.02205.

\bibitem[{\textit{Liu et al.}(2024b)}]{LZet2024b}
Liu, Y., Zhou, H., Guo, Z., Shareghi, E., Vulić, I., Korhonen, A., $\&$ Collier, N. (2024), Aligning with human judgement: The role of pairwise preference in large language model evaluators, arXiv:2403.16950.

\bibitem[{\textit{Liu and Shah}(2023)}]{LS2023}
Liu, R., $\&$ Shah, N. B. (2023), Reviewergpt? an exploratory study on using large language models for paper reviewing, arXiv:2306.00622.

\bibitem[{\textit{Liusie et al.}(2023)}]{LMet2023}
Liusie, A., Manakul, P., $\&$ Gales, M. J. (2023), LLM comparative assessment: Zero-shot NLG evaluation through pairwise comparisons using large language models, arXiv:2307.07889.

\bibitem[{\textit{Qin et al.}(2023)}]{QJet2023}
Qin, Z., Jagerman, R., Hui, K., Zhuang, H., Wu, J., Yan, L., ... $\&$ Bendersky, M. (2023), Large language models are effective text rankers with pairwise ranking prompting, arXiv:2306.17563.

\bibitem[{\textit{Shen et al.}(2023)}]{SCet2023}
Shen, C., Cheng, L., Nguyen, X. P., You, Y., $\&$ Bing, L. (2023), Large language models are not yet human-level evaluators for abstractive summarization, arXiv:2305.13091.

\bibitem[{\textit{Shin et al.}(2025)}]{STet2025}
Shin, H., Tang, J., Lee, Y., Kim, N., Lim, H., Cho, J. Y., ... $\&$ Kim, J. (2025), Automatically Evaluating the Paper Reviewing Capability of Large Language Models, arXiv:2502.17086.

\bibitem[{\textit{Tao et al.}(2024)}]{TOet2024}
Tao, K., Osman, Z. A., Tzou, P. L., Rhee, S. Y., Ahluwalia, V., $\&$ Shafer, R. W. (2024), GPT-4 performance on querying scientific publications: reproducibility, accuracy, and impact of an instruction sheet, \textit{BMC Medical Research Methodology}, 24(1), 139.

\bibitem[{\textit{Wang et al.}(2020)}]{WZet2020}
Wang, Q., Zeng, Q., Huang, L., Knight, K., Ji, H., $\&$ Rajani, N. F. (2020, December). ReviewRobot: Explainable Paper Review Generation based on Knowledge Synthesis, In \textit{Proceedings of the 13th International Conference on Natural Language Generation} (pp. 384-397).

\bibitem[{\textit{Yu et al.}(2024)}]{YDet2024}
Yu, J., Ding, Z., Tan, J., Luo, K., Weng, Z., Gong, C., ... $\&$ Li, X. (2024, November), Automated Peer Reviewing in Paper SEA: Standardization, Evaluation, and Analysis, In \textit{Findings of the Association for Computational Linguistics: EMNLP 2024} (pp. 10164-10184).

\bibitem[{\textit{Yuan et al.}(2022)}]{YLet2022}
Yuan, W., Liu, P., $\&$ Neubig, G. (2022), Can we automate scientific reviewing? \textit{Journal of Artificial Intelligence Research}, 75, 171-212.

\bibitem[{\textit{Zeng et al.}(2024a)}]{ZSet2024}
Zeng, Q., Sidhu, M., Blume, A., Chan, H. P., Wang, L., $\&$ Ji, H. (2024), Scientific opinion summarization: Paper meta-review generation dataset, methods, and evaluation, In \textit{AI4Research 2024 Proceedings} (p. 20). Springer Nature.

\bibitem[{\textit{Zeng et al.}(2024b)}]{ZTet2024}
Zeng, Y., Tendolkar, O., Baartmans, R., Wu, Q., Chen, L., $\&$ Wang, H. (2024), LLM-RankFusion: Mitigating Intrinsic Inconsistency in LLM-based Ranking, arXiv:2406.00231.

\bibitem[{\textit{Zheng et al.}(2023)}]{ZCet2023}
Zheng, L., Chiang, W. L., Sheng, Y., Zhuang, S., Wu, Z., Zhuang, Y., ... $\&$ Stoica, I. (2023), Judging llm-as-a-judge with mt-bench and chatbot arena, \textit{Advances in Neural Information Processing Systems}, 36, 46595-46623.

\bibitem[{\textit{Zhou et al.}(2024)}]{ZCet2024}
Zhou, R., Chen, L., $\&$ Yu, K. (2024, May), Is LLM a reliable reviewer? A comprehensive evaluation of LLM on automatic paper reviewing tasks, In \textit{Proceedings of the 2024 Joint International Conference on Computational Linguistics, Language Resources and Evaluation (LREC-COLING 2024)} (pp. 9340-9351).

\bibitem[{\textit{Zhuang et al.}(2024)}]{ZZet2024}
Zhuang, S., Zhuang, H., Koopman, B., $\&$ Zuccon, G. (2024, July), A setwise approach for effective and highly efficient zero-shot ranking with large language models, In \textit{Proceedings of the 47th International ACM SIGIR Conference on Research and Development in Information Retrieval} (pp. 38-47).

\end{thebibliography}
\end{document}